\documentclass{article}
\usepackage{graphicx} 
\usepackage{amsmath}
\usepackage{amssymb}
\usepackage{cite}
\usepackage{subcaption}
\title{Spectral Introspection Identifies Group Training Dynamics in Deep Neural Networks for Neuroimaging}
\author{Bradley T. Baker$^{*,\dagger}$, Vince D. Calhoun$^{*,\dagger,\diamond}$, Sergey M. Plis$^{*,\dagger}$}
\date{$*$The Georgia State University, Georgia Institute of Technology, Emory University Center for Translational Research in Neuroimaging and Data Science (TReNDS)\\$\dagger$Georgia State University\\$\star$Georgia Institute of Technology\\$\diamond$Emory University}

\begin{document}

\maketitle

\begin{abstract}
Neural networks, whice have had a profound effect on how researchers study complex phenomena, do so through a complex, nonlinear mathematical structure which can be difficult for human researchers to interpret. This obstacle can be especially salient when researchers want to better understand the emergence of particular model behaviors such as bias, overfitting, overparametrization, and more. In Neuroimaging, the understanding of how such phenomena emerge is fundamental to preventing and informing users of the potential risks involved in practice. In this work, we present a novel introspection framework for Deep Learning on Neuroimaging data, which exploits the natural structure of gradient computations via the singular value decomposition of gradient components during reverse-mode auto-differentiation. Unlike post-hoc introspection techniques, which require fully-trained models for evaluation, our method allows for the study of training dynamics on the fly, and even more interestingly, allow for the decomposition of gradients based on which samples belong to particular groups of interest. We demonstrate how the gradient spectra for several common deep learning models differ between schizophrenia and control participants from the COBRE study, and illustrate how these trajectories may reveal specific training dynamics helpful for further analysis. 
\end{abstract}

\section{Introduction}

In recent years, a broad library of ``introspection'' methods have emerged to mitigate the difficulty of straightforward interpretation \cite{sundararajan2017axiomatic, selvaraju2017grad, baehrens2010explain, simonyan2013deep, bach2015pixel,shrikumar2017learning}.  In particular, most state of the art research on DNN interpretation has utilized post-hoc analysis of model gradients. For example, the popular integrated gradients measure~\cite{sundararajan2017axiomatic}, saliency maps \cite{shrikumar2017learning}, and other gradient-based methods \cite{baehrens2010explain, simonyan2013deep} have proved popular in applications to convolutional neural networks as they provide simple spatial visualizations which make for easy building of intuition. While these methods are extremely popular, the resulting visualizations tend to be noisy and inconsistent \cite{smilkov2017smoothgrad,montavon2017explaining, samek2016evaluating, sturmfels2020visualizing}, and recent research in the field has attempted to mitigate these limitations, for example by using refining maps based on region attribution density \cite{kapishnikov2019xrai, xu2020attribution}, providing an adaptive interpolation path \cite{kapishnikov2021guided}, or imposing geometric
constraints on produced maps \cite{rahman2022saliency, rahman2022ig}.

While post-hoc gradient analysis techniques can be useful for interpretation on pre-trained models, these methods are not useful for evaluating training dynamics; they can explain where a model currently is, but now how it arrived there. While theoretical analyses of dynamics \cite{saxe2013exact, saxe2019mathematical} provide some guidance to describing model dynamics, new empirical metrics and visualizations are also important for building intuition about model behavior while also accumulating evidence for further theoretical interpretation. 

One existing method which empirically describes model dynamics is the information-plane method \cite{tishby2015deep, shwartz2017opening, baker2022information}, in which the mutual information between layer weights and the input and output spaces is computed and visualized during training. While this method provides fascinating visualizations of dynamic behavior, the general interpretation of the dynamics adhering to the information bottleneck principle \cite{tishby2000information} is not immediately clear \cite{saxe2019information} and can be sensitive to architectural choices or the empirical method used to estimate mutual information. 

We present a novel empirical method for analyzing model learning dynamics, which builds off of our theoretical work studying gradient rank. We call our method AutoSpec, as it utilizes unique opportunities within auto-differnetiation to analyze model learning dynamics. While we previously showed that model architecture can impose theoretical limits on gradient rank, there is more that can be observed at work within auto-differentiation. Namely, we show that the singular values of the gradient and of the component matrices which are used to compute it can be studied they dynamically evolve during model training. Furthermore, because we can analyze gradients on-the-fly within the auto-differentiation mechanism, we have the unique opportunity to analyze these dynamics as they adhere to individual samples from the training data set. As long as these samples have some kind of common group labelling, we can thus do statistical comparisons of gradient trajectories between groups without breaking normal training behavior. This further allows our method to stand out from post-hoc methods which not only occur outside of training, but can only be evaluated for between different classes in disjoint contexts. 

\subsection{Auto Differentiation}

\section{Methods}

In this section, we provide an overview of the methods at work in AutoSpec. First, we provide a brier review of how the gradient of the weights is computed within auto-differentiation, recall how the spectrum is bounded by particular architectural decisions, and show how the spectrum of the gradient relates to the spectrum of the input activations and adjoint variables accumulated within auto-differentiation. We then describe how auto-differentiation uniquely allows us to analyze dynamics between particular groups of samples.

\subsection{Gradient Spectra via Auto-Differentiation}

Recall that during reverse-mode auto-differentiation, the gradient of the weights at a given layer $i$ is computed as a product of the input activations $\mathbb{A}$ and the adjoint variable $\mathbb{\Delta}$ which is the partial derivative computed on the output neurons during back-propagation. Formally, if we have weights $\mathbb{W}_i \in \mathbb{R}^{h_{i-1} \times h_{i}}$ where $h_{i-1}$ and $h_i$ are the number of input and output neurons respectively. Formally, we write this as:

\begin{align}
    \nabla_{\mathbf{W}_i} &= \mathbf{A}^\top_{i-1}\mathbf{\Delta}_i
\end{align}

where $\mathbb{A} \in \mathbb{R}^{N\times h_{i-1}}$ and $\mathbb{\Delta} \in \mathbb{R}^{N\times h_i}$ are the input activations and adjoint variables with batch size $N$.  The Singular Value Decompositions of $\nabla_{\mathbf{W}_i}$, $\mathbf{A}_{i-1}$ and $\mathbf{\Delta}_{i}$ can be written as:
\begin{align}
    \nabla_{\mathbf{W}_i} &= \mathbf{U}_{\nabla_i}\mathbf{\Sigma}_{\nabla_i} \mathbf{V}_{\nabla_i}^\top,\,\,
    \mathbf{A}_{i-1} = \mathbf{U}_{A_{i-1}}\mathbf{\Sigma}_{A_{i-1}} \mathbf{V}_{A_{i-1}}^\top,\,\,
    \mathbf{\Delta}_{i} = \mathbf{U}_{\Delta_{i}}\mathbf{\Sigma}_{\Delta_{i}} \mathbf{V}_{\Delta_{i}}^\top
\end{align}

We can then write $\mathbf{\nabla_{\mathbf{W}_i}}$ as a product of the SVDs of $\mathbf{A}_{i-1}$ and $\mathbf{\Delta}_i$, and use the fact that the $\mathbf{U}$ matrices are orthogonal to get:

\begin{align}
    \nabla_{\mathbf{W}_i} &= \mathbf{V}_{A_{i-1}} \mathbf{\Sigma}_{A_{i-1}} \mathbf{U}_{A_{i-1}} \mathbf{U}_{\Delta_i}^\top \mathbf{\Sigma}_{\Delta_{i}} \mathbf{V}_{\Delta_i}^\top\\
    &= \mathbf{V}_{A_{i-1}} \mathbf{\Sigma}_{A_{i-1}} \mathbf{\Sigma}_{\Delta_{i}} \mathbf{V}_{\Delta_i}^\top
\end{align}

Thus, we can see that the singular values of $\nabla_{\mathbf{W}_i}$ are just the singular values of the first $\min(h_{i-1}, h)$ singular values from the input activations and adjoint variable. 

For the sake of analysis, we can compute the SVD of all three statistics-of-interest just by computing the SVD and the input activations and adjoint matrices; however, because the batch size $N$ might be large, if the gradient is the only statistic of interest, it would often be more efficient to compute the SVD of $\nabla_{\mathbf{W}_i}$ directly.

For networks which utilize parameter tying, such as Convolutional or Recurrent Neural Networks, the gradients are often accumulated over time and space. If desired, our unique perspective from within Auto-Differentiation allows us to peek further into these dimensions, characterizing the spectra not only of the gradient of the weights, but of the gradient of the weights over the dimension of tying.

\subsection{Identifying Group Differences}

One of the advantages of computing our introspection statistics within auto-differentiation is that we have access to the individual gradients for each input sample to the model. Thus, we can evaluate how particular groups of samples individually contribute to the aggregated gradients prior to the aggregated gradient. Formally, if we have $C$ distinct groups of samples in our training set, we can compute the set of $C$ gradients and their SVD as

\begin{align}
\{\nabla_c &= \mathbf{U}_c\mathbf{\Sigma}_c\mathbf{V}_c^\top\}_{c=1}^C
\end{align}

we can then perform statistical testing by accumulating these group-specific statistics during training, and evaluating the differences between groups. For example, if we perform a two-tailed T-test, we can obtain a measure of which training steps were significant between groups if we treat the number of singular values as features. Additionally, we can obtain a measure of per-singular-value significance by taking the T-tests between the transpose. 

\subsection{Data sets and Experimental Design}

To demonstrate the kinds of dynamics which AutoSpec can reveal, we have organized a battery of experiments across different data modalities and architecture types. 

First, we use two numerical data sets to show how AutoSpec allows for dynanic introspection on Multi-Layer Perceptrons, Elman Cell RNNs, and 2-D CNNs. Our choice of numerical data sets are the MNIST and Sinusoid data sets. 

We then move from numerical data to an application in Neuroimaging analysis. We first apply a Multi-Layer perceptron on FreeSurfer volumes, and then move to analyzing functional MRI from the COBRE data set \cite{mayer2013functional}, which is a well-studied data set especially for applications of deep learning \cite{patel2016classification,oh2020identifying,zhu2020weighted,mahmood2020whole}. For our demonstration, we perform Spatially Constrained Independent Component Analysis using the NeuroMark template \cite{du2020neuromark}, which provides us with 53 neurologically relevant spatially independent maps and associated time-series. Using the time-series data, we demonstrate how AutoSpec can reveal group-specific gradient dynamics in 1D and 2D CNNs, LSTMs and the BERT transformer \cite{vaswani2017attention}. We then utiliez the spatial maps (aggregated over the number of components by taking the maximum over the voxel dimension) to demonstrate group-specific gradient dynamics in 3D-CNNs.

For all architectures and all data sets, we evaluate a few different scenarios demonstrating the diversity of dynamics available in gradient spectra. For all models and datasets, we perform two tasks: classification and auto-encoding; however, we only demonstrate one model instance and group differences for the auto-encoding task for clarity. Additionally, we evaluate ``wide/shallow'' (1 layer with 128 neurons) and ``deep/thin'' (3 layers with 8 neurons) variants of each model. We finally compare how a different choice of activation function can affect dynamics by evaluating each model with Sigmoid and Tanh activations in constrast to ReLU activations which we use elsewhere. For all analyses, we perform two-tailed T-Tests between each pair of classes in a given data set to demonstrate where significant group differences emerge within a particular scenario. Group comparisons for all architectural variants are included in supplementary material.

A detailed outline of our experimental is included in \ref{tab:autospec_panels} and \ref{tab:autospec_architectures}. When not otherwise specified, we use ReLU activations in all models, a learning rate of $1\times 10^{-3}$, and 1000 epochs of training. Where possible, we perform full-batch training rather than SGD, as the batch size can artificially restrict the rank of the gradient. For the sake of this demonstration, we set all models to use the same seed, and we only evaluate the models in a one-shot training scenario.

\section{Results}

In this section, we present the empirical results demonstrating how AutoSpec can be used to reveal group gradient dynamics in deep neural networks. All of our figures follow the same format as follows: panels A and B compare the dynamics between a model trained for sample reconstruction (panel A) and for classification (panel B); panels C and D compare dynamics between tanh (panel C) and relu (panel D) activations, panels E and F compare dynamics between ``thin'' (panel E) and ``wide'' (panel F) variants of the base network with 8 and 64 neurons respectively. For a review of how each experiment is organized into panels see table \ref{tab:autospec_panels}.

The experiments on the MNIST data set are included in \ref{ch6-fig:mlp_mnist} and \ref{ch6-fig:mlp_cnn2d} for the MLP and 2DCNN architectures respectively. The experiments for the sinusoid data set evaluated with an RNN can be found in  \ref{ch6-fig:rnn_sin}. The MLP applied to FSL data can be found in \ref{ch6-fig:mlp_fsl}. The experiments on COBRE ICA time-series can be found in figures 6.5, 6.6 and 6.7 for the LSTM, BERT, and 1D-CNN architectures respectively. Finally, the experiments on COBRE ICA spatial maps can be found in figure 6.8. See table \ref{tab:autospec_architectures} for a review of which data set and architecture combinations can be found in particular figures.

\begin{table}
    \centering
        \caption{}
    \begin{tabular}{c|c|c|c}
    \centering
    \textbf{Panel} & \textbf{Task} & \textbf{Model Dims} & \textbf{Activation}\\\hline\hline
    A & Auto-Encoding & [32] & ReLU\\
    B & Classification & [32] & ReLU\\
    C & Auto-Encoding & [32] & Tanh\\
    D & Auto-Encoding & [32] & Sigmoid\\
    E & Auto-Encoding & [8] & ReLU\\
    F & Auto-Encoding & [64] & ReLU\\
    G & Auto-Encoding (Group Differences) & [32] & ReLU\\
    H & Classification (Group Differences) & [32] & ReLU
    \end{tabular}

    \label{tab:autospec_panels}
\end{table}

\begin{table}
    \centering
    \caption{}
    \begin{tabular}{c|c|c|c}
    \centering
    \textbf{Dataset} & \textbf{Modality} &\textbf{Model} & \textbf{Figure}\\\hline\hline
    MNIST & Image & MLP & \ref{ch6-fig:mlp_mnist}\\
    MNIST & Image & 2DCNN & \ref{ch6-fig:mlp_cnn2d}\\
    SINUSOID & Time-Series & RNN & \ref{ch6-fig:rnn_sin}\\
    FreeSurfer & Tabular & MLP & \ref{ch6-fig:mlp_fsl}\\
    COBRE & ICA Time-Series & LSTM & \ref{ch6-fig:lstm_cobre}\\
    COBRE & ICA Time-Series & BERT & \ref{ch6-fig:bert_cobre}\\
    COBRE & ICA Time-Series & 1D-CNN & \ref{ch6-fig:cnn1d_cobre}\\    
    COBRE & ICA Spatial Mapps & 3D-CNN & \ref{ch6-fig:cnn3d_cobre}
    \end{tabular}

    \label{tab:autospec_architectures}
\end{table}

\section{Discussion}

In this work, we have introduced a new method for model introspection called AutoSpec, in which we utilize the singular value decomposition to study the dynamic evolution of the spectrum of the gradient and its component matrices. Our method reveals fascinating dynamics at work in a number of model architectures, and also allows us to identify unique dynamics belonging to particular groups within a data set. We demonstrated our model on numerical datasets for sequence and image reconstruction and classification, and also demonstrated the identification of group differences on a real neuroimaging data set. 

We will provide a brief discussion of some of the observed differences in dynamics; however, we would like to stipulate that any general conclusions regarding these dynamics will require further experimentation testing specific architecture choices systematically over many repetitions, seeds, and data sets. First of all we notice that for all data sets and architectures, the dynamics we find with AutoSpec show very different trajectories between and AutoEncoding and Classification task (see \ref{ch6-fig:mlp_mnist_a} and \ref{ch6-fig:mlp_mnist_b} for the MLP applied to MNIST for example). Particularly the input layers across all cases are affected by the change in the output structure, and corresponding singular values in output layers are also different between the two tasks. The choice of activation function can affect the gradient spectrum as well - we notice for example when we compare the Sigmoid and Tanh activated LSTM (\ref{ch6-fig:lstm_cobre}) and 1D CNN (\ref{ch6-fig:cnn1d_cobre}), we can see slight differences in the trajectory toward the start of training, but the overall evolution stays the same. In the 2D (\ref{ch6-fig:mlp_cnn2d}) and 3D (\ref{ch6-fig:cnn3d_cobre}) CNNs, however, we notice that Sigmoid and Tanh activations affect the spectrum quite differently, with the effects in the 3D CNN particularly noticeable early in the training period. The dynamics in the layers pulled from the BERT architecture (\ref{ch6-fig:bert_cobre}) are difficult to interpret as singular values tend to jump drastically between individual epochs, perhaps indicated; however, even in this noisy scenario, the BERT architecture reflects a general trend of shrinking singular values which occurs in other architures during training. 

In each of our analyses, we also compute group differences between the observed dynamics. In general, smaller observed singular values tend to change more drastically; however, because the values are so small and near the threshold for machine epsilon of floating point numbers, it is unclear if any observed differences are merely the result of noise. We do see larger singular values show significant differences during training across a few different architectures however, and interestingly, these differences are often contained with a few layers. For example, the MLP autoencoder trained on FSL data shows significant differences between SZ and HC groups in the middle singular values of the output layer (see \ref{ch6-fig:mlp_fsl_g}), while the corresponding classifier shows more group differences in the input layer (see \ref{ch6-fig:mlp_fsl_h}).  The LSTM, BERT and 1D CNN models all show significant differences between male SZ and HC groups, with the LSTM showing differences mostly the Hidden-to-Hidden gradients (see \ref{ch6-fig:lstm_cobre_g} and \ref{ch6-fig:lstm_cobre_h}), the 1D CNN showing significant differences at the output layer (see \ref{ch6-fig:cnn1d_cobre_g} and \ref{ch6-fig:cnn1d_cobre_h}), and BERT showing differences across the entire model for the autoencoder task (see \ref{ch6-fig:bert_cobre_g}) with almost no significant differences in the classifier ((see \ref{ch6-fig:bert_cobre_h}). While more work is needed to investigate what these differences mean for how the model treats different sample groups differently, our finding of significant effects across multiple tasks and architectures demonstrates that these kinds of dynamics may be useful for further, targeted investigation. 

The major limitation of our AutoSpec framework is the computational overhead required for computing the singular values on-the-fly during training. In general for a matrix $A \in \mathbb{R}^{m\times n}$, the complexity required for computing the SVD is $O(\min(m n^2, m^2 n))$. If we want to analyze $L$ different layers at $T$ different training periods, the complexity of our method further increases to $O(LT\min(m n^2, m^2 n))$. Even more overhead is accumulated if we perform group-specific analyses. As such, AutoSpec in practical usage will require long runtimes during training to perform a comprehensive analysis; however, limiting the analysis to specific singular values, layers, periods during training, or groups would reduce this overhead. One potential direction of future work could derive an analytic update to the gradient spectrum based on the initial SVD of the weights and input data, and thus avoid recomputing the SVD entirely for each update. 

The computational overhead has limited our experiments in this work to smaller architectures, or variants of large architectures with smaller dimension sizes. In future work, we would like to expand the analysis to one or two larger scale architectures to provide principled insights into the dynamics of these models; however, for the sake of surveying many model types in this work we have kept the scope smaller for scalability. 

\begin{figure*}
     \caption{Differences in Auto-Differentiation Spectra Dynamics for the first 4 classes in the MNIST data set, trained with various architectures and tasks with a Multi-Layer Perceptron.}
     \centering
         \begin{subfigure}[b]{1\linewidth}
         \centering
         \includegraphics[width=\linewidth] {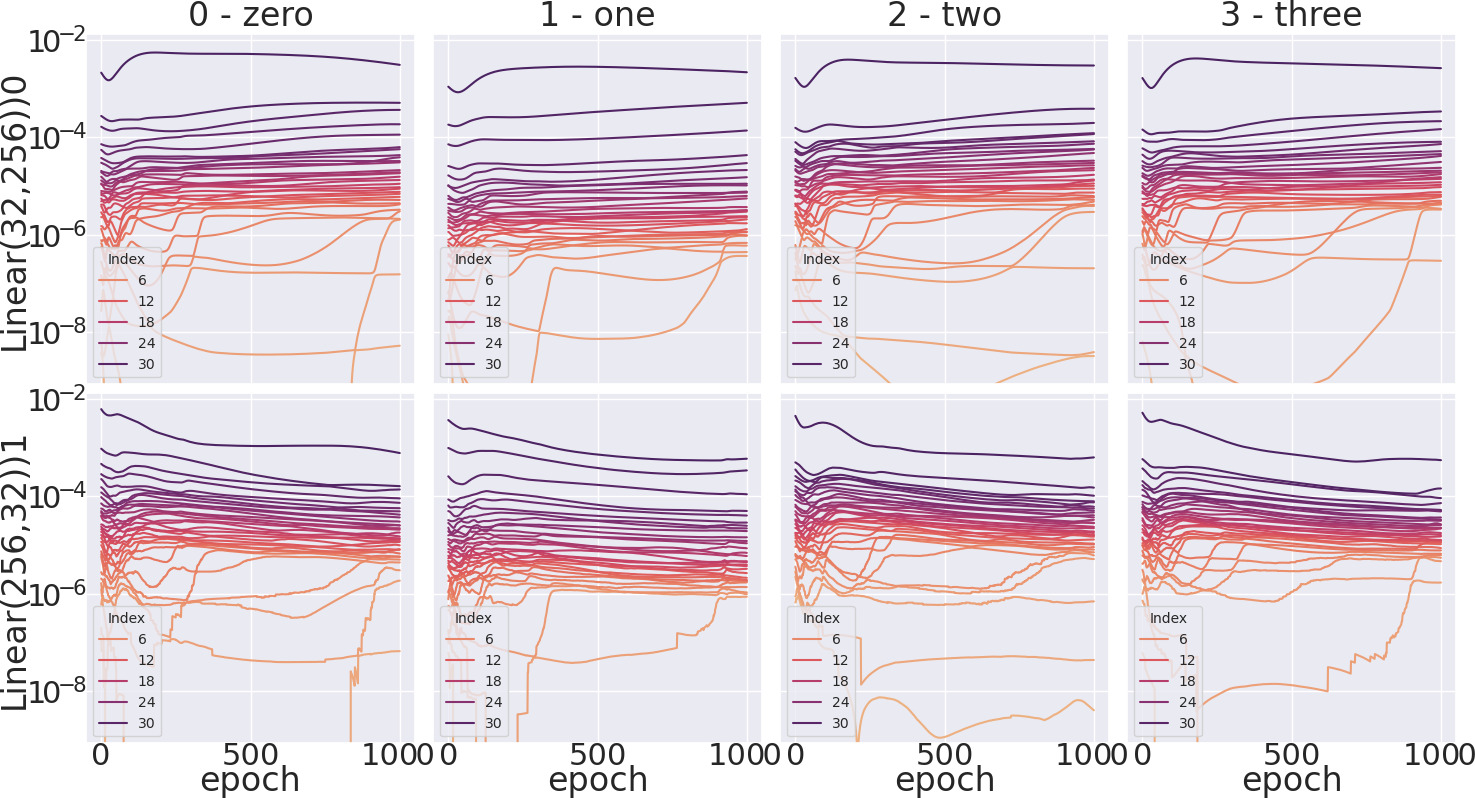}
         \caption{MLP Autoencoder with hidden dim 32 and ReLU activations}
         \label{ch6-fig:mlp_mnist_a}
     \end{subfigure}
              \begin{subfigure}[b]{1\linewidth}
         \centering
         \includegraphics[width=\linewidth] {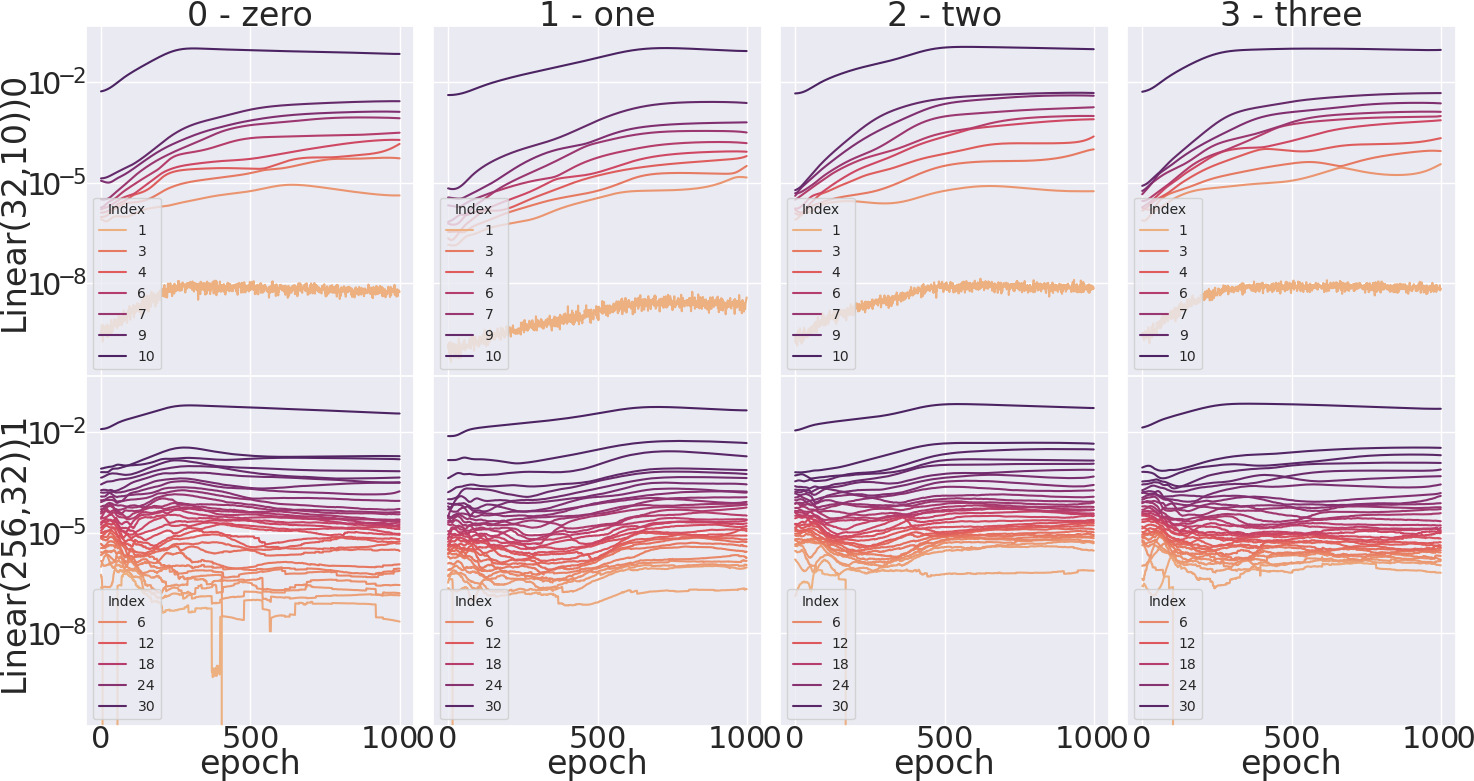}
         \caption{MLP Classifier with hidden dim 32 and ReLU activations}
         \label{ch6-fig:mlp_mnist_b}
     \end{subfigure}
\end{figure*}
\begin{figure*}\ContinuedFloat
\centering
     \begin{subfigure}[b]{1\linewidth}
         \centering
         \includegraphics[width=\linewidth] {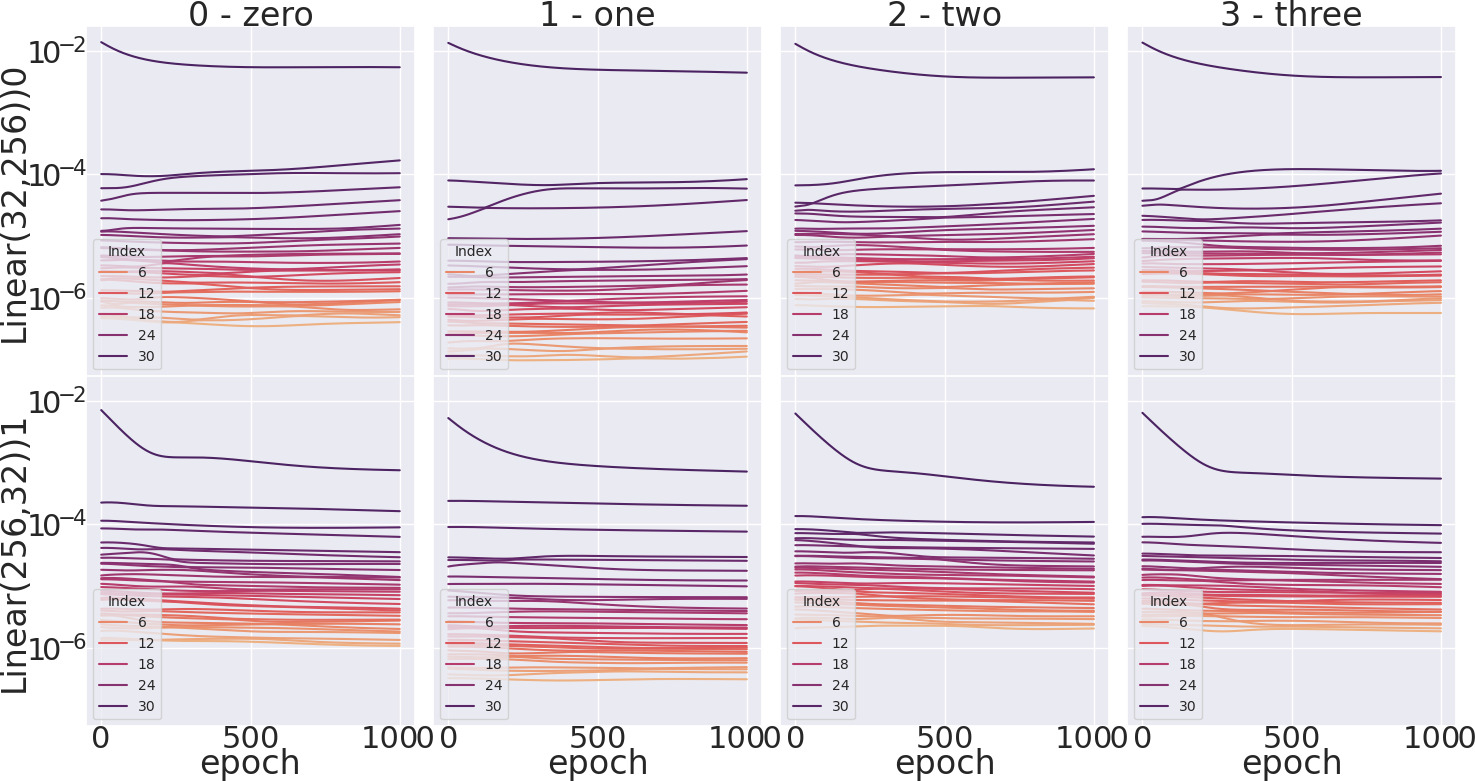}
         \caption{MLP Autoencoder with hidden dim 32 and Sigmoid activations}
         \label{ch6-fig:mlp_mnist_c}
     \end{subfigure}
     \begin{subfigure}[b]{1\linewidth}
         \centering
         \includegraphics[width=\linewidth] {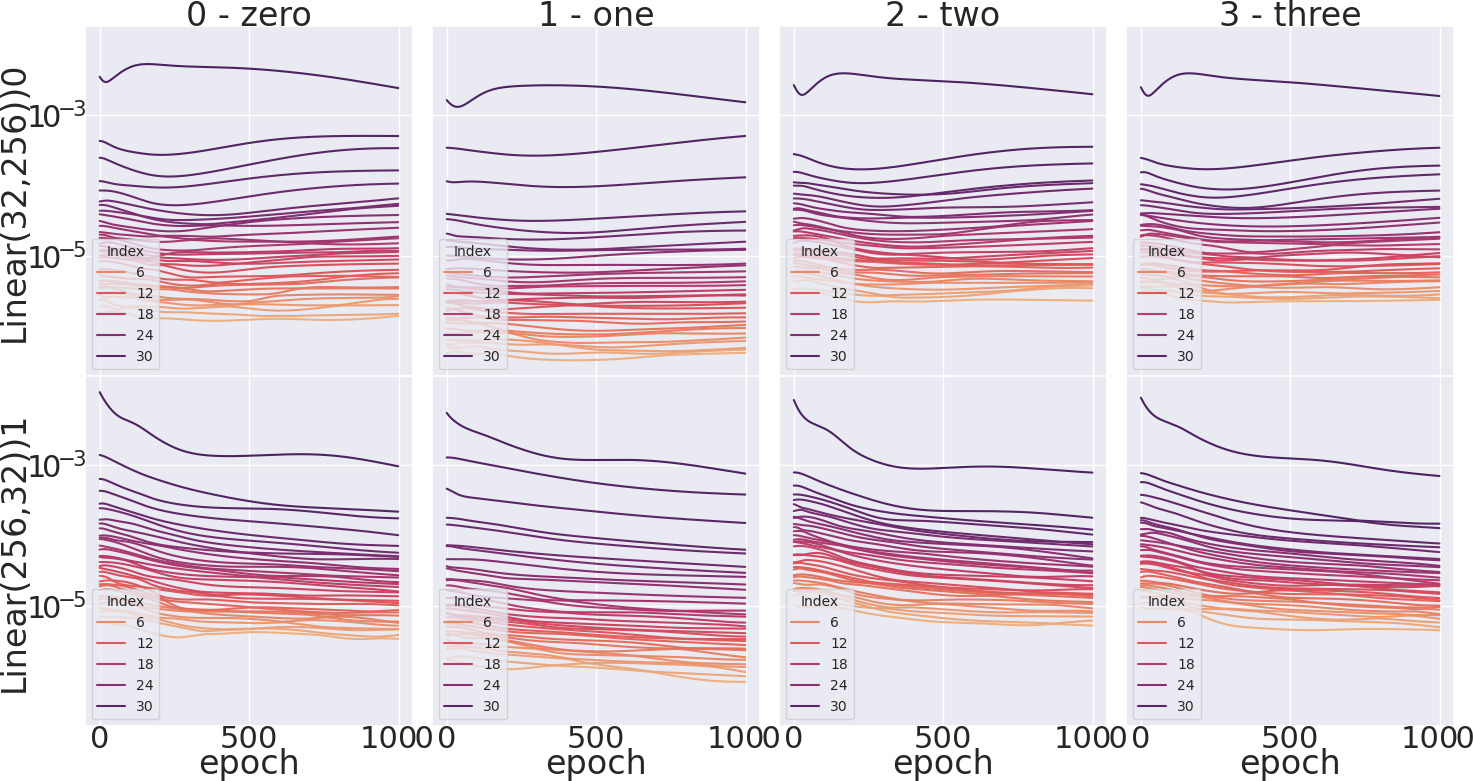}
         \caption{MLP Autoencoder with hidden dim 32 and Tanh activations}
         \label{ch6-fig:mlp_mnist_d}
     \end{subfigure}
\end{figure*}
\begin{figure*}\ContinuedFloat
\centering
     \begin{subfigure}[b]{1\linewidth}
         \centering
         \includegraphics[width=\linewidth] {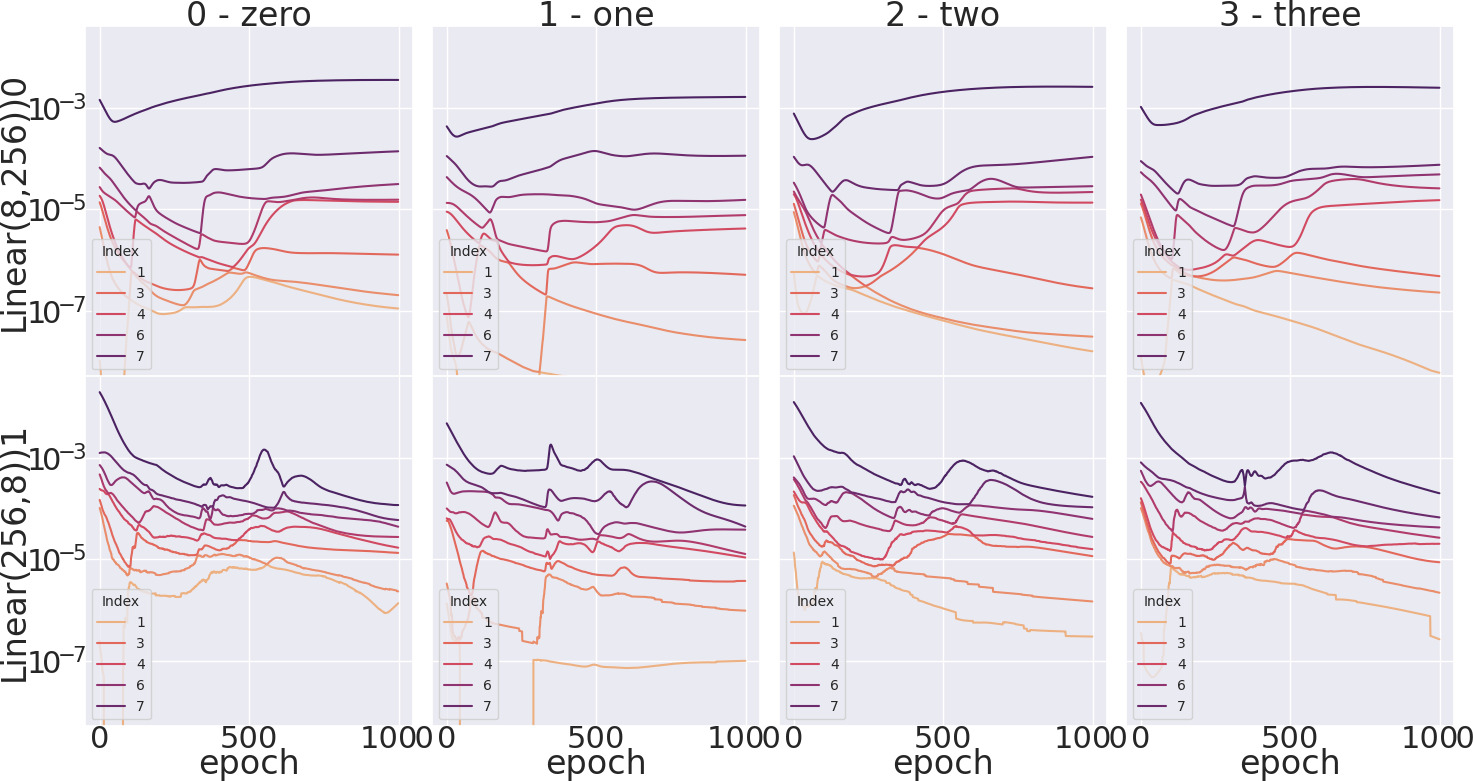}
         \caption{MLP Autoencoder with hidden dim 8 and ReLU activations}
         \label{ch6-fig:mlp_mnist_e}
     \end{subfigure}
          \begin{subfigure}[b]{1\linewidth}
         \centering
         \includegraphics[width=\linewidth] {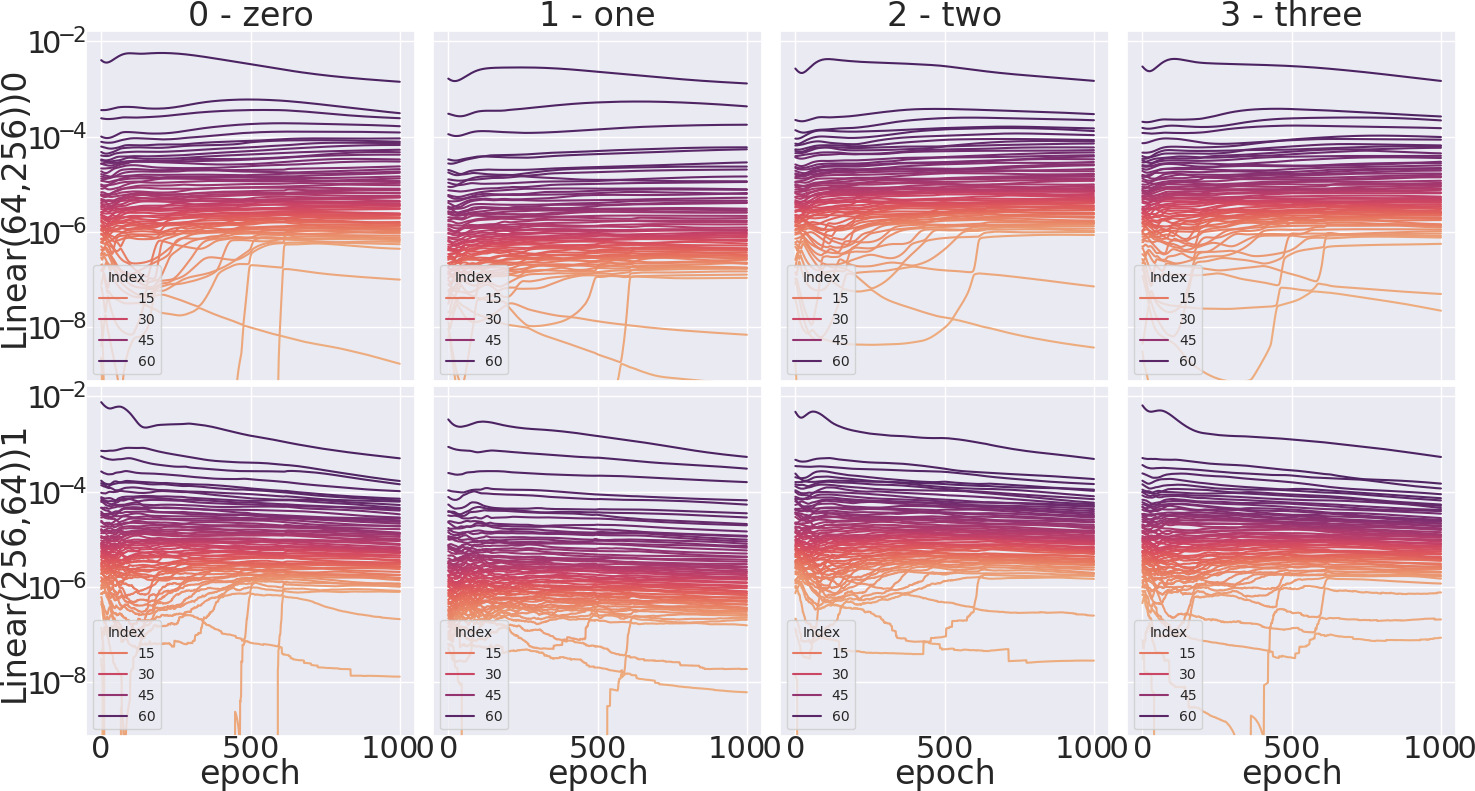}
         \caption{MLP Autoencoder with hidden dim 64 and ReLU activations}
         \label{ch6-fig:mlp_mnist_f}
     \end{subfigure}
\end{figure*}
\begin{figure*}\ContinuedFloat
     \begin{subfigure}[b]{1\linewidth}
         \centering
         \includegraphics[width=\linewidth] {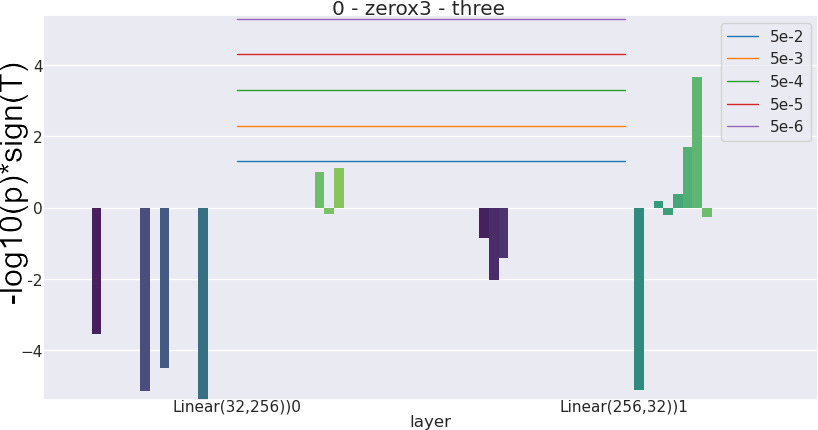}
         \caption{Significant spectral differences between groups 0 and 3: MLP Autoencoder with hidden dim 32 and ReLU activations}
         \label{ch6-fig:mlp_mnist_g}
     \end{subfigure}
     \begin{subfigure}[b]{1\linewidth}
         \centering
         \includegraphics[width=\linewidth] {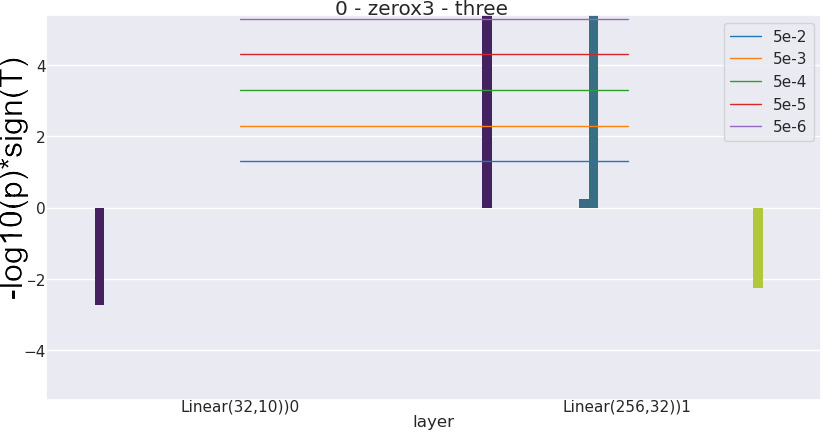}
         \caption{Significant spectral differences between groups 0 and 3: MLP Classifier with hidden dim 32 and ReLU activations}
         \label{ch6-fig:mlp_mnist_h}
     \end{subfigure}
     \caption{Differences in Auto-Differentiation Spectra Dynamics for the first 4 classes in the MNIST data set, trained with various architectures and tasks with a Multi-Layer Perceptron.}
     \label{ch6-fig:mlp_mnist}
\end{figure*}

\begin{figure*}
\caption{Differences in Auto-Differentiation Spectra Dynamics for the first 4 classes in the MNIST data set, trained with various architectures and tasks with a 2D CNN.}
         \begin{subfigure}[b]{\linewidth}
         \centering
         \includegraphics[width=\linewidth] {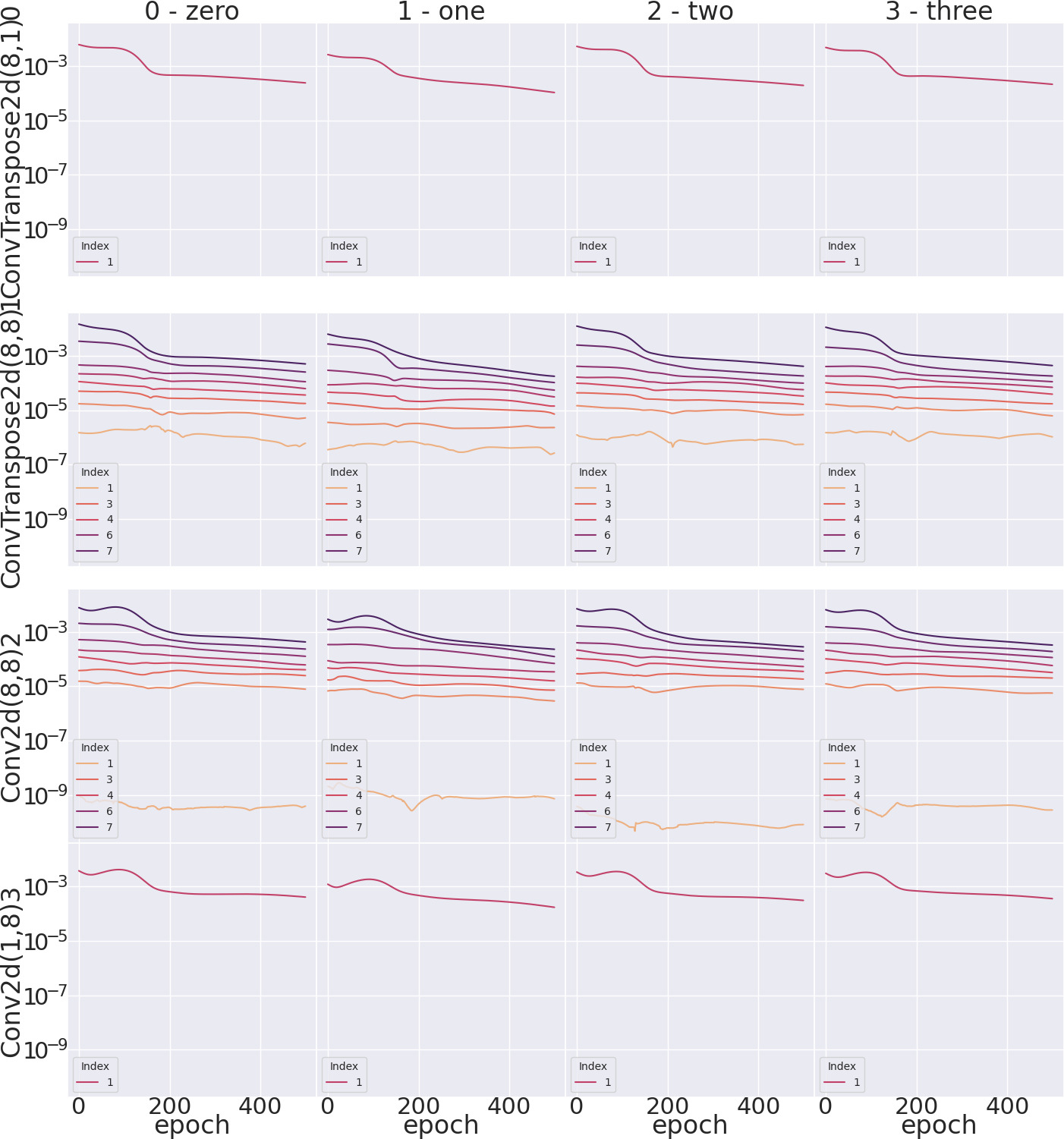}
         \caption{CNN2D Autoencoder with hidden dim 8 and ReLU activations}
         \label{ch6-fig:mlp_cnn2d_a}
     \end{subfigure}
\end{figure*}
\begin{figure*}\ContinuedFloat
              \begin{subfigure}[b]{\linewidth}
         \centering
         \includegraphics[width=\linewidth] {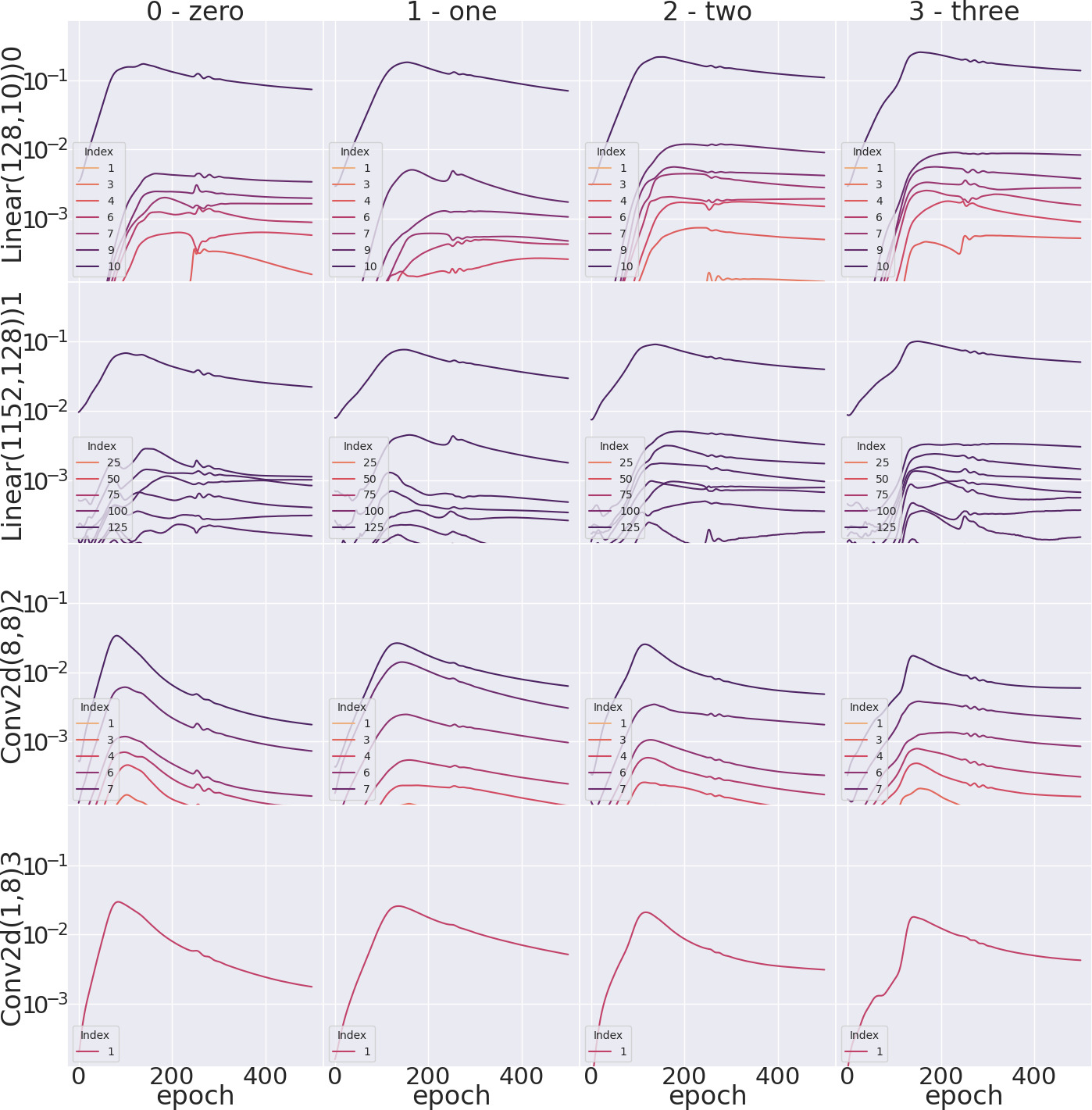}
         \caption{CNN2D Classifier with hidden dim 8 and ReLU activations}
         \label{ch6-fig:mlp_cnn2d_b}
     \end{subfigure}
\end{figure*}
\begin{figure*}\ContinuedFloat
     \begin{subfigure}[b]{\linewidth}
         \centering
         \includegraphics[width=\linewidth] {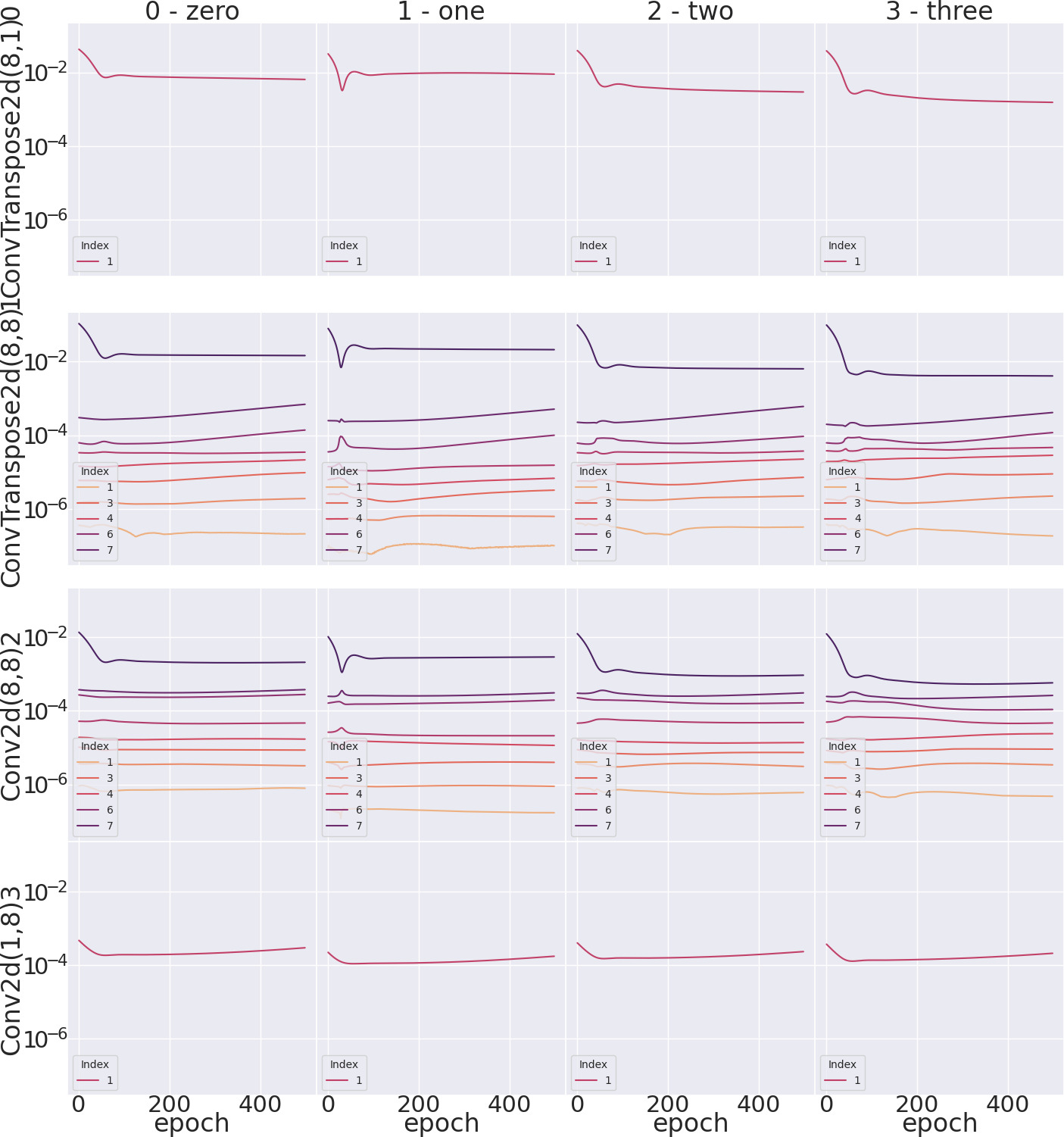}
         \caption{CNN2D Autoencoder with hidden dim 8 and Sigmoid activations}
         \label{ch6-fig:mlp_cnn2d_c}
     \end{subfigure}
\end{figure*}
\begin{figure*}\ContinuedFloat
     \begin{subfigure}[b]{\linewidth}
         \centering
         \includegraphics[width=\linewidth] {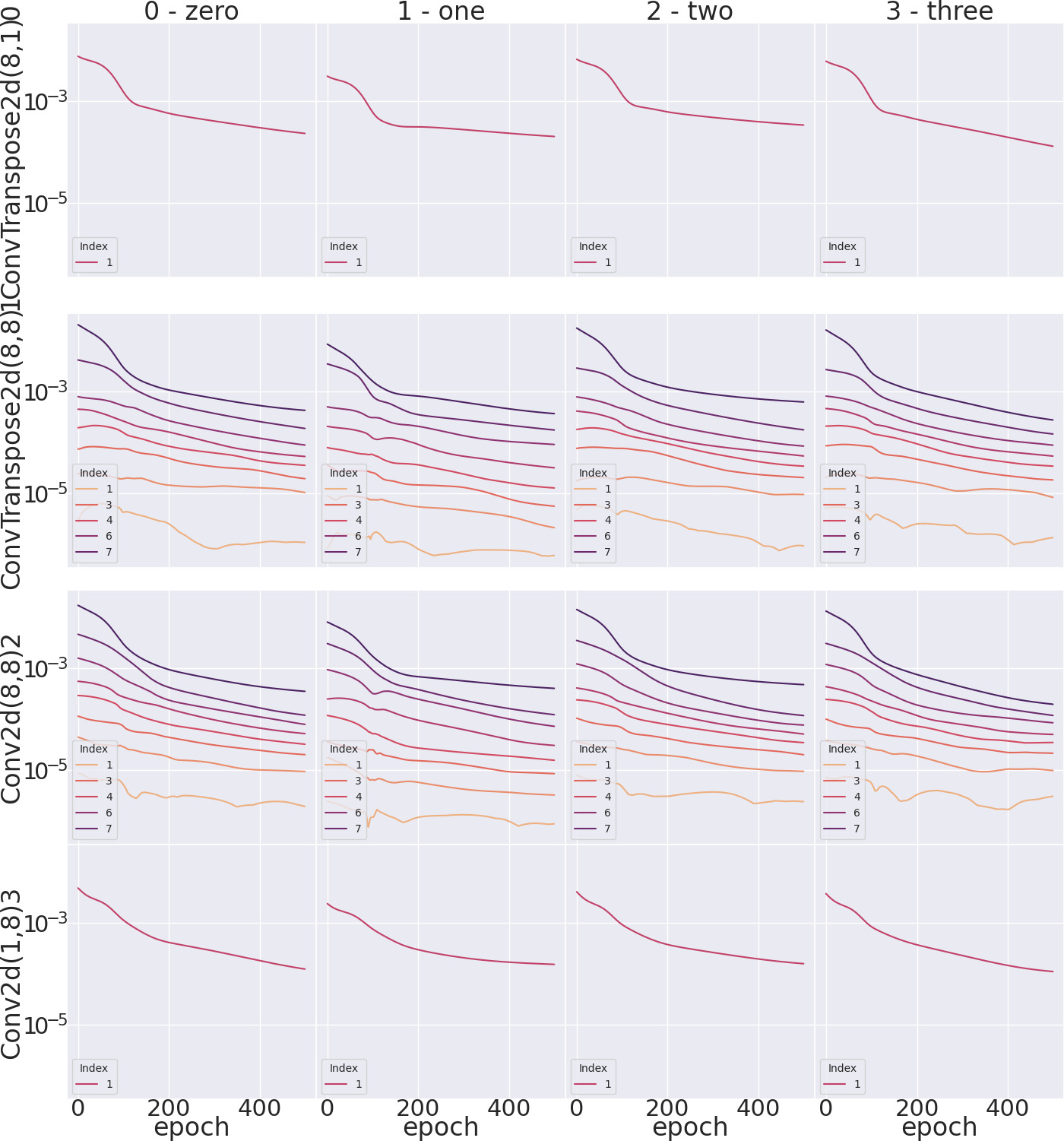}
         \caption{CNN2D Autoencoder with hidden dim 8 and Tanh activations}
         \label{ch6-fig:mlp_cnn2d_d}
     \end{subfigure}
\end{figure*}
\begin{figure*}\ContinuedFloat
     \begin{subfigure}[b]{\linewidth}
         \centering
         \includegraphics[width=\linewidth] {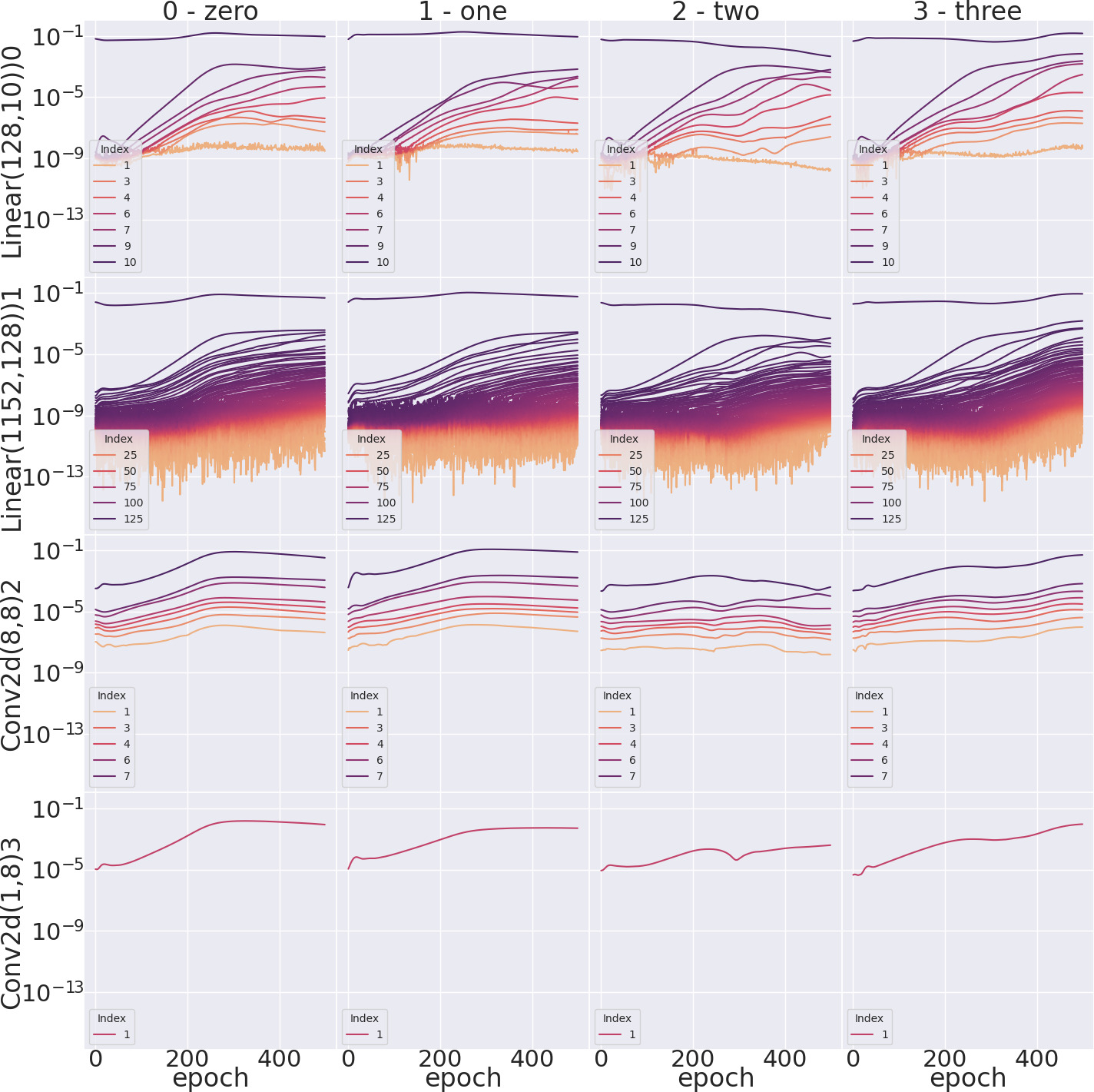}
         \caption{CNN2D Classifier with hidden dim 8 and Sigmoid activations}
         \label{ch6-fig:mlp_cnn2d_e}
     \end{subfigure}
\end{figure*}
\begin{figure*}\ContinuedFloat
          \begin{subfigure}[b]{\linewidth}
         \centering
         \includegraphics[width=\linewidth] {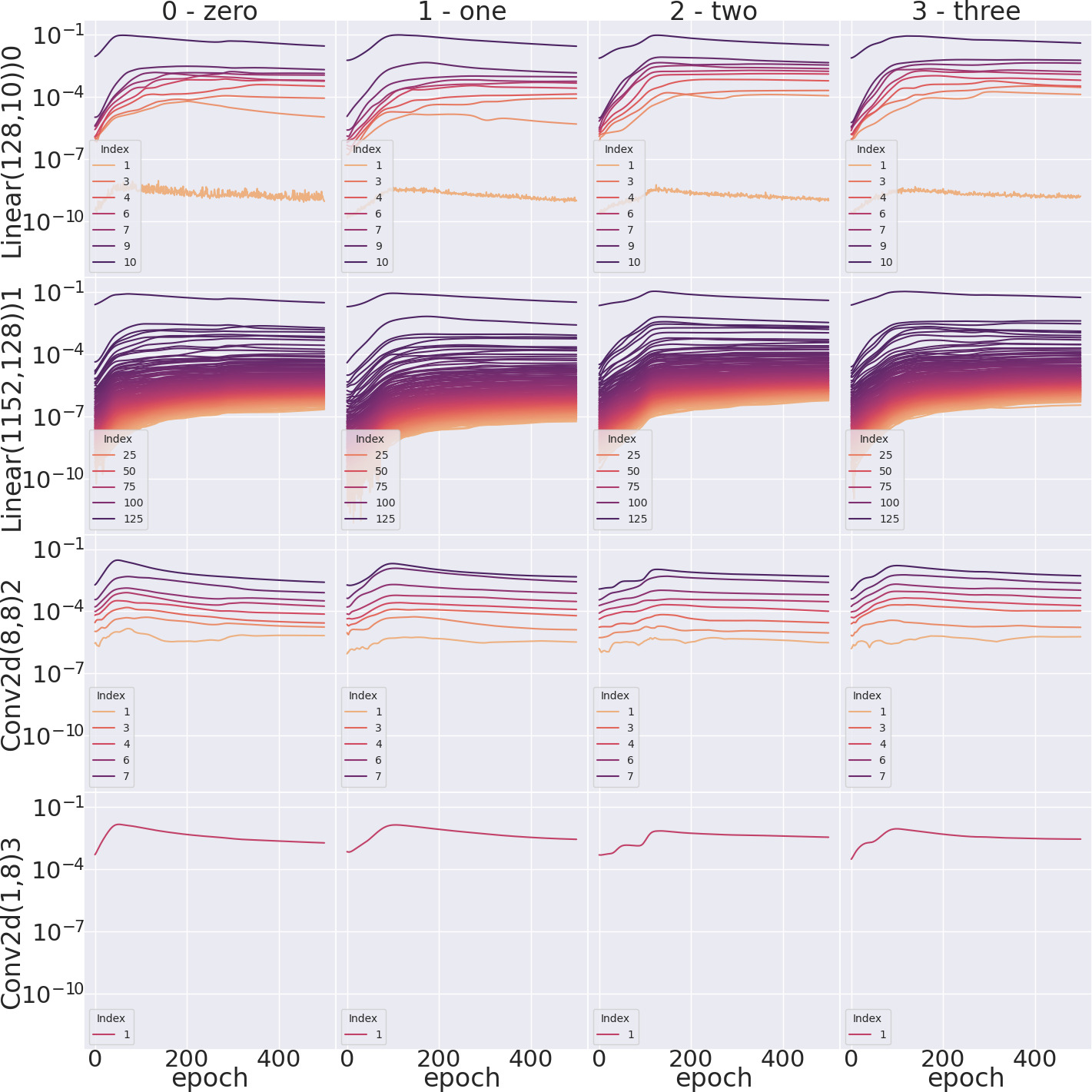}
         \caption{CNN2D Classifier with hidden dim 8 and Tanh activations}
         \label{ch6-fig:mlp_cnn2d_f}
     \end{subfigure}
\end{figure*}
\begin{figure*}\ContinuedFloat
     \begin{subfigure}[b]{1\linewidth}
         \centering
         \includegraphics[width=\linewidth] {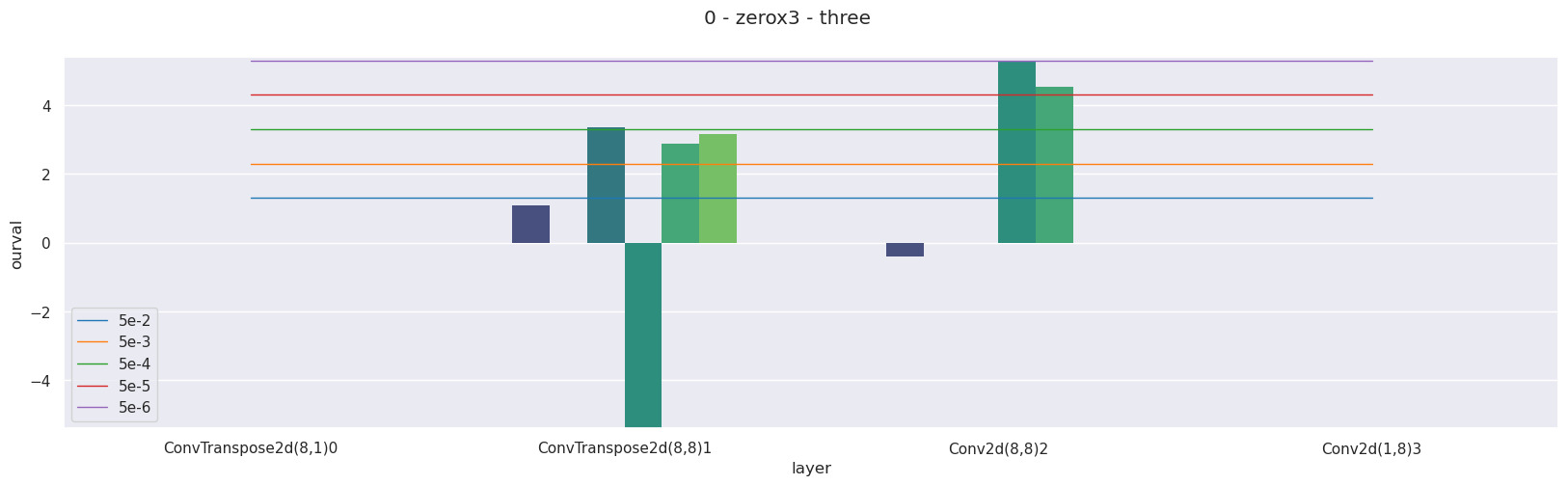}
         \caption{Significant spectral differences between groups 0 and 3: CNN2D Autoencoder with hidden dim 8 and ReLU activations}
         \label{ch6-fig:mlp_cnn2d_g}
     \end{subfigure}
     \begin{subfigure}[b]{1\linewidth}
         \centering
         \includegraphics[width=\linewidth] {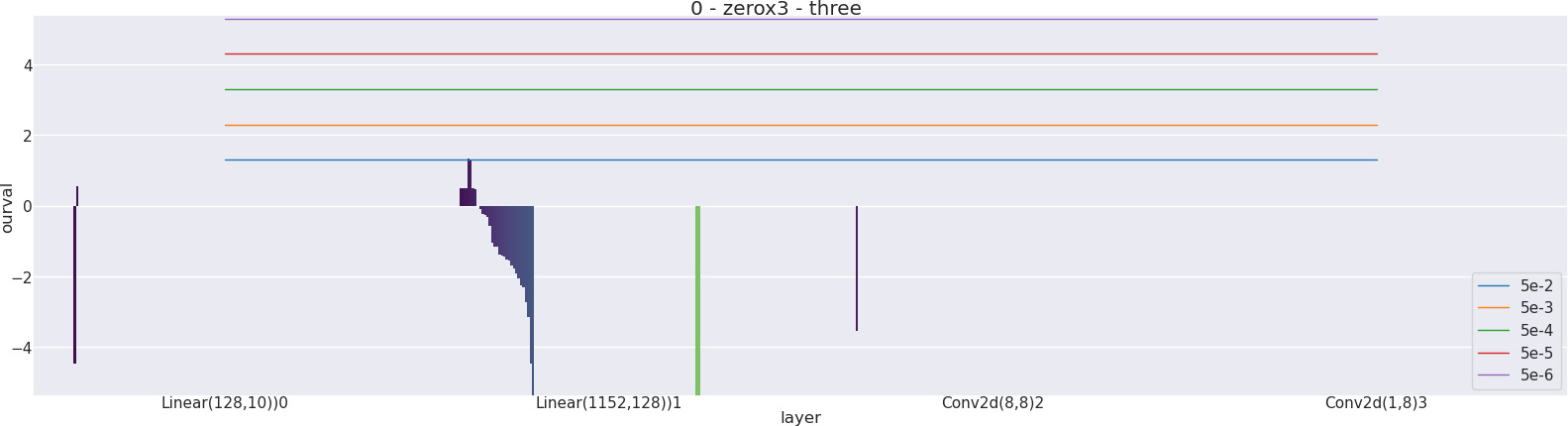}
         \caption{Significant spectral differences between groups 0 and 3: CNN2D Classifier with hidden dim 8 and ReLU activations}
         \label{ch6-fig:mlp_cnn2d_h}
     \end{subfigure}
     \caption{Differences in Auto-Differentiation Spectra Dynamics for the first 4 classes in the MNIST data set, trained with various architectures and tasks with a 2D CNN.}
     \label{ch6-fig:mlp_cnn2d}
\end{figure*}

\begin{figure*}
     \caption{Differences in Auto-Differentiation Spectra Dynamics for the first 4 classes in the Sinusoid data set, trained with various architectures and tasks with an RNN}
     \centering
         \begin{subfigure}[b]{\linewidth}
         \centering
         \includegraphics[width=\linewidth] {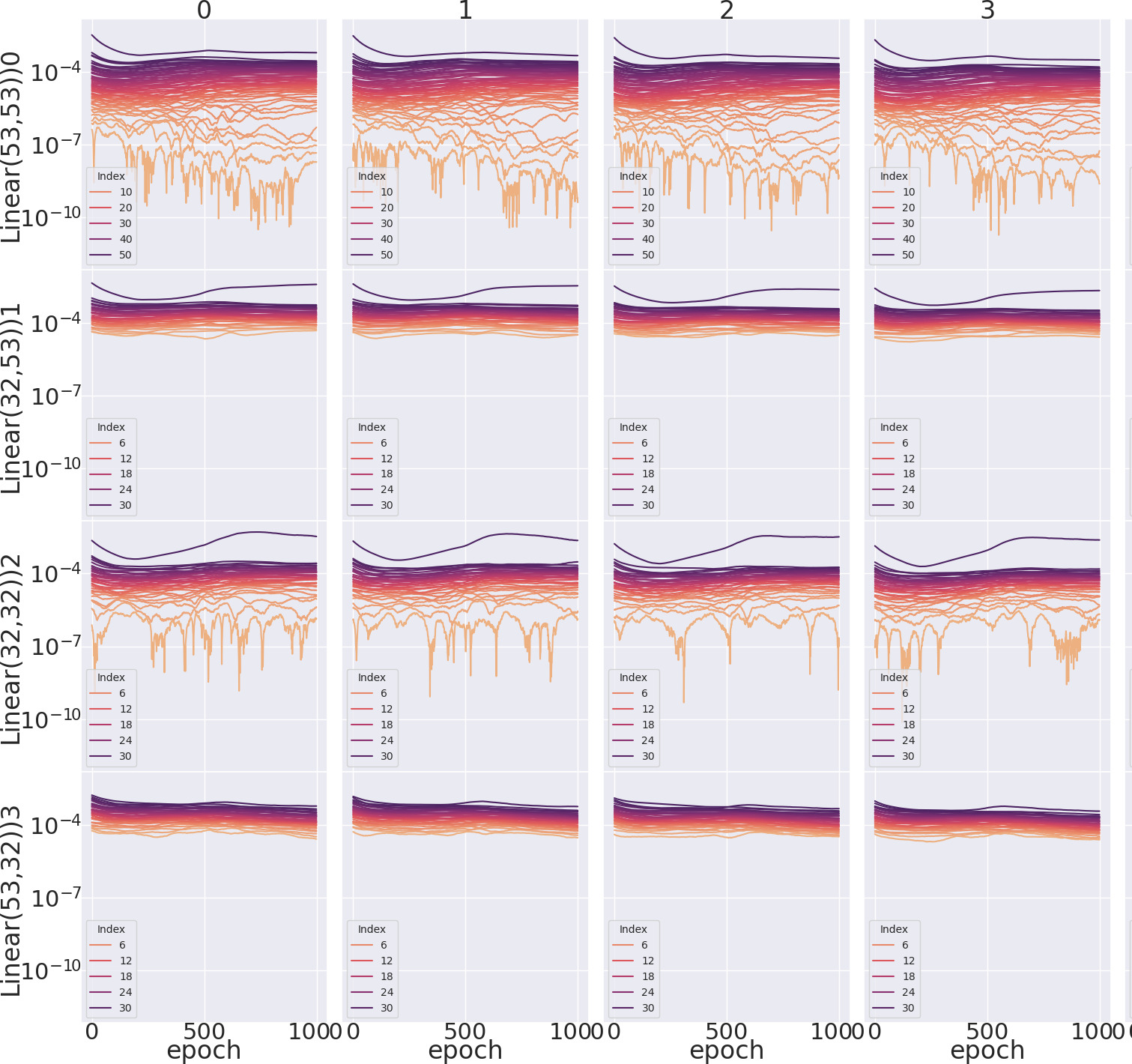}
         \caption{RNN Autoencoder with hidden dim 32 and ReLU activations}
         \label{ch6-fig:rnn_sin_a}
     \end{subfigure}
\end{figure*}
\begin{figure*}\ContinuedFloat
              \begin{subfigure}[b]{\linewidth}
         \centering
         \includegraphics[width=\linewidth] {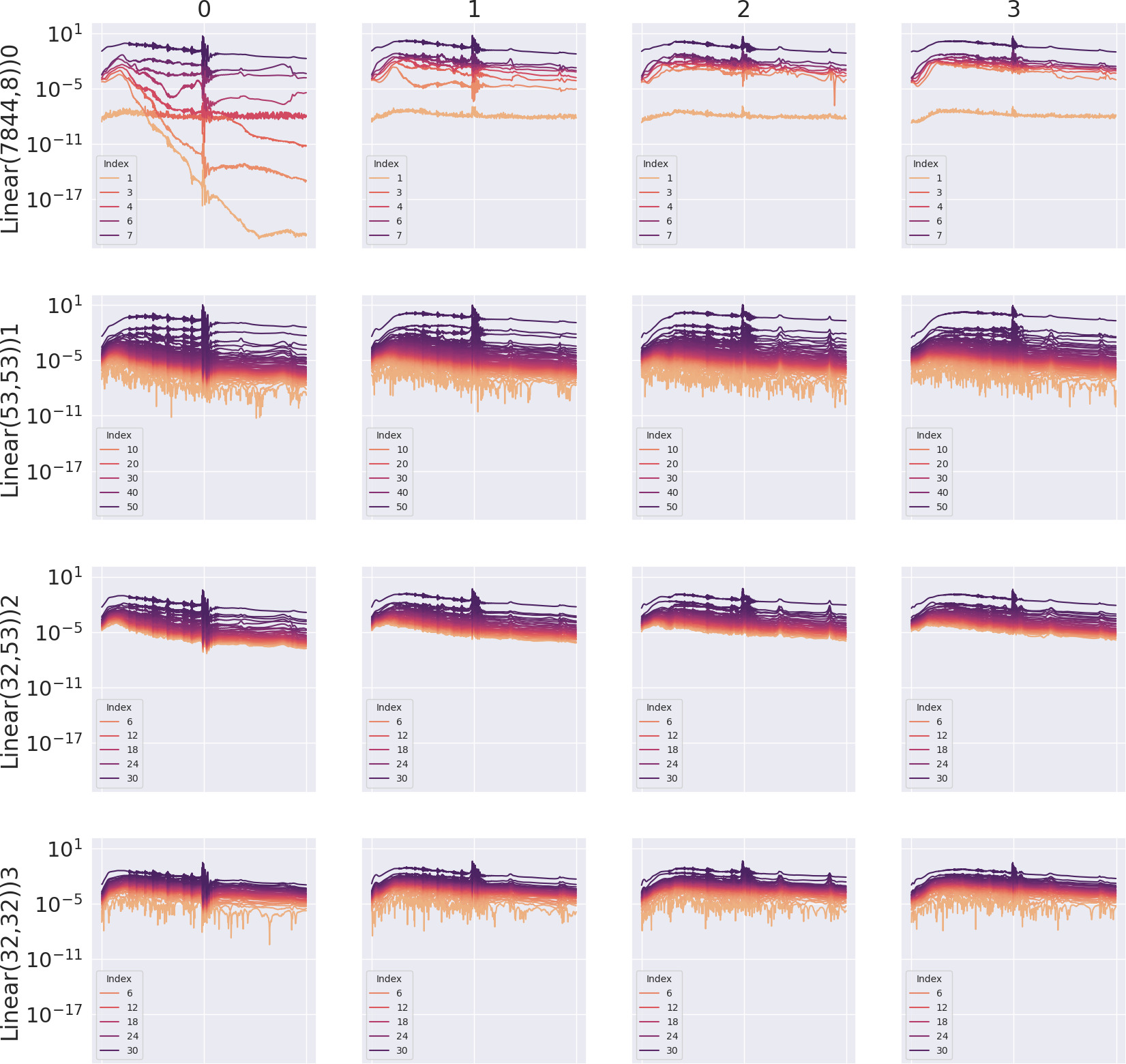}
         \caption{RNN Classifier with hidden dim 32 and ReLU activations}
         \label{ch6-fig:rnn_sin_b}
     \end{subfigure}
\end{figure*}
\begin{figure*}\ContinuedFloat
\centering
     \begin{subfigure}[b]{\linewidth}
         \centering
         \includegraphics[width=\linewidth] {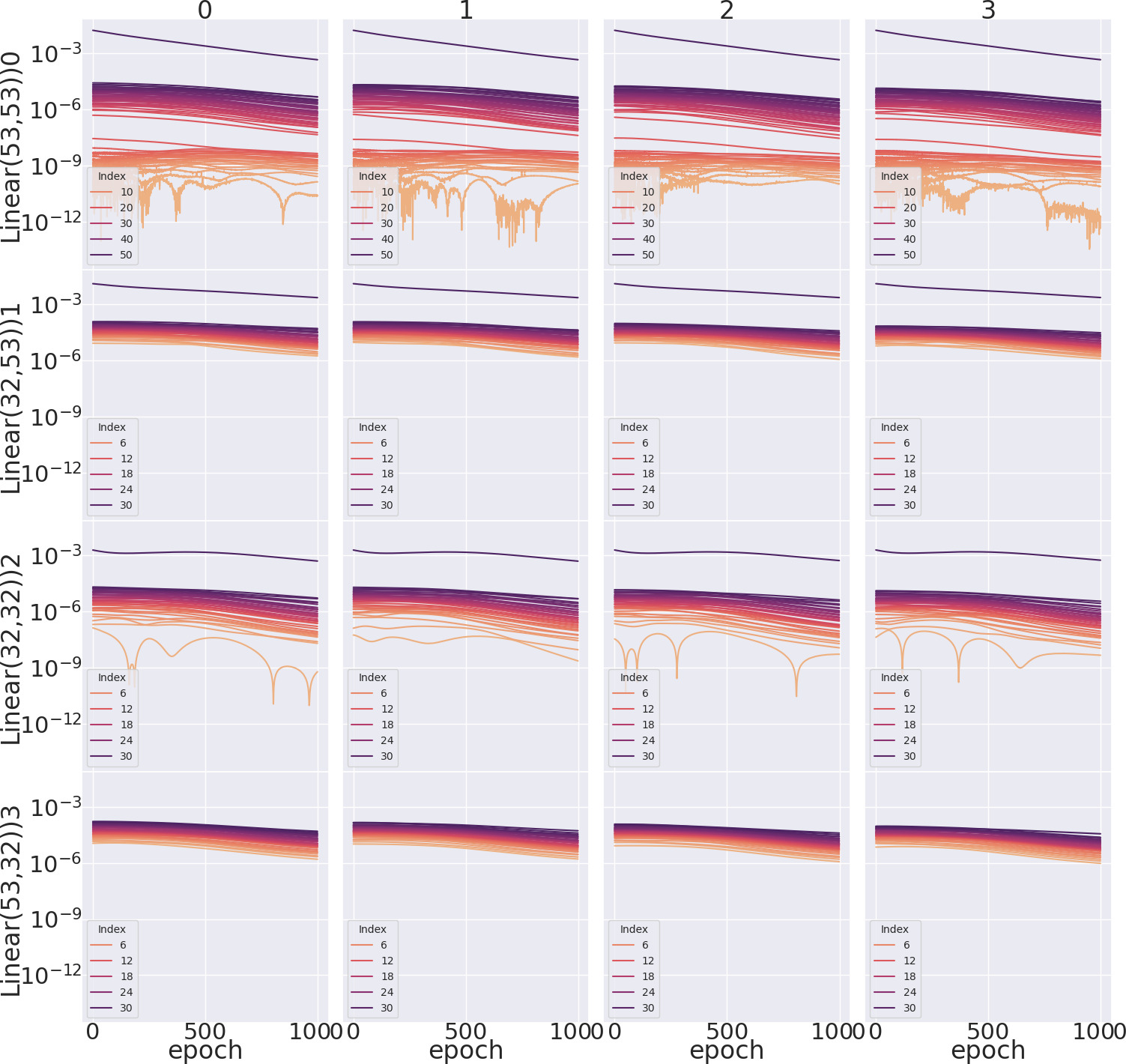}
         \caption{RNN Autoencoder with hidden dim 32 and Sigmoid activations}
         \label{ch6-fig:rnn_sin_c}
     \end{subfigure}
\end{figure*}
\begin{figure*}\ContinuedFloat
     \begin{subfigure}[b]{\linewidth}
         \centering
         \includegraphics[width=\linewidth] {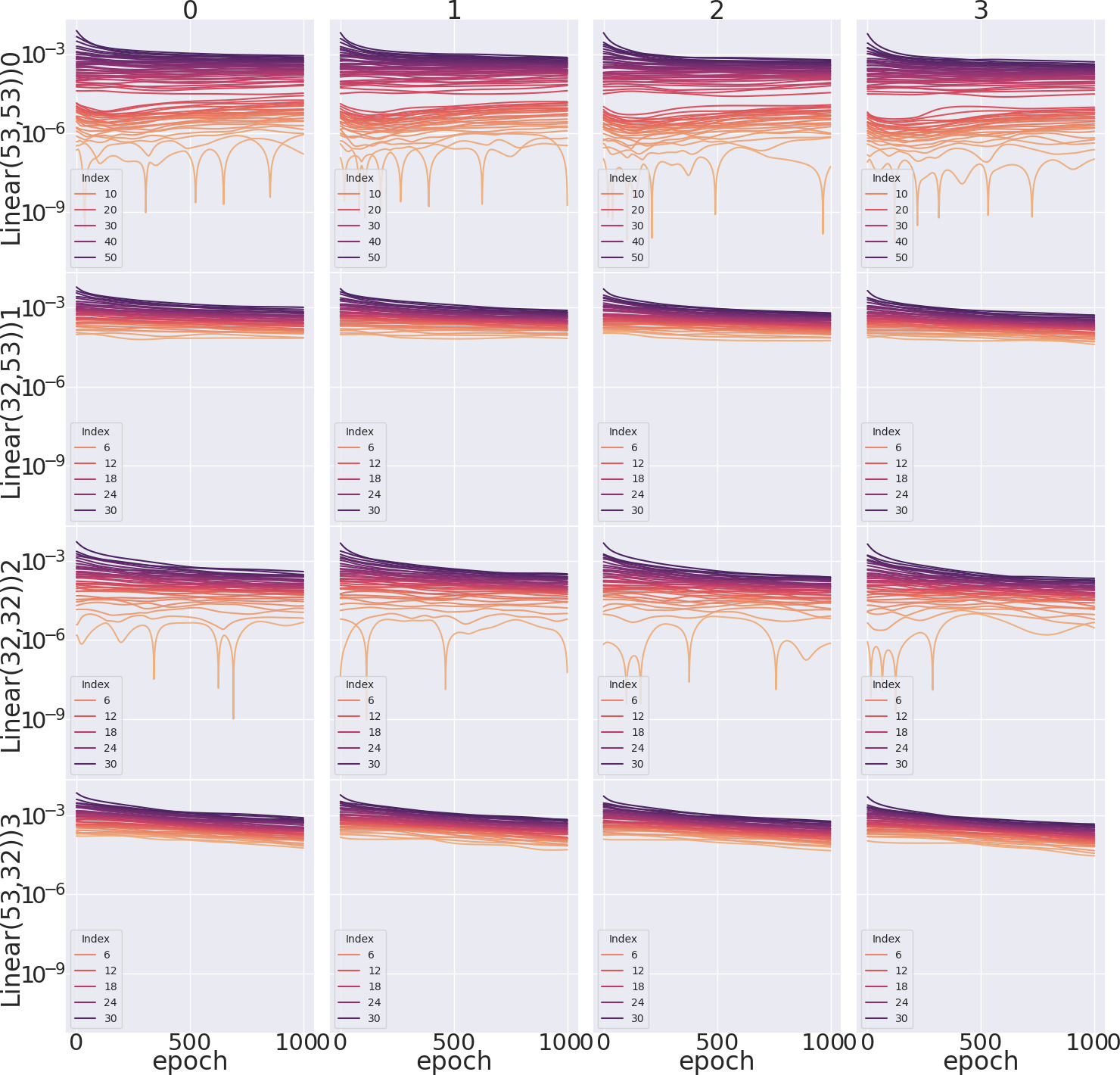}
         \caption{RNN Autoencoder with hidden dim 32 and Tanh activations}
         \label{ch6-fig:rnn_sin_d}
     \end{subfigure}
\end{figure*}
\begin{figure*}\ContinuedFloat
\centering
     \begin{subfigure}[b]{\linewidth}
         \centering
         \includegraphics[width=\linewidth] {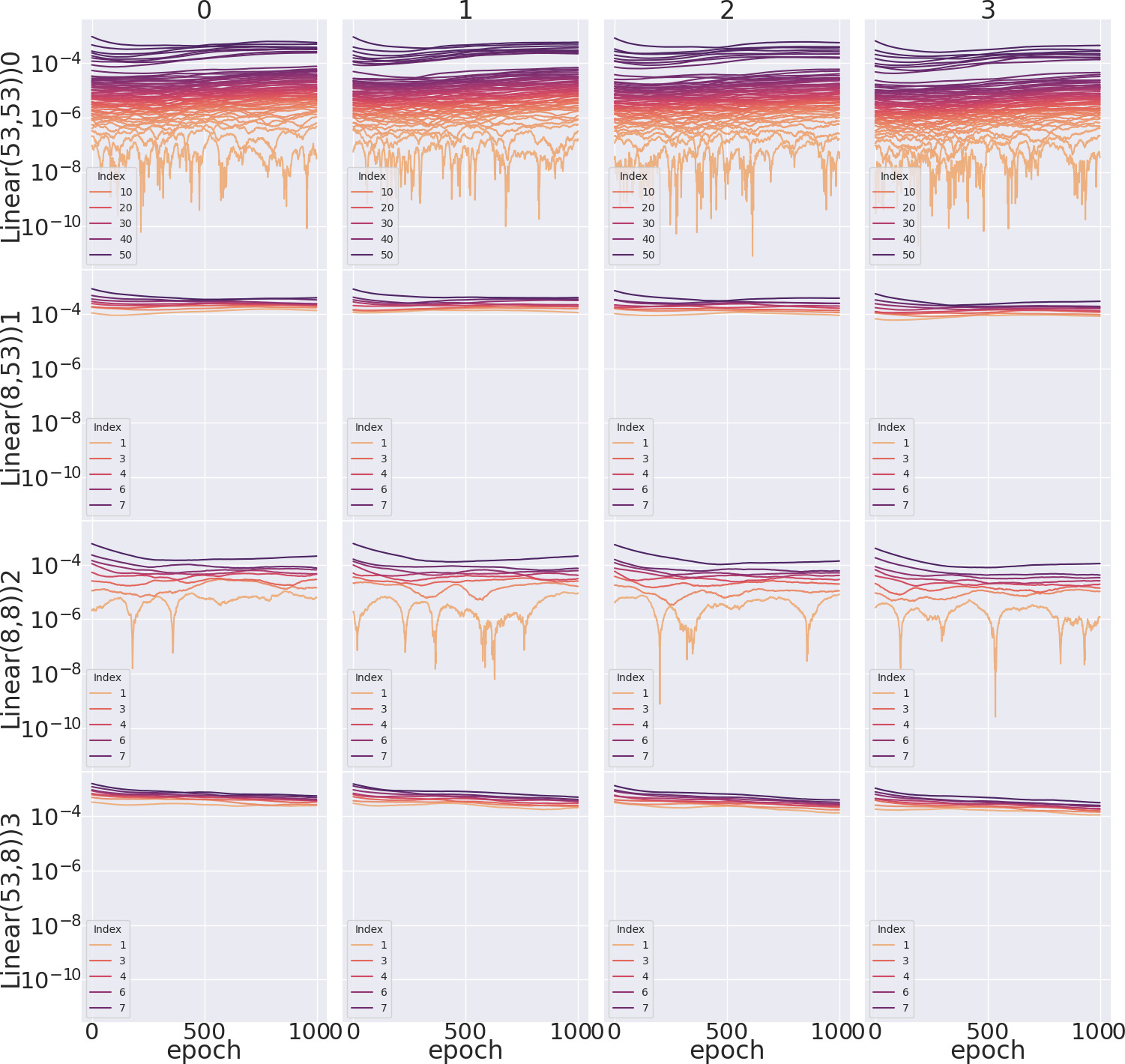}
         \caption{RNN Autoencoder with hidden dim 8 and ReLU activations}
         \label{ch6-fig:rnn_sin_e}
     \end{subfigure}
\end{figure*}
\begin{figure*}\ContinuedFloat
          \begin{subfigure}[b]{\linewidth}
         \centering
         \includegraphics[width=\linewidth] {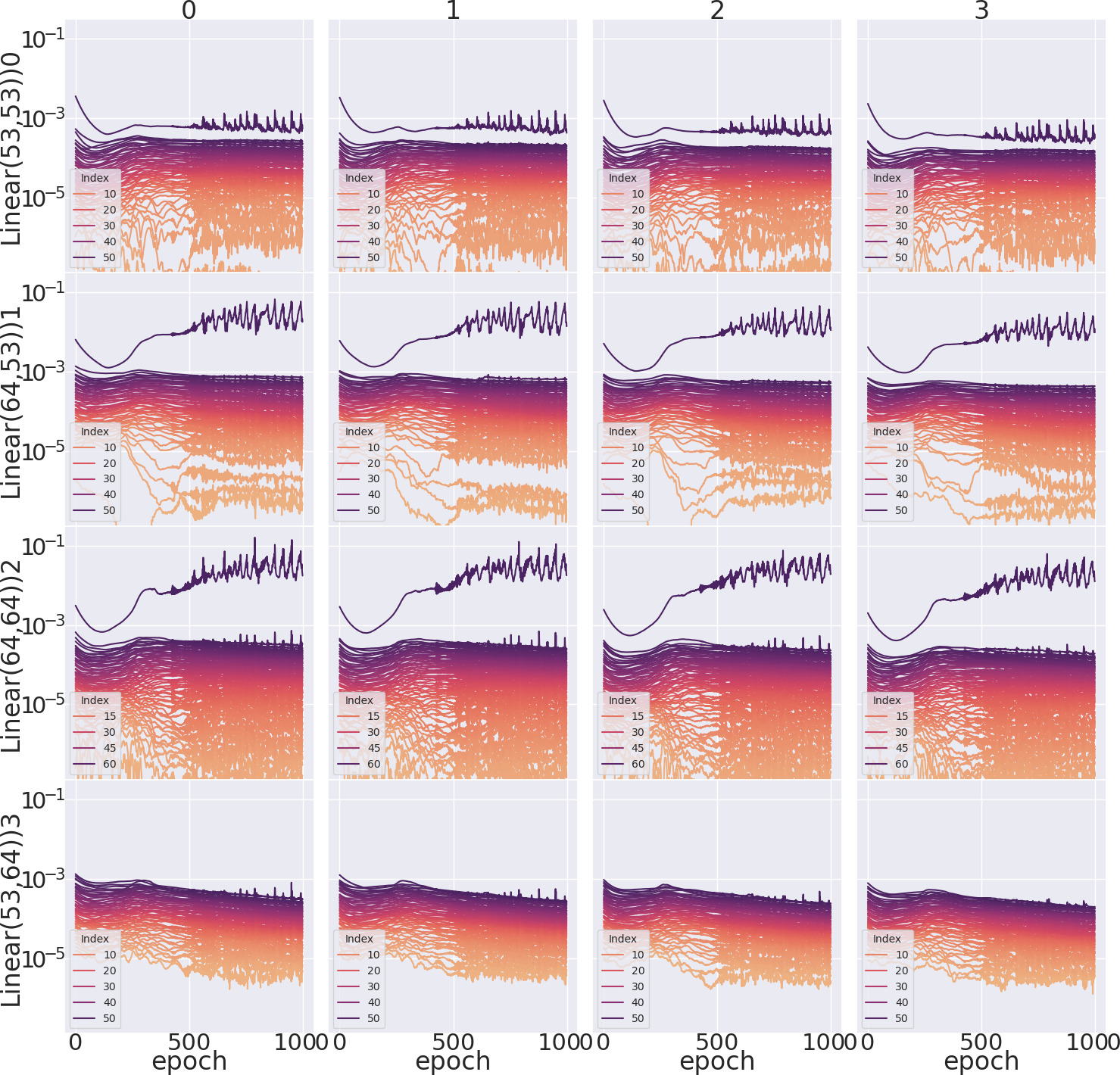}
         \caption{RNN Autoencoder with hidden dim 64 and ReLU activations}
         \label{ch6-fig:rnn_sin_f}
     \end{subfigure}
\end{figure*}
\begin{figure*}\ContinuedFloat
     \begin{subfigure}[b]{1\linewidth}
         \centering
         \includegraphics[width=\linewidth] {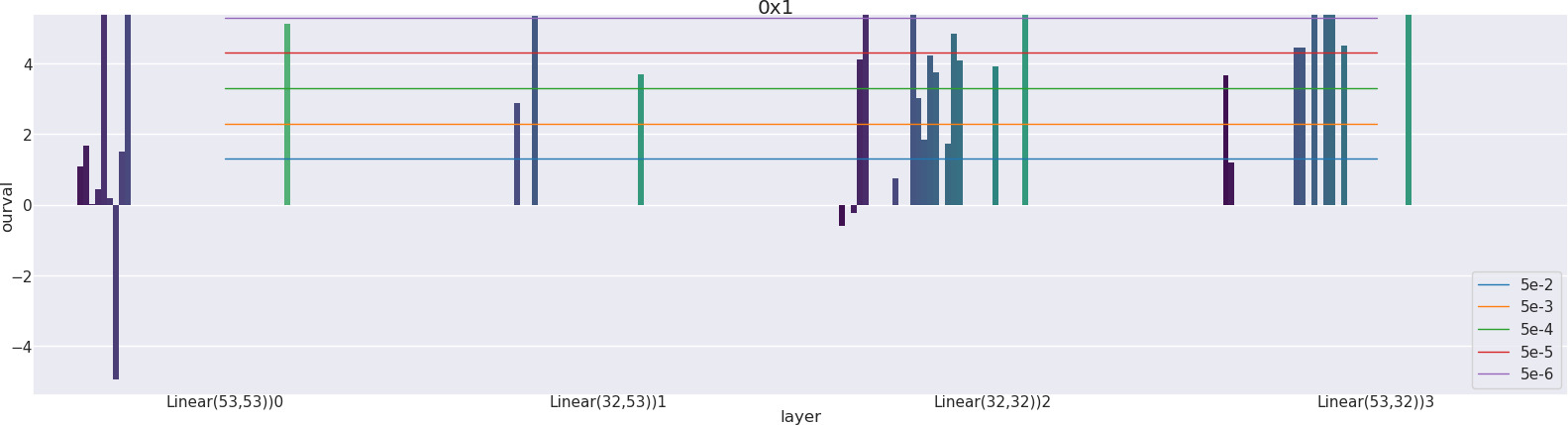}
         \caption{Significant spectral differences between groups 0 and 3: RNN Autoencoder with hidden dim 32 and ReLU activations}
         \label{ch6-fig:rnn_sin_g}
     \end{subfigure}
     \begin{subfigure}[b]{1\linewidth}
         \centering
         \includegraphics[width=\linewidth] {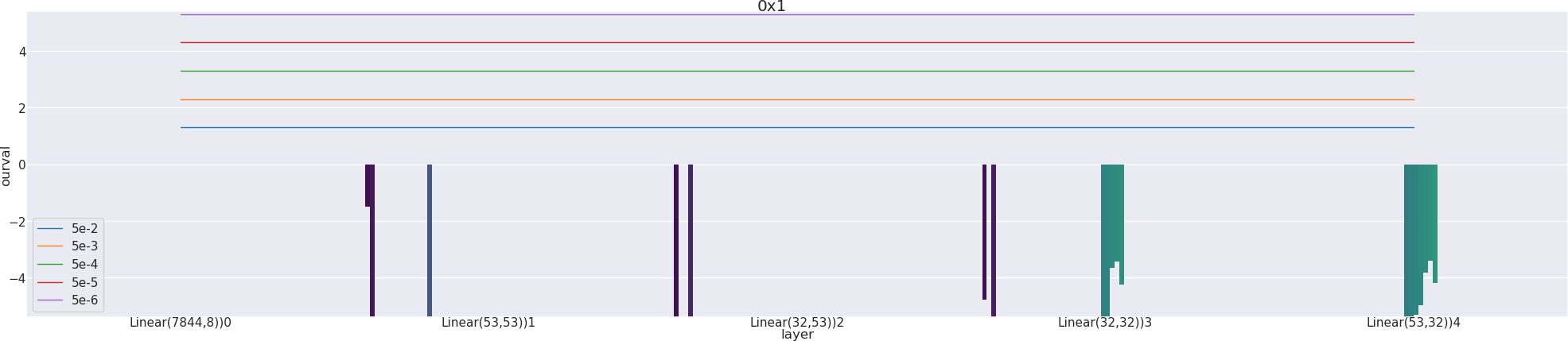}
         \caption{Significant spectral differences between groups 0 and 3: RNN Classifier with hidden dim 32 and ReLU activations}
         \label{ch6-fig:rnn_sin_h}
     \end{subfigure}
     \caption{Differences in Auto-Differentiation Spectra Dynamics for the first 4 classes in the Sinusoid data set, trained with various architectures and tasks with an RNN}
     \label{ch6-fig:rnn_sin}
\end{figure*}

\begin{figure*}
     \caption{Differences in Auto-Differentiation Spectra Dynamics on the FSL data set, trained with various architectures and tasks with a Multi-Layer Perceptron.}
     \centering
         \begin{subfigure}[b]{\linewidth}
         \centering
         \includegraphics[width=\linewidth] {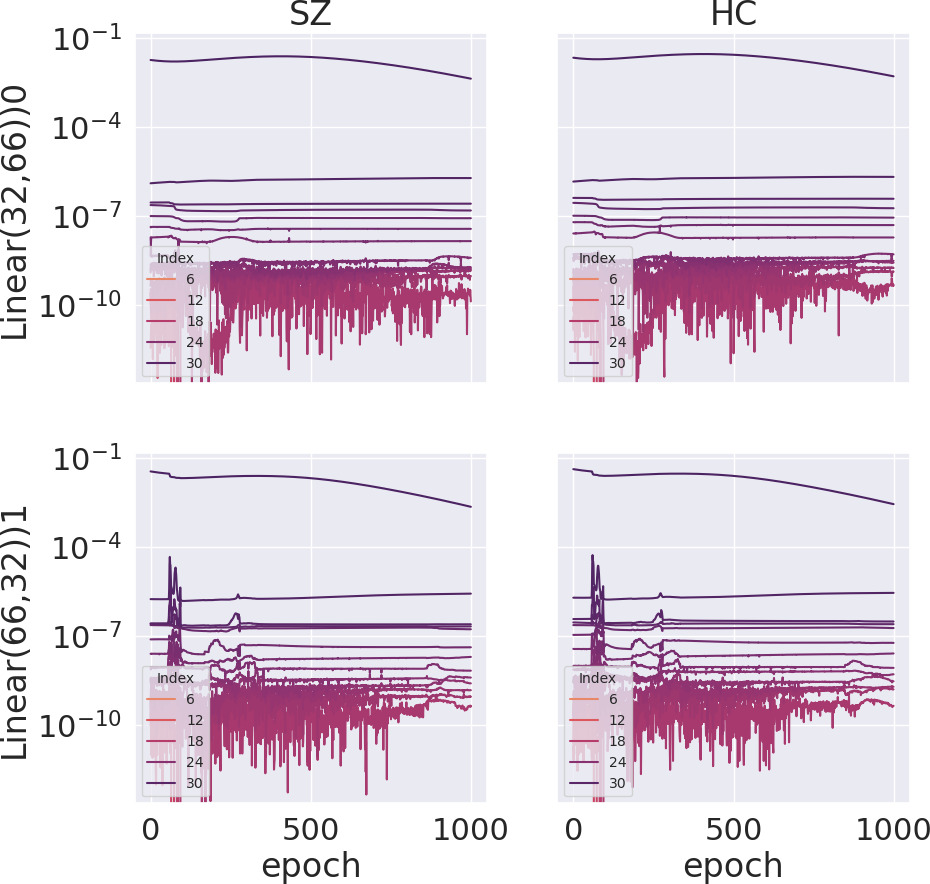}
         \caption{MLP Autoencoder with hidden dim 32 and ReLU activations}
         \label{ch6-fig:mlp_fsl_a}
     \end{subfigure}
\end{figure*}
\begin{figure*}\ContinuedFloat
              \begin{subfigure}[b]{\linewidth}
         \centering
         \includegraphics[width=\linewidth] {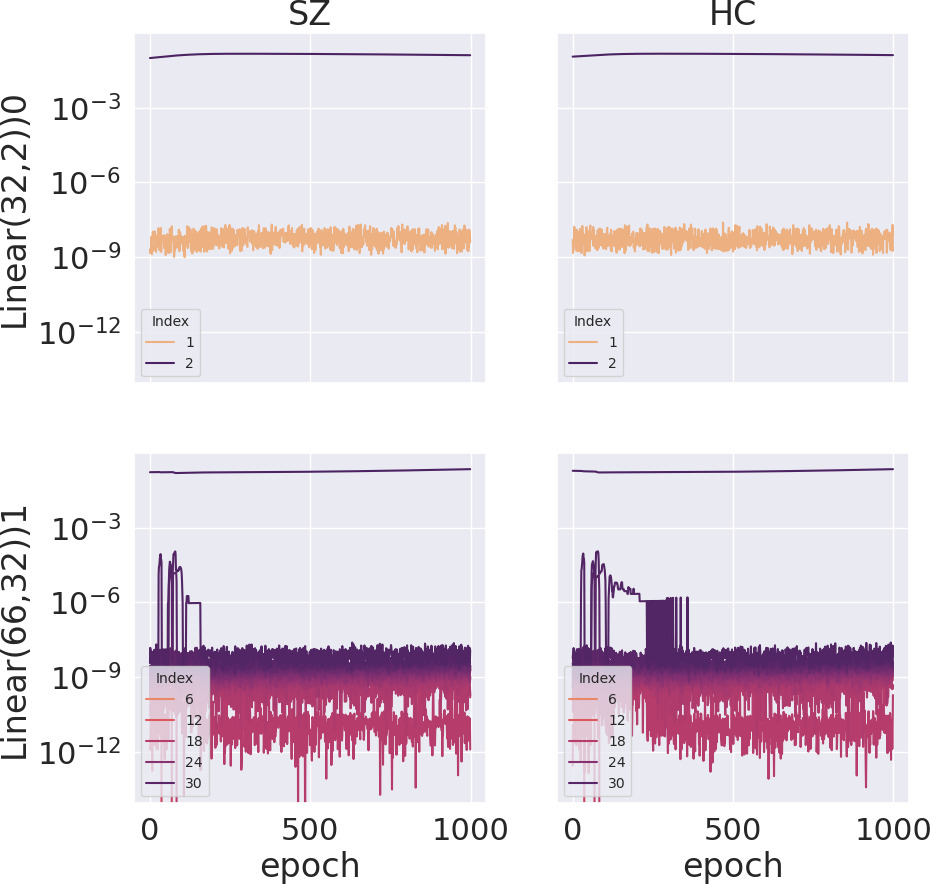}
         \caption{MLP Classifier with hidden dim 32 and ReLU activations}
         \label{ch6-fig:mlp_fsl_b}
     \end{subfigure}
\end{figure*}
\begin{figure*}\ContinuedFloat
\centering
     \begin{subfigure}[b]{\linewidth}
         \centering
         \includegraphics[width=\linewidth] {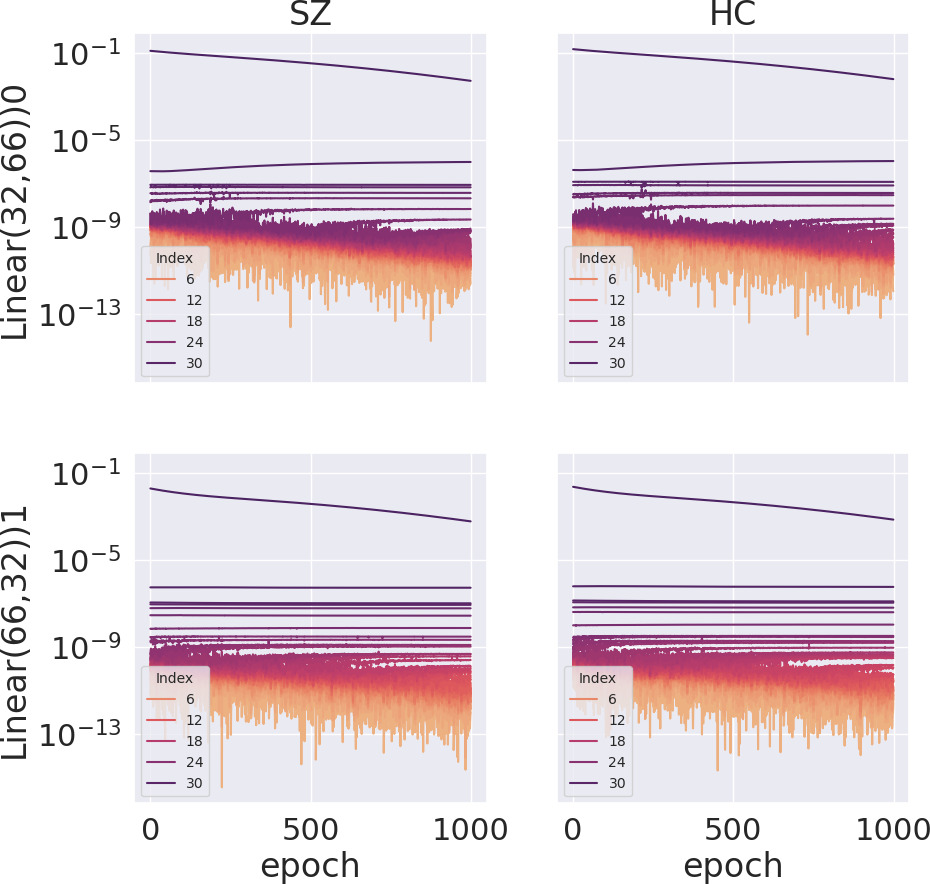}
         \caption{MLP Autoencoder with hidden dim 32 and Sigmoid activations}
         \label{ch6-fig:mlp_fsl_c}
     \end{subfigure}
\end{figure*}
\begin{figure*}\ContinuedFloat
     \begin{subfigure}[b]{\linewidth}
         \centering
         \includegraphics[width=\linewidth] {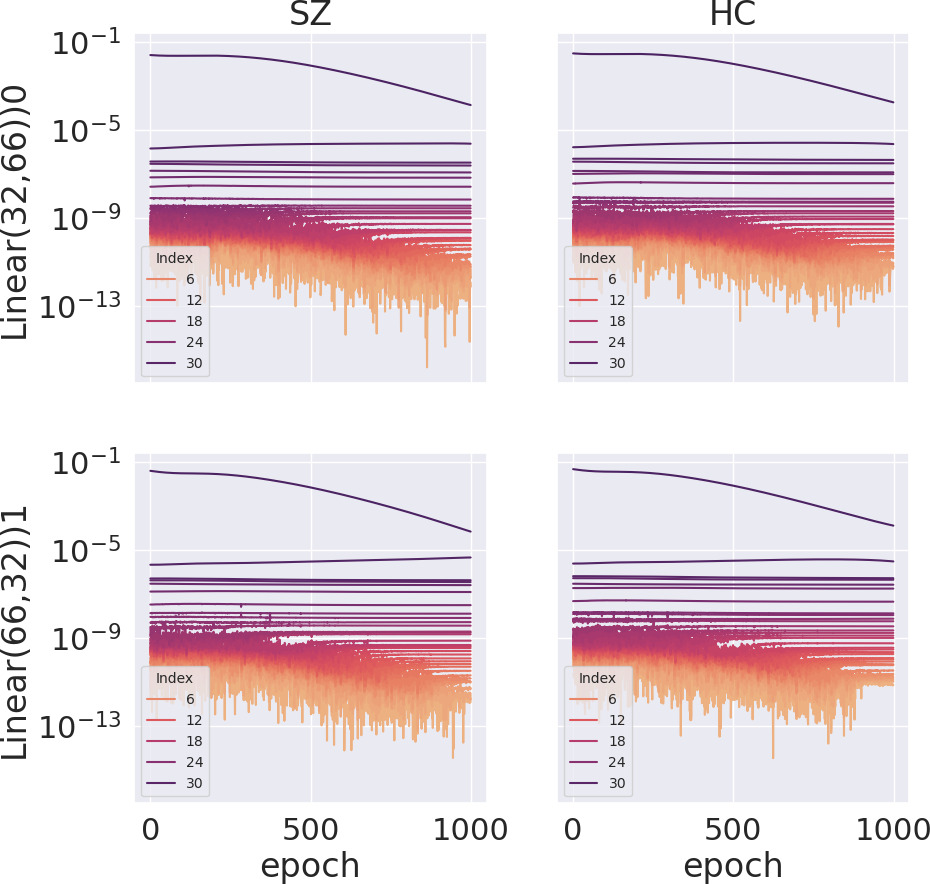}
         \caption{MLP Autoencoder with hidden dim 32 and Tanh activations}
         \label{ch6-fig:mlp_fsl_d}
     \end{subfigure}
\end{figure*}
\begin{figure*}\ContinuedFloat
\centering
     \begin{subfigure}[b]{\linewidth}
         \centering
         \includegraphics[width=\linewidth] {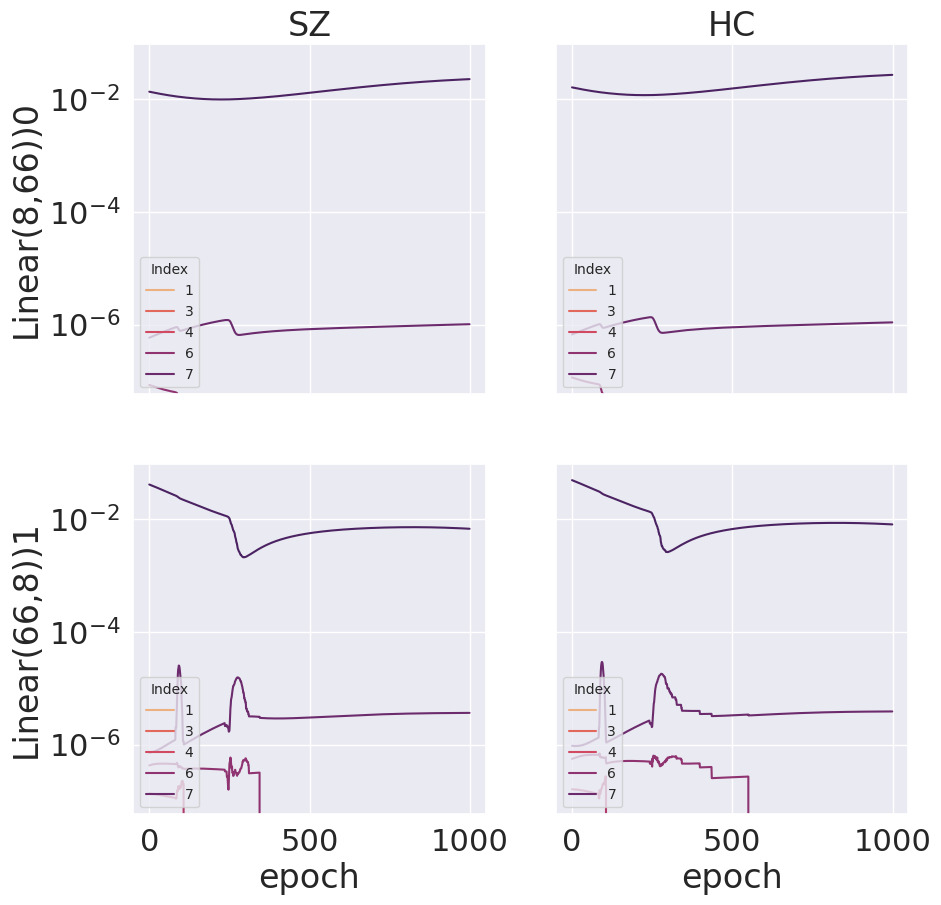}
         \caption{MLP Autoencoder with hidden dim 8 and ReLU activations}
         \label{ch6-fig:mlp_fsl_e}
     \end{subfigure}
\end{figure*}
\begin{figure*}\ContinuedFloat
          \begin{subfigure}[b]{\linewidth}
         \centering
         \includegraphics[width=\linewidth] {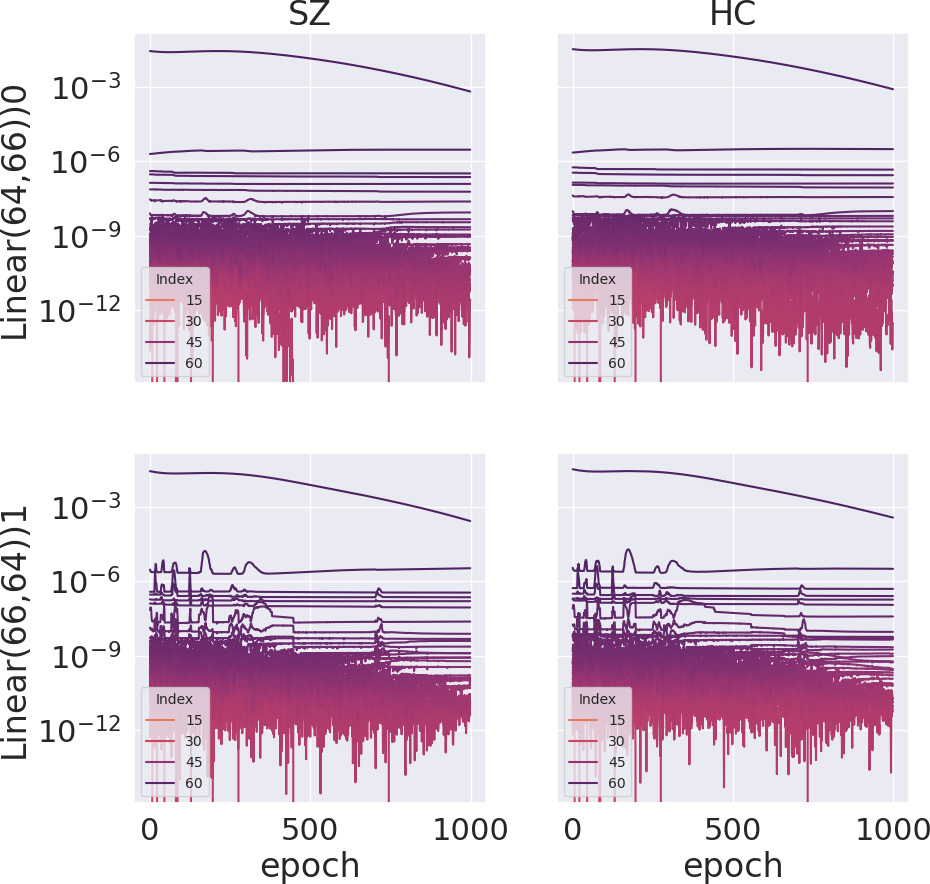}
         \caption{MLP Autoencoder with hidden dim 64 and ReLU activations}
         \label{ch6-fig:mlp_fsl_f}
     \end{subfigure}
\end{figure*}
\begin{figure*}\ContinuedFloat
\centering
     \begin{subfigure}[b]{1\linewidth}
         \centering
         \includegraphics[width=\linewidth] {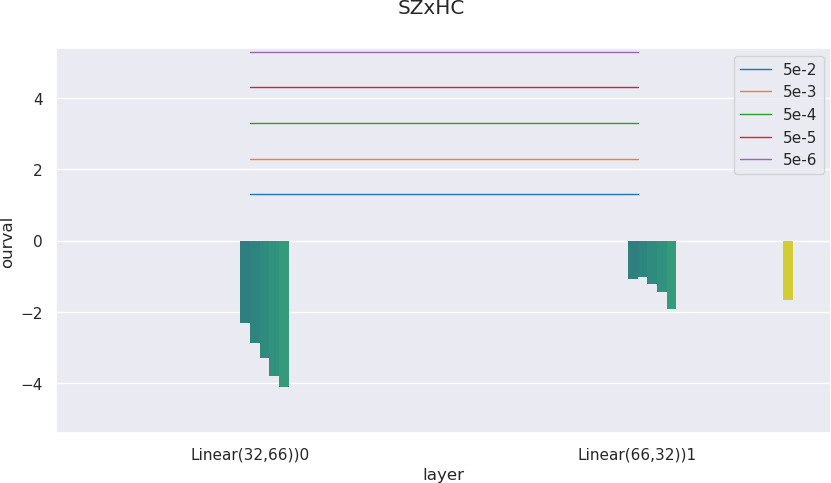}
         \caption{Significant spectral differences between groups SZ and HC: MLP Autoencoder with hidden dim 32 and ReLU activations}
         \label{ch6-fig:mlp_fsl_g}
     \end{subfigure}
     \begin{subfigure}[b]{1\linewidth}
         \centering
         \includegraphics[width=\linewidth] {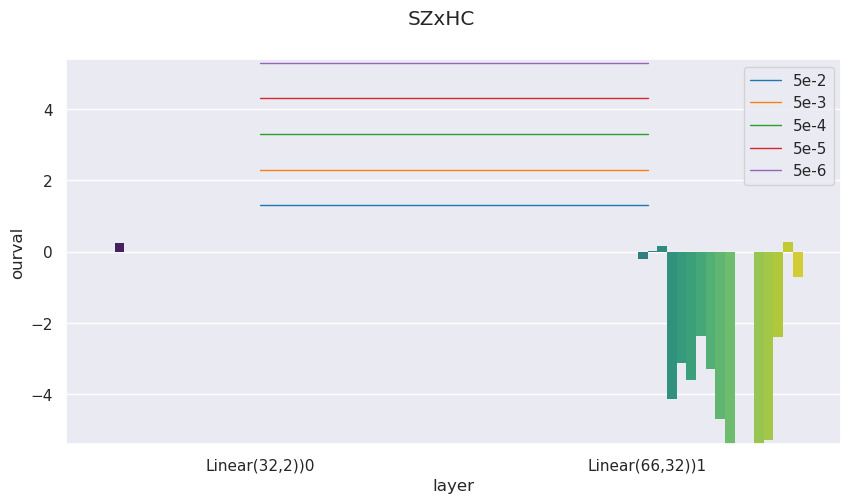}
         \caption{Significant spectral differences between groups SZ and HC: MLP Classifier with hidden dim 32 and ReLU activations}
         \label{ch6-fig:mlp_fsl_h}
     \end{subfigure}
     \caption{Differences in Auto-Differentiation Spectra Dynamics on the FSL data set, trained with various architectures and tasks with a Multi-Layer Perceptron.}
     \label{ch6-fig:mlp_fsl}
\end{figure*}

\begin{figure*}
     \caption{Differences in Auto-Differentiation Spectra Dynamics on the COBRE data set, trained with various architectures and tasks with an LSTM.}
     \centering
         \begin{subfigure}[b]{\linewidth}
         \centering
         \includegraphics[width=\linewidth] {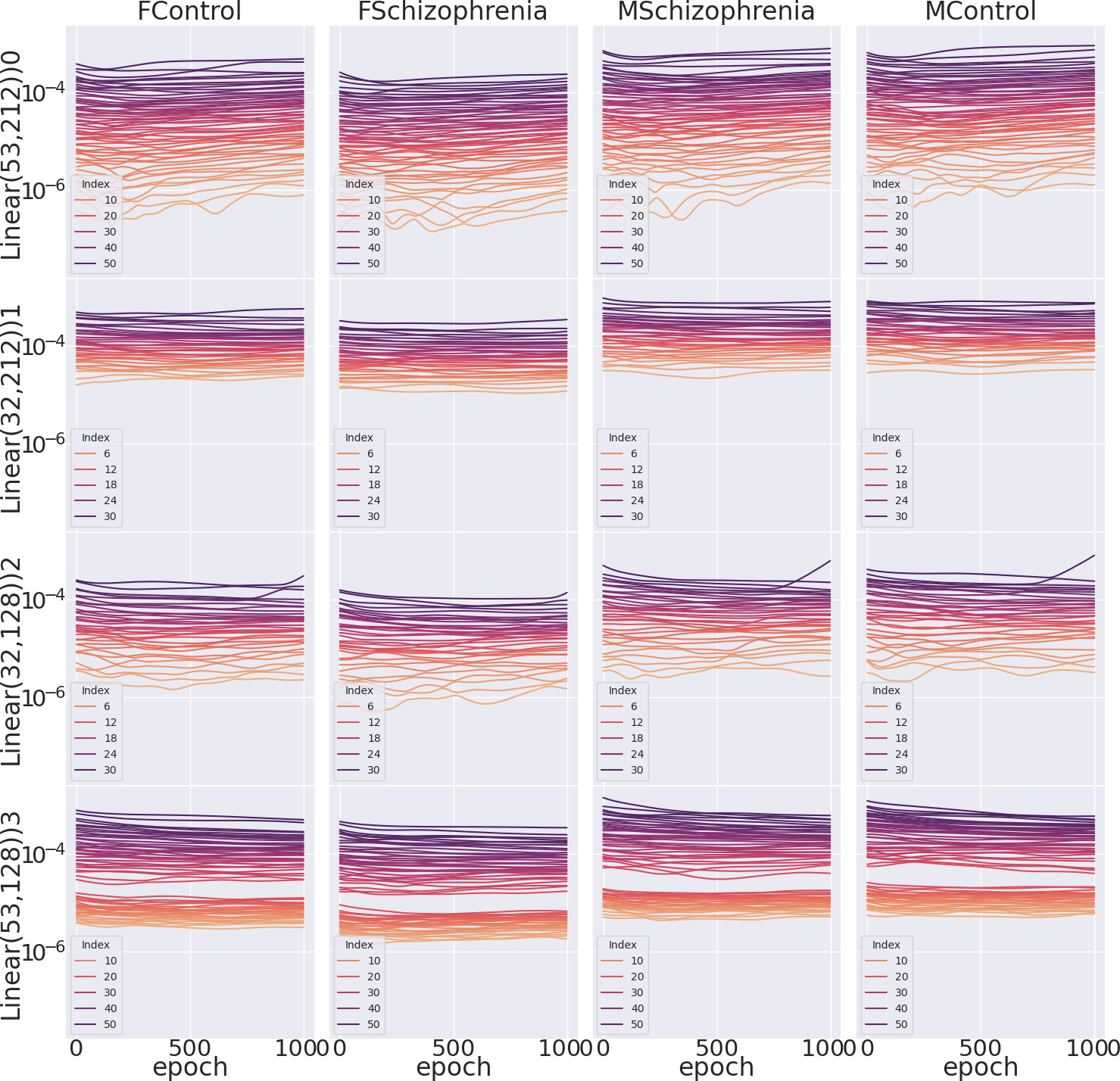}
         \caption{LSTM Autoencoder with hidden dim 32 and ReLU activations}
         \label{ch6-fig:lstm_cobre_a}
     \end{subfigure}
\end{figure*}
\begin{figure*}\ContinuedFloat     
              \begin{subfigure}[b]{\linewidth}
         \centering
         \includegraphics[width=\linewidth] {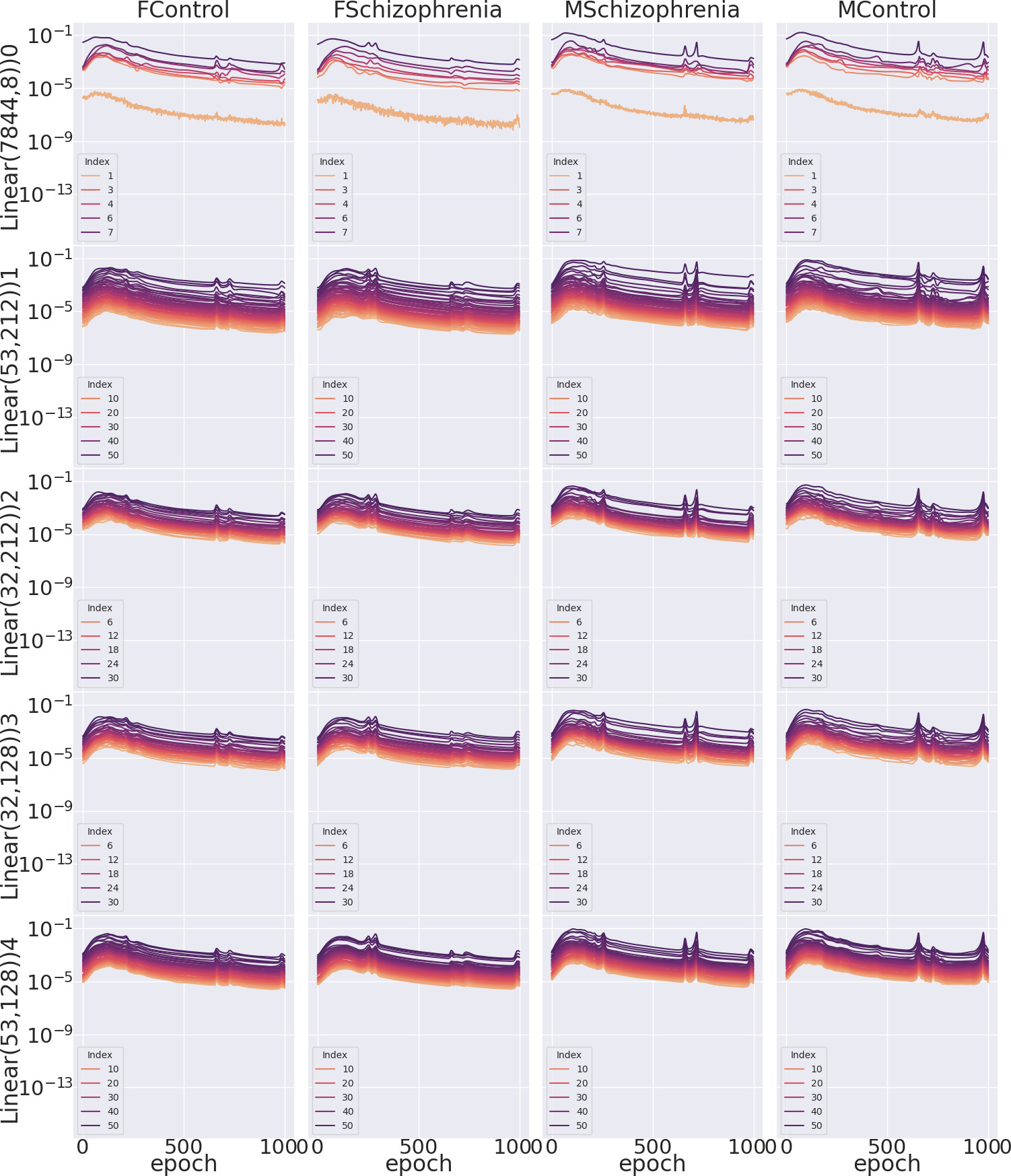}
         \caption{LSTM Classifier with hidden dim 32 and ReLU activations}
         \label{ch6-fig:lstm_cobre_b}
     \end{subfigure}
\end{figure*}
\begin{figure*}\ContinuedFloat
\centering
     \begin{subfigure}[b]{\linewidth}
         \centering
         \includegraphics[width=\linewidth] {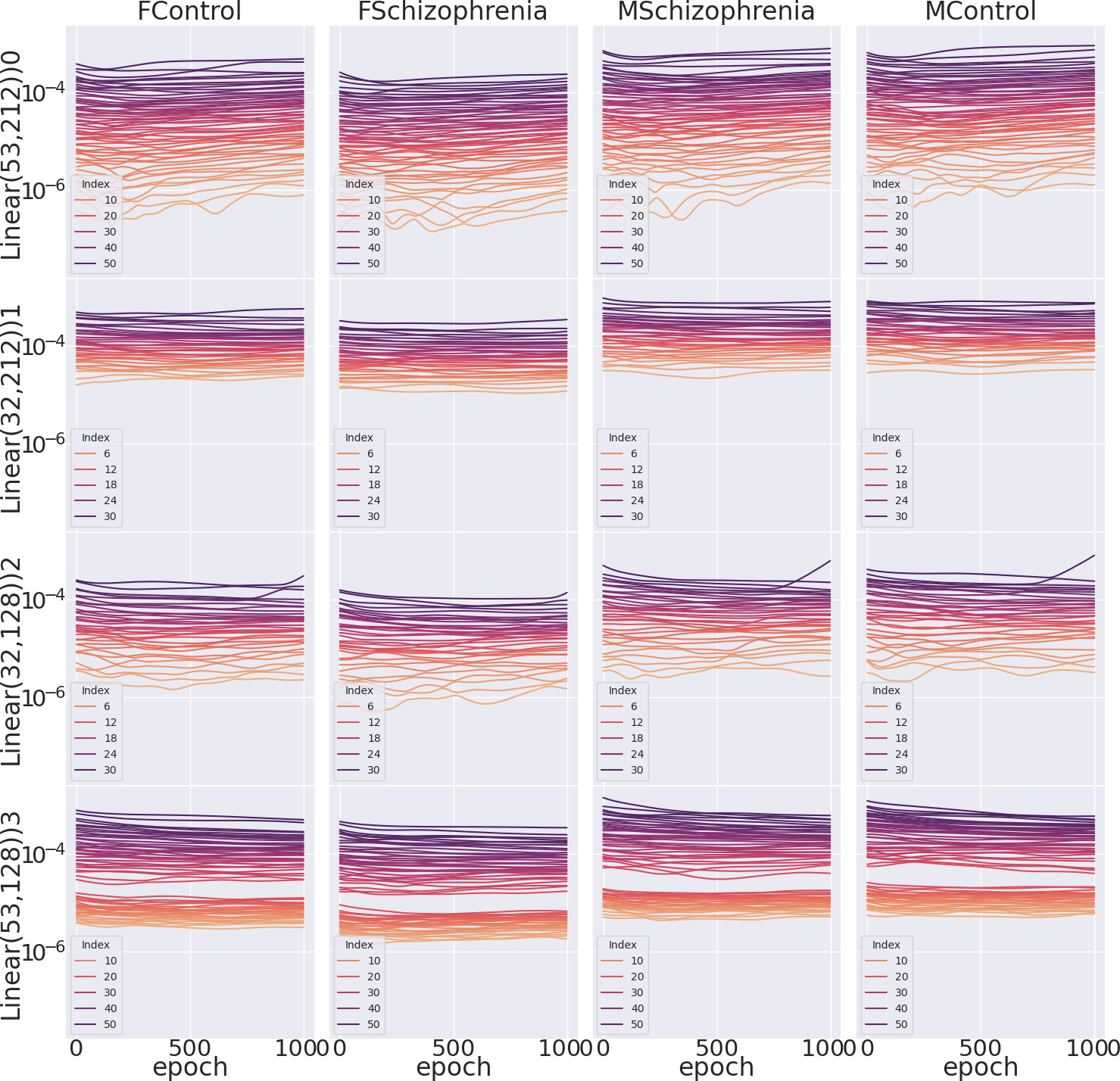}
         \caption{LSTM Autoencoder with hidden dim 32 and Sigmoid activations}
         \label{ch6-fig:lstm_cobre_c}
     \end{subfigure}
\end{figure*}
\begin{figure*}\ContinuedFloat     
     \begin{subfigure}[b]{\linewidth}
         \centering
         \includegraphics[width=\linewidth] {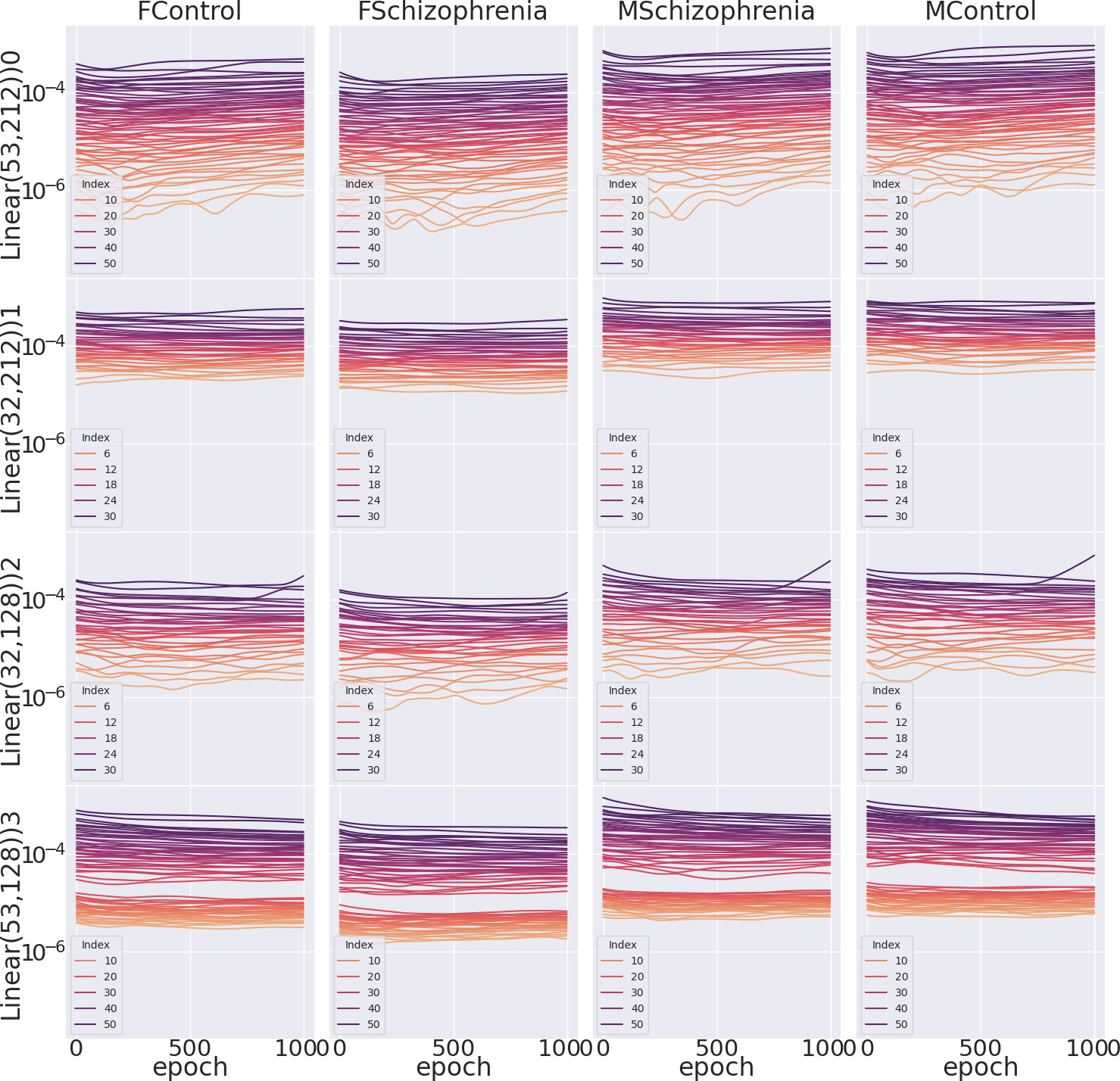}
         \caption{LSTM Autoencoder with hidden dim 32 and Tanh activations}
         \label{ch6-fig:lstm_cobre_d}
     \end{subfigure}
\end{figure*}
\begin{figure*}\ContinuedFloat
\centering
     \begin{subfigure}[b]{\linewidth}
         \centering
         \includegraphics[width=\linewidth] {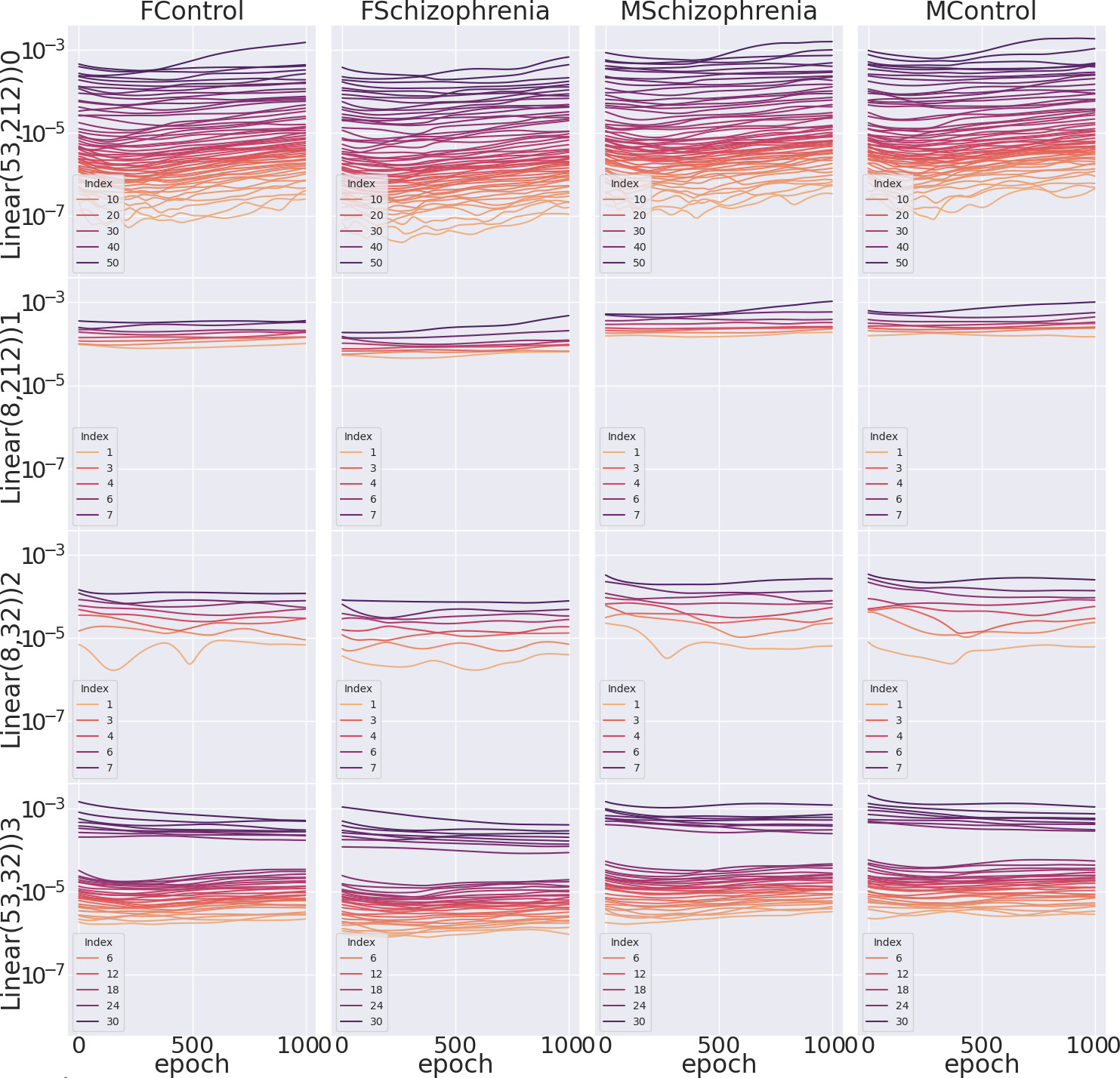}
         \caption{LSTM Autoencoder with hidden dim 8 and ReLU activations}
         \label{ch6-fig:lstm_cobre_e}
     \end{subfigure}
\end{figure*}
\begin{figure*}\ContinuedFloat     
          \begin{subfigure}[b]{\linewidth}
         \centering
         \includegraphics[width=\linewidth] {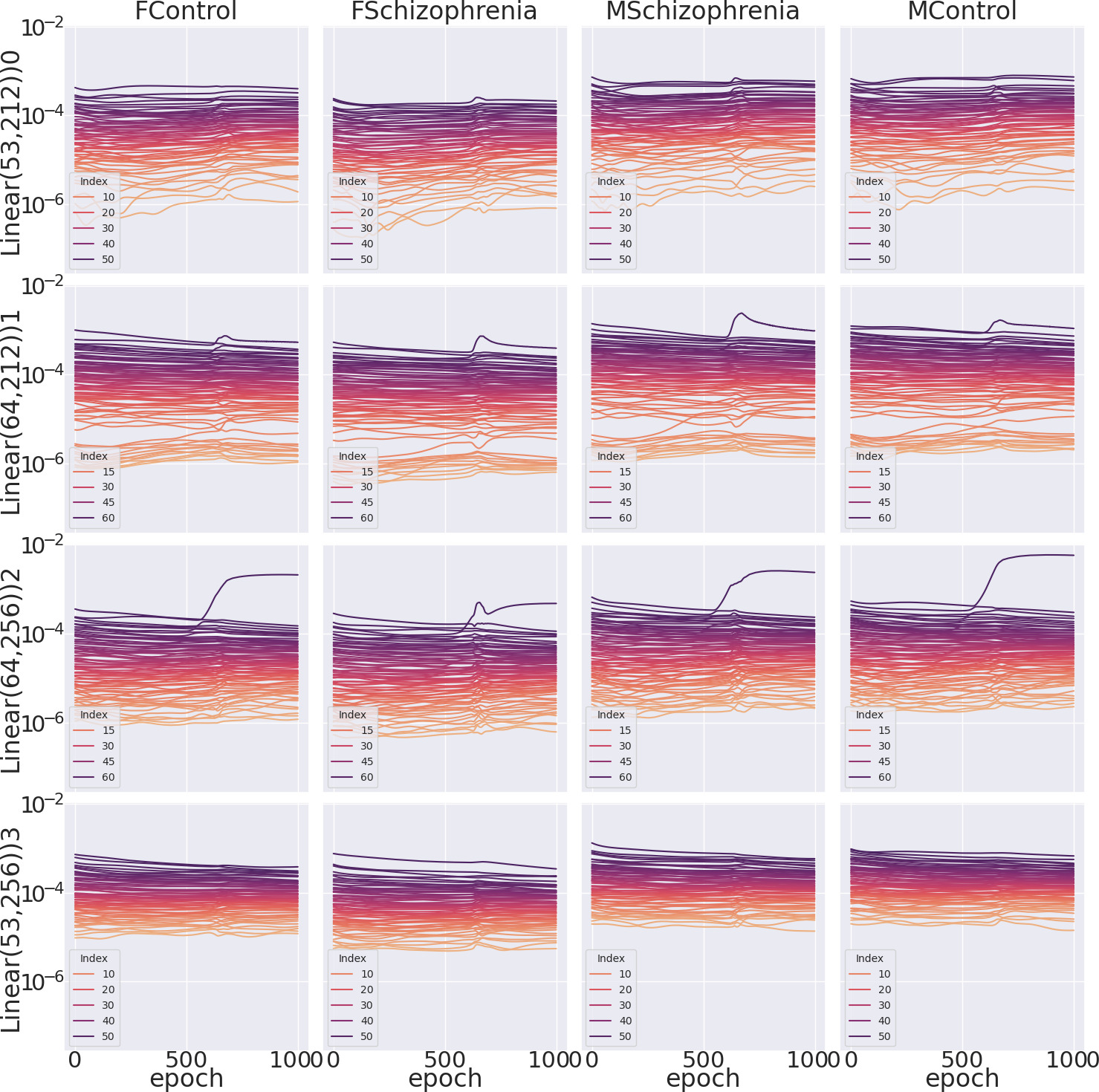}
         \caption{LSTM Autoencoder with hidden dim 64 and ReLU activations}
         \label{ch6-fig:lstm_cobre_f}
     \end{subfigure}
\end{figure*}
\begin{figure*}\ContinuedFloat
\centering
     \begin{subfigure}[b]{1\linewidth}
         \centering
         \includegraphics[width=\linewidth] {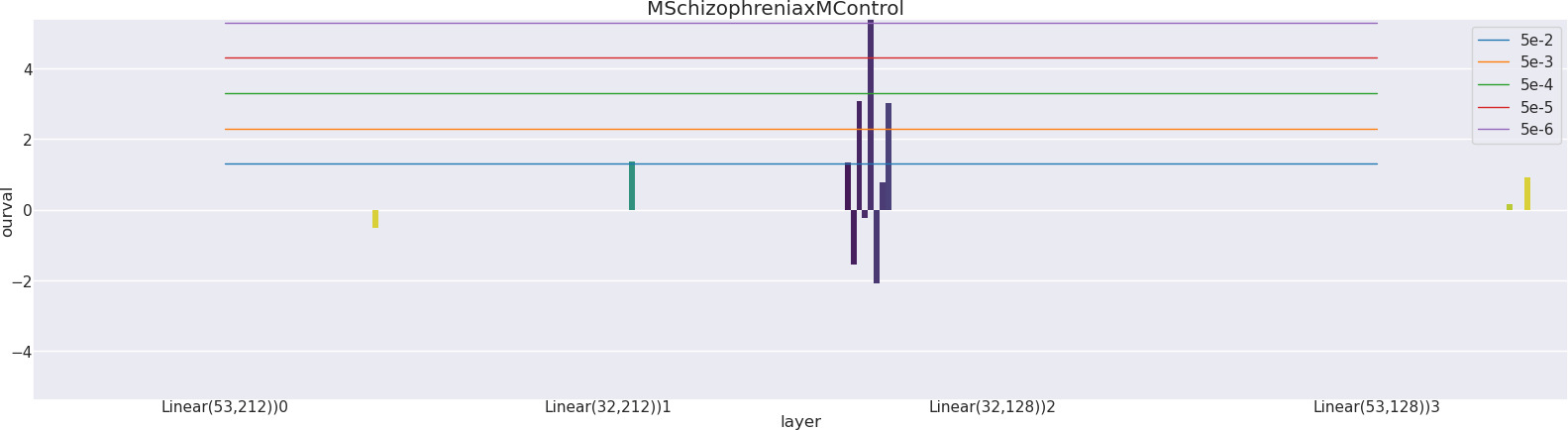}
         \caption{Significant spectral differences between HC, SZ and Sex: LSTM Autoencoder with hidden dim 32 and ReLU activations}
         \label{ch6-fig:lstm_cobre_g}
     \end{subfigure}
     \begin{subfigure}[b]{1\linewidth}
         \centering
         \includegraphics[width=\linewidth] {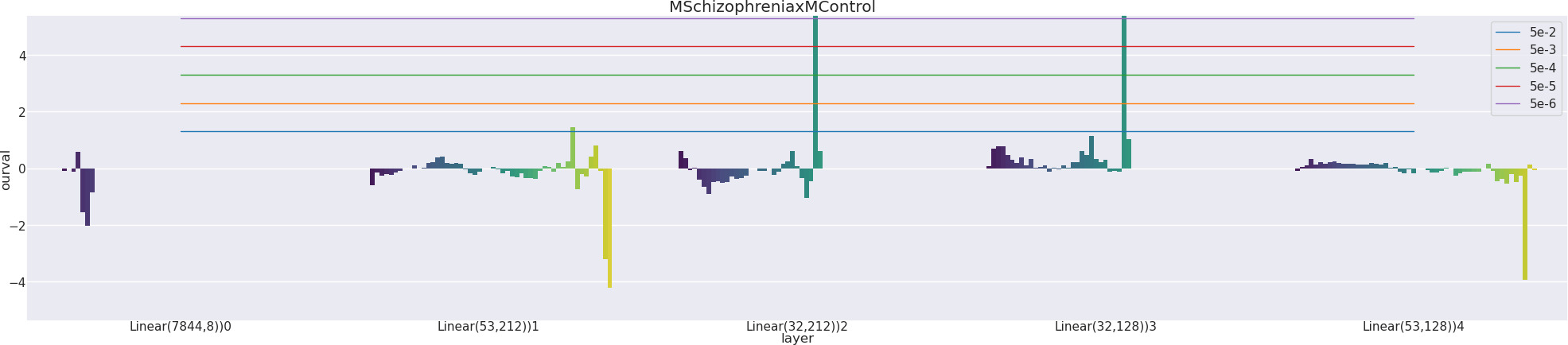}
         \caption{Significant spectral differences between HC, SZ and Sex: LSTM Classifier with hidden dim 32 and ReLU activations}
         \label{ch6-fig:lstm_cobre_h}
     \end{subfigure}
     \caption{Differences in Auto-Differentiation Spectra Dynamics on the COBRE data set, trained with various architectures and tasks with an LSTM.}
     \label{ch6-fig:lstm_cobre}
\end{figure*}

\begin{figure*}
     \caption{Differences in Auto-Differentiation Spectra Dynamics on the COBRE data set, trained with various architectures and tasks with a BERT Transformer.}
     \centering
         \begin{subfigure}[b]{\linewidth}
         \centering
         \includegraphics[width=\linewidth] {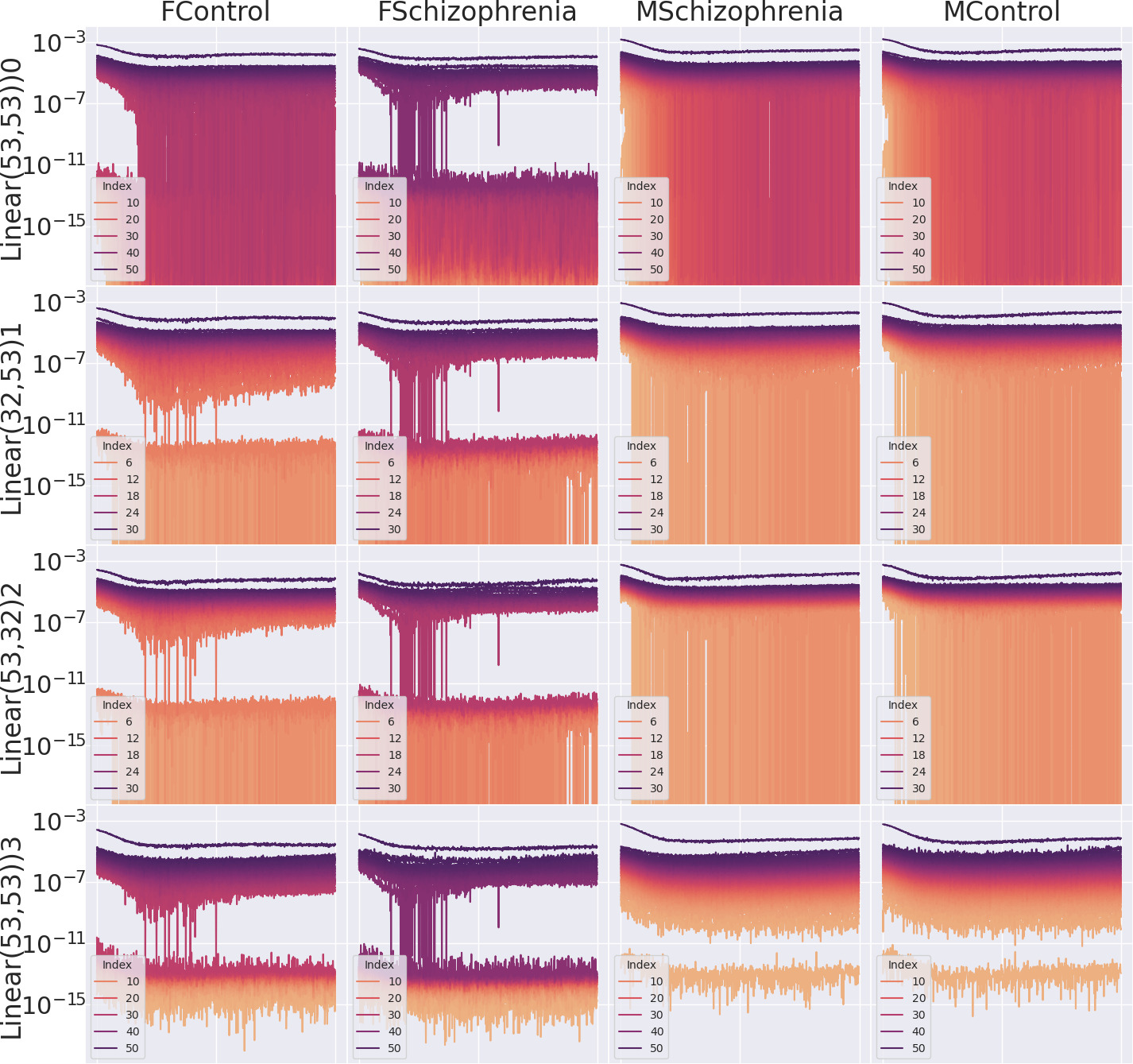}
         \caption{BERT Autoencoder with hidden dim 32 and ReLU activations}
         \label{ch6-fig:bert_cobre_a}
     \end{subfigure}
\end{figure*}
\begin{figure*}\ContinuedFloat
              \begin{subfigure}[b]{\linewidth}
         \centering
         \includegraphics[width=\linewidth] {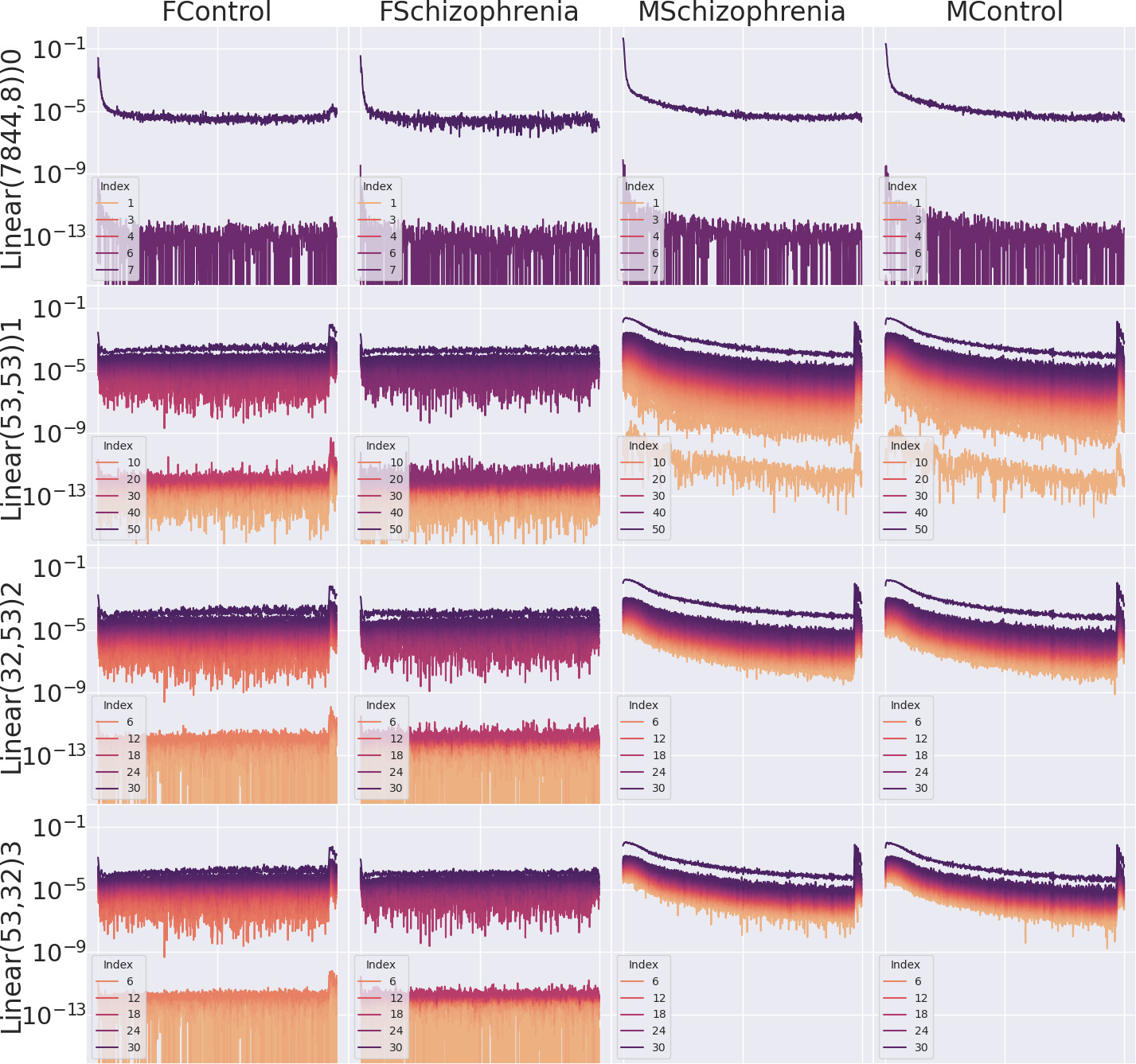}
         \caption{BERT Classifier with hidden dim 32 and ReLU activations}
         \label{ch6-fig:bert_cobre_b}
     \end{subfigure}
\end{figure*}
\begin{figure*}\ContinuedFloat
\centering
     \begin{subfigure}[b]{\linewidth}
         \centering
         \includegraphics[width=\linewidth] {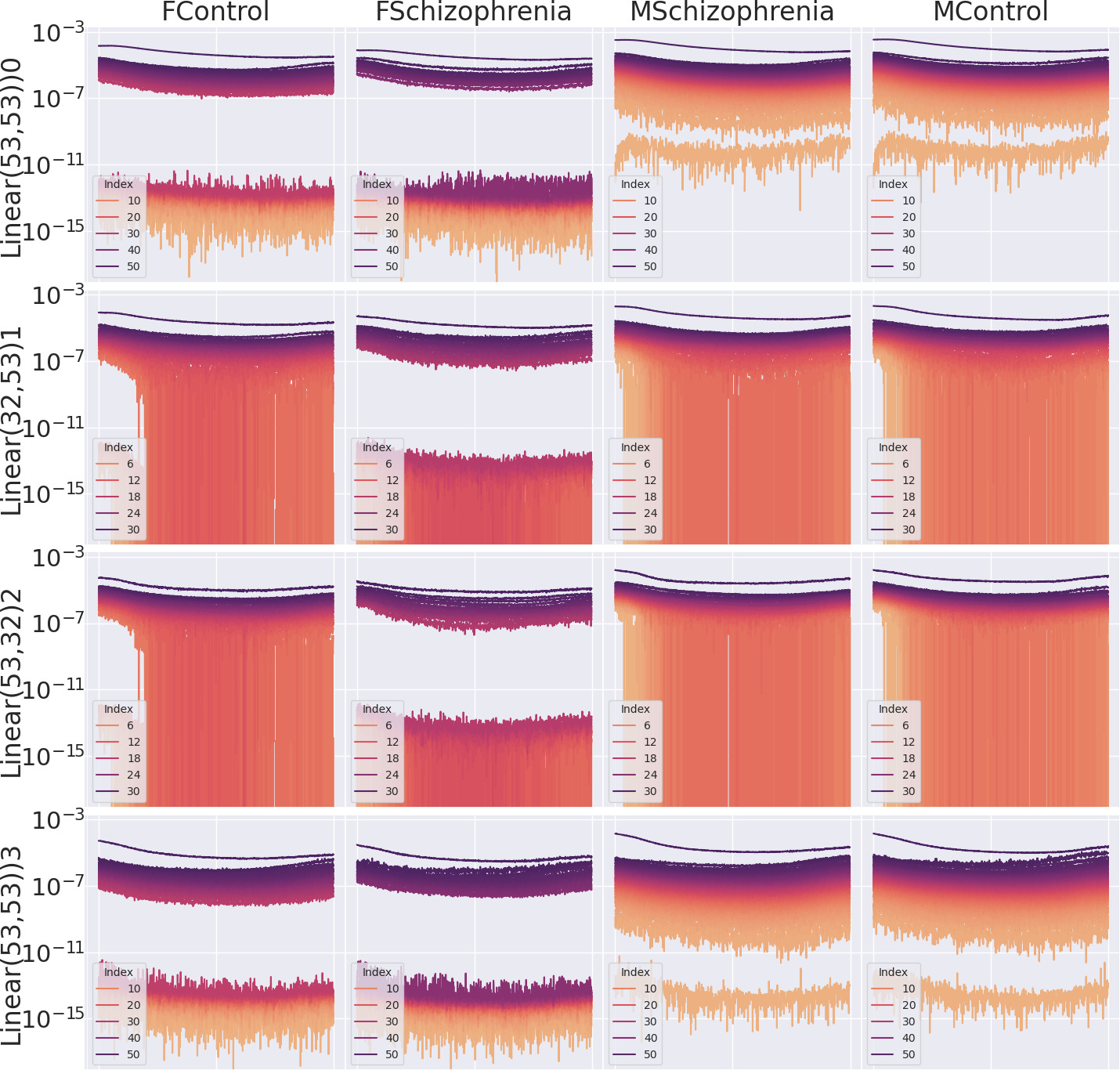}
         \caption{BERT Autoencoder with hidden dim 32 and Sigmoid activations}
         \label{ch6-fig:bert_cobre_c}
     \end{subfigure}
\end{figure*}
\begin{figure*}\ContinuedFloat
     \begin{subfigure}[b]{\linewidth}
         \centering
         \includegraphics[width=\linewidth] {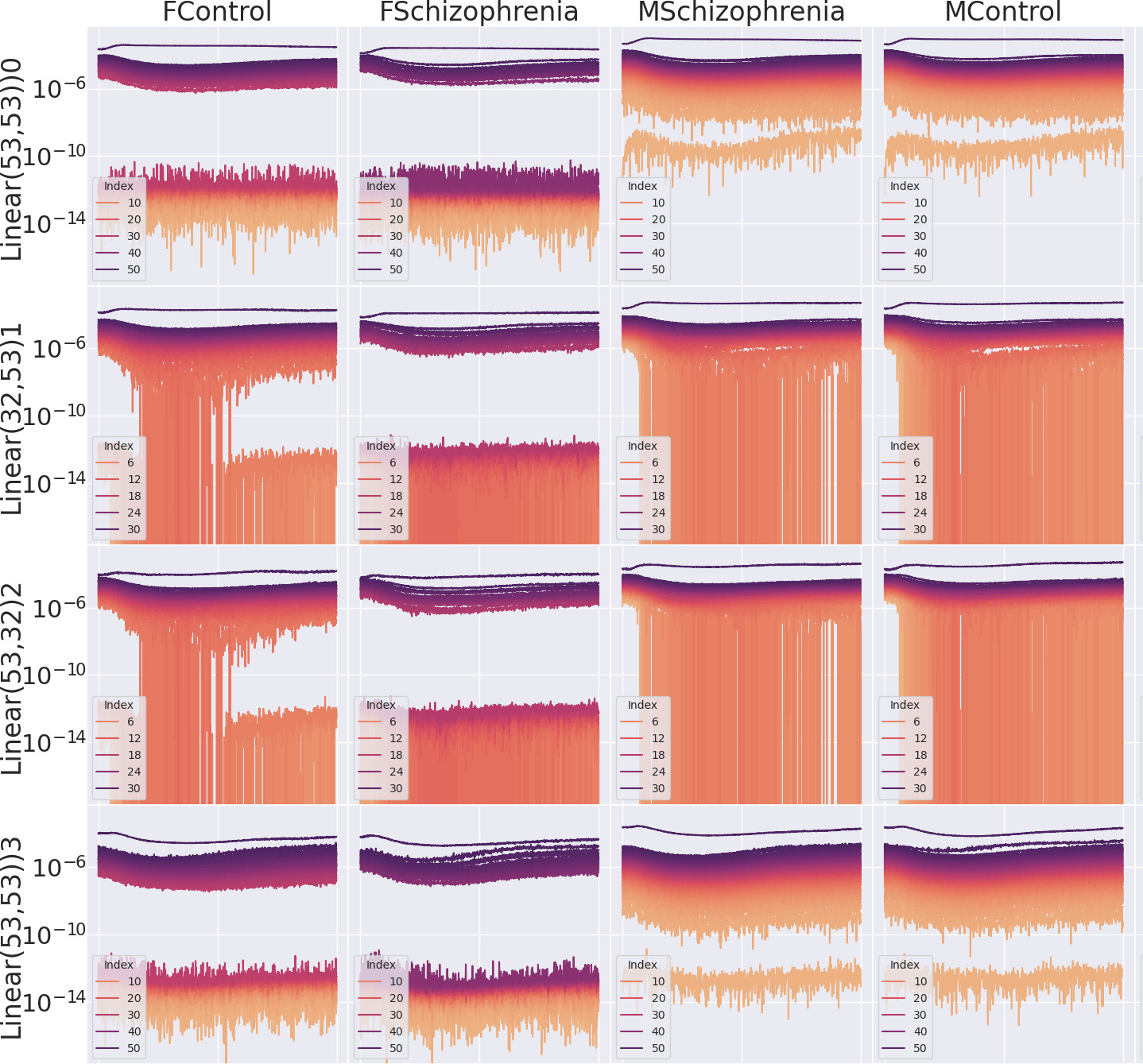}
         \caption{BERT Autoencoder with hidden dim 32 and Tanh activations}
         \label{ch6-fig:bert_cobre_d}
     \end{subfigure}
\end{figure*}
\begin{figure*}\ContinuedFloat
\centering
     \begin{subfigure}[b]{\linewidth}
         \centering
         \includegraphics[width=\linewidth] {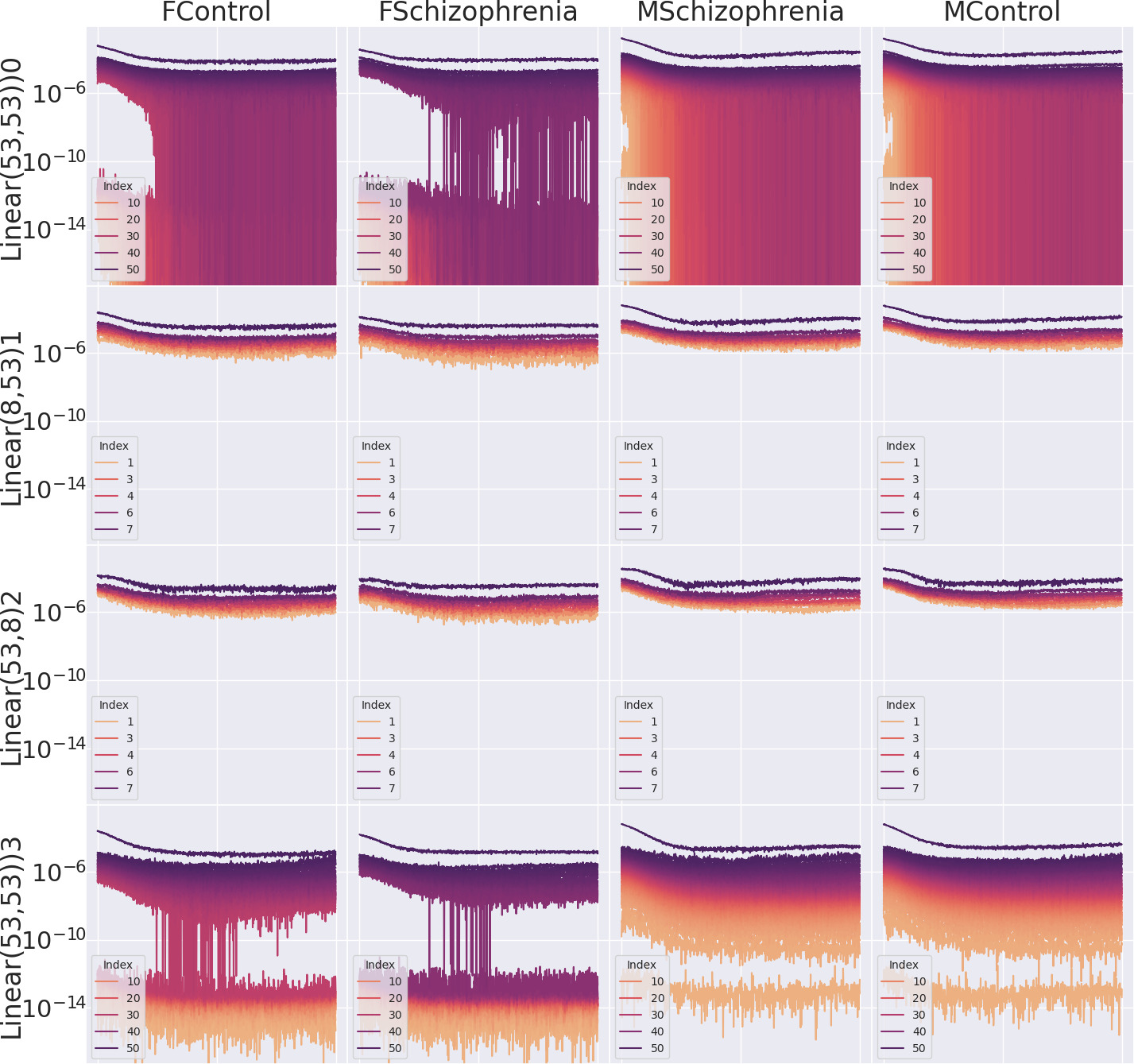}
         \caption{BERT Autoencoder with hidden dim 8 and ReLU activations}
         \label{ch6-fig:bert_cobre_e}
     \end{subfigure}
\end{figure*}
\begin{figure*}\ContinuedFloat     
          \begin{subfigure}[b]{\linewidth}
         \centering
         \includegraphics[width=\linewidth] {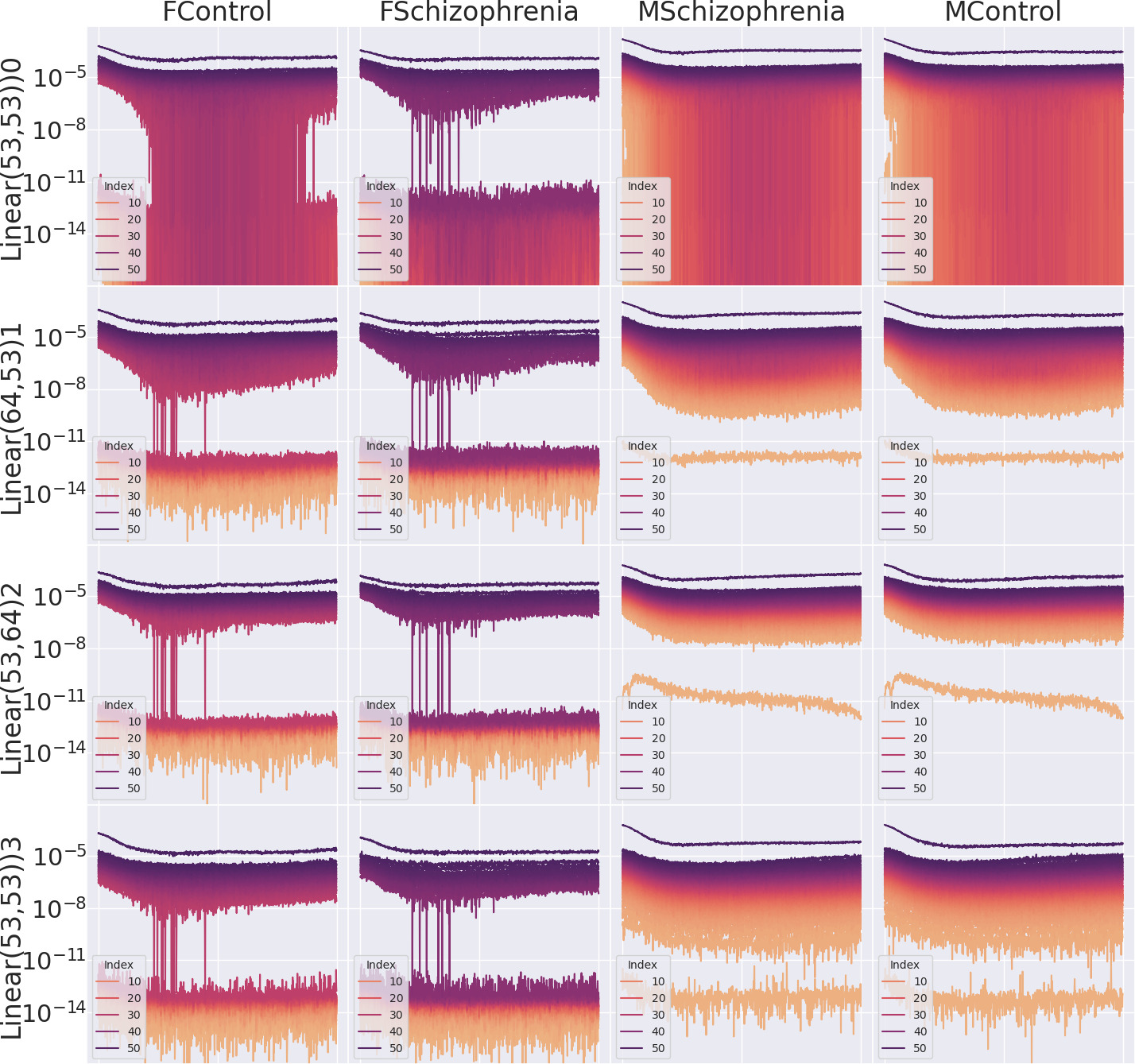}
         \caption{BERT Autoencoder with hidden dim 64 and ReLU activations}
         \label{ch6-fig:bert_cobre_f}
     \end{subfigure}
\end{figure*}
\begin{figure*}\ContinuedFloat
\centering
     \begin{subfigure}[b]{1\linewidth}
         \centering
         \includegraphics[width=\linewidth] {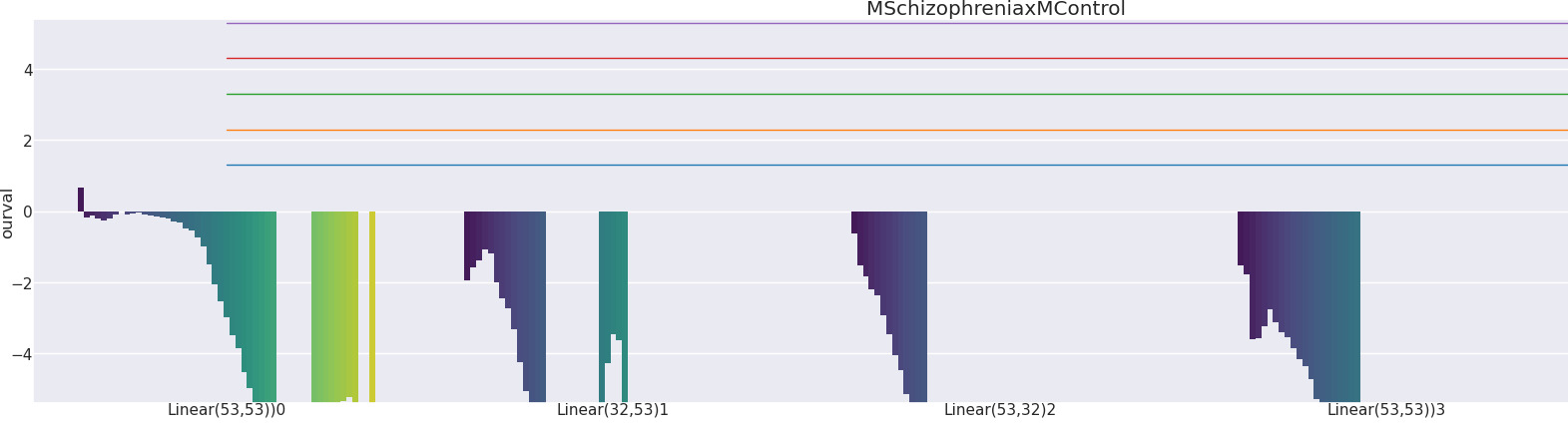}
         \caption{Significant spectral differences between HC, SZ and Sex: BERT Autoencoder with hidden dim 32 and ReLU activations}
         \label{ch6-fig:bert_cobre_g}
     \end{subfigure}
     \begin{subfigure}[b]{1\linewidth}
         \centering
         \includegraphics[width=\linewidth] {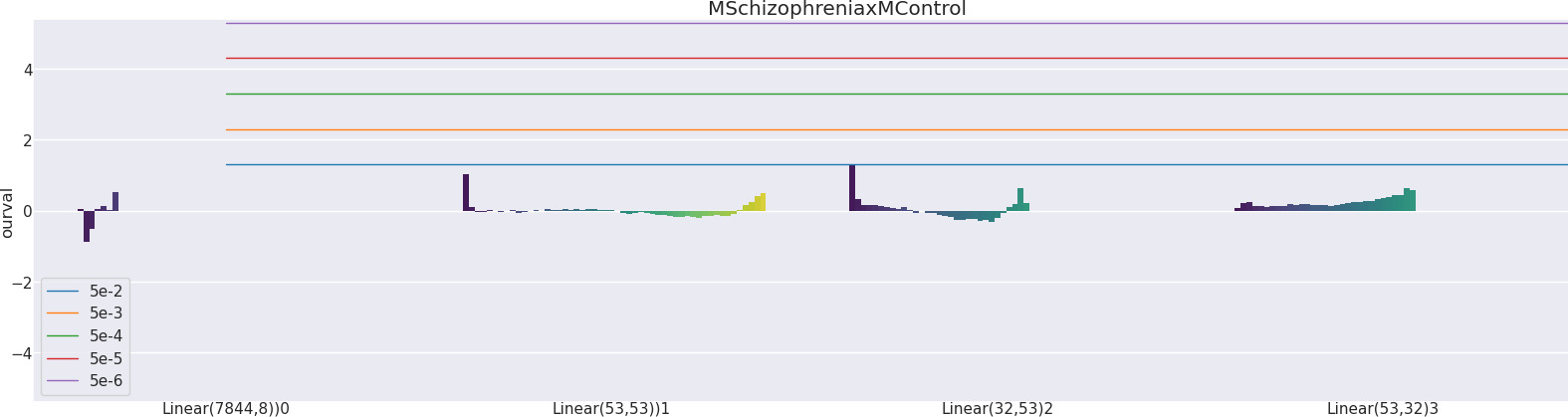}
         \caption{Significant spectral differences between HC, SZ and Sex: BERT Classifier with hidden dim 32 and ReLU activations}
         \label{ch6-fig:bert_cobre_h}
     \end{subfigure}
     \caption{Differences in Auto-Differentiation Spectra Dynamics on the COBRE data set, trained with various architectures and tasks with a BERT Transformer.}
     \label{ch6-fig:bert_cobre}
\end{figure*}

\begin{figure*}
     \caption{Differences in Auto-Differentiation Spectra Dynamics on the COBRE data set, trained with various architectures and tasks with a 1D CNN.}
     \centering
         \begin{subfigure}[b]{\linewidth}
         \centering
         \includegraphics[width=\linewidth] {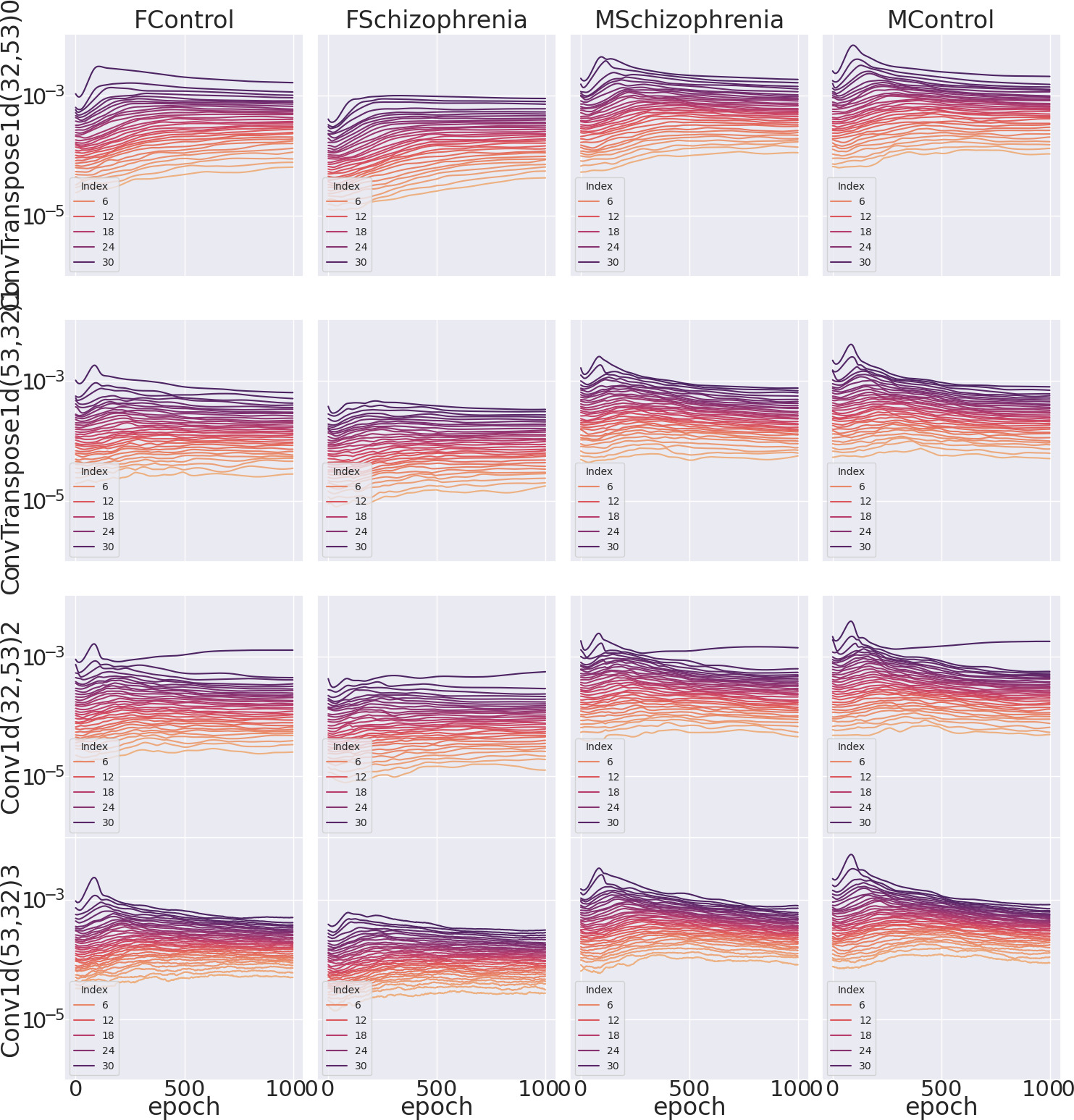}
         \caption{CNN1D Autoencoder with hidden dim 32 and ReLU activations}
         \label{ch6-fig:cnn1d_cobre_a}
     \end{subfigure}
\end{figure*}
\begin{figure*}\ContinuedFloat
              \begin{subfigure}[b]{\linewidth}
         \centering
         \includegraphics[width=\linewidth] {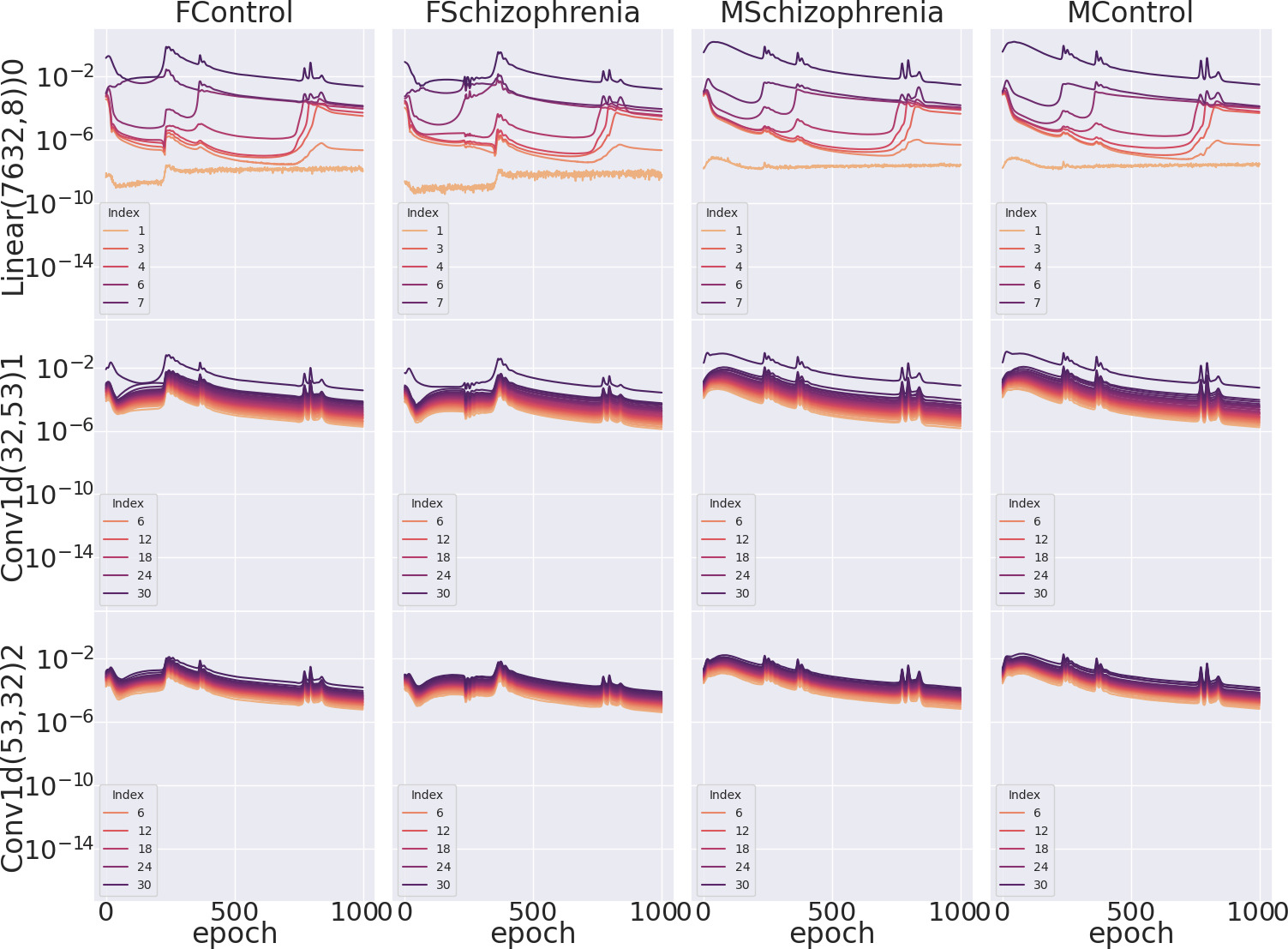}
         \caption{CNN1D Classifier with hidden dim 32 and ReLU activations}
         \label{ch6-fig:cnn1d_cobre_b}
     \end{subfigure}
\end{figure*}
\begin{figure*}\ContinuedFloat
\centering
     \begin{subfigure}[b]{\linewidth}
         \centering
         \includegraphics[width=\linewidth] {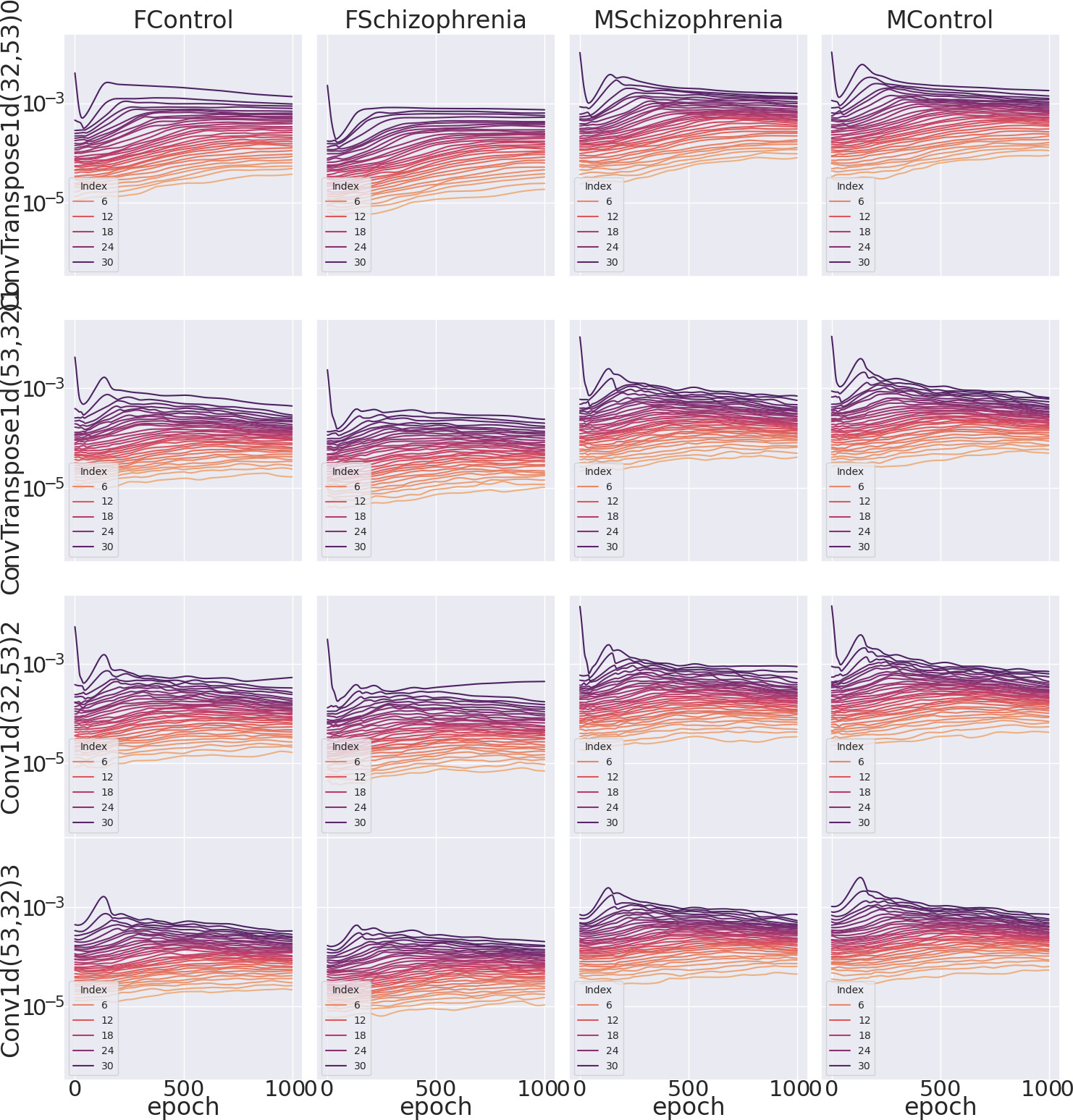}
         \caption{CNN1D Autoencoder with hidden dim 32 and Sigmoid activations}
         \label{ch6-fig:cnn1d_cobre_c}
     \end{subfigure}
\end{figure*}
\begin{figure*}\ContinuedFloat
     \begin{subfigure}[b]{\linewidth}
         \centering
         \includegraphics[width=\linewidth] {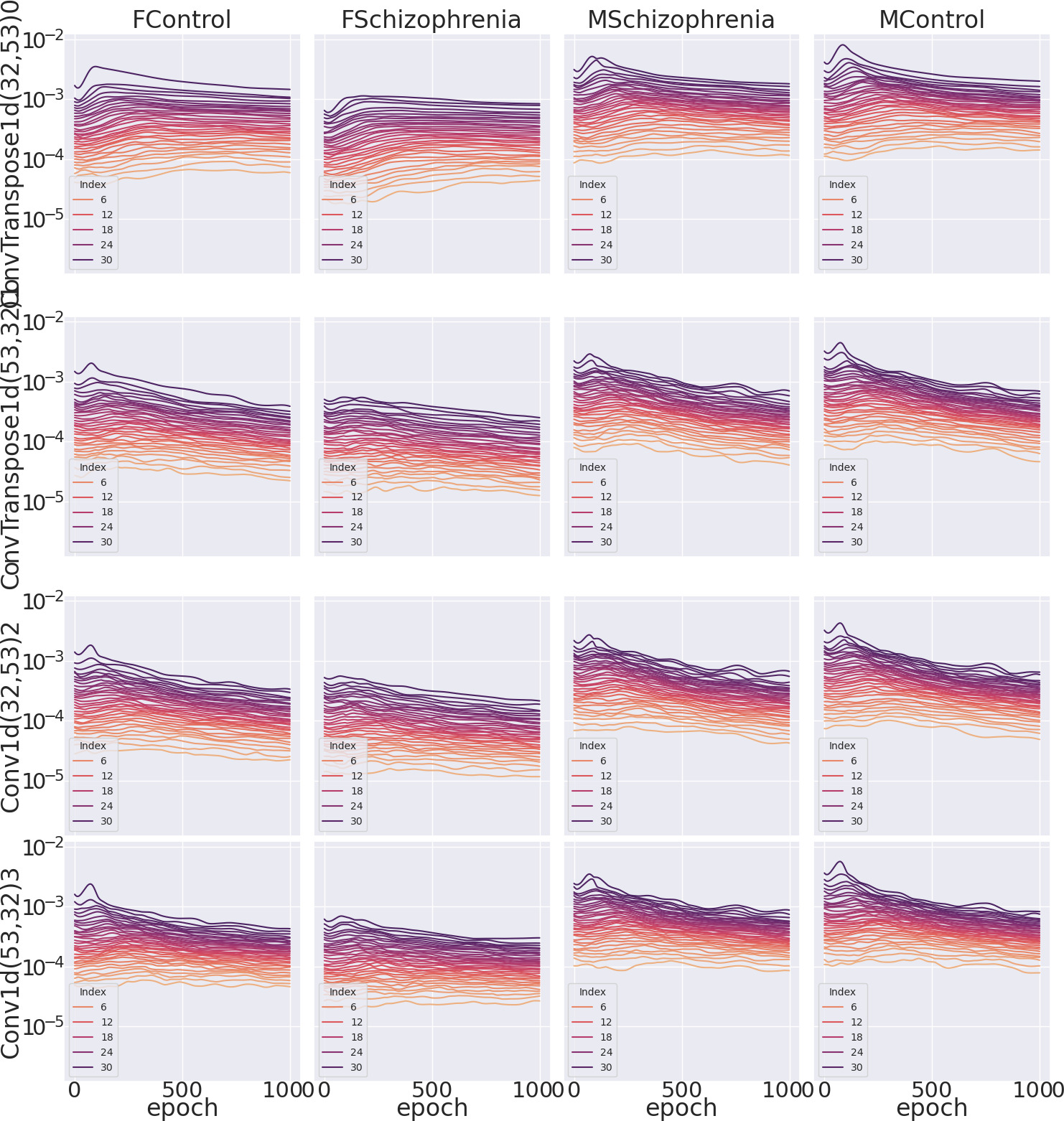}
         \caption{CNN1D Autoencoder with hidden dim 32 and Tanh activations}
         \label{ch6-fig:cnn1d_cobre_d}
     \end{subfigure}
\end{figure*}
\begin{figure*}\ContinuedFloat
\centering
     \begin{subfigure}[b]{\linewidth}
         \centering
         \includegraphics[width=\linewidth] {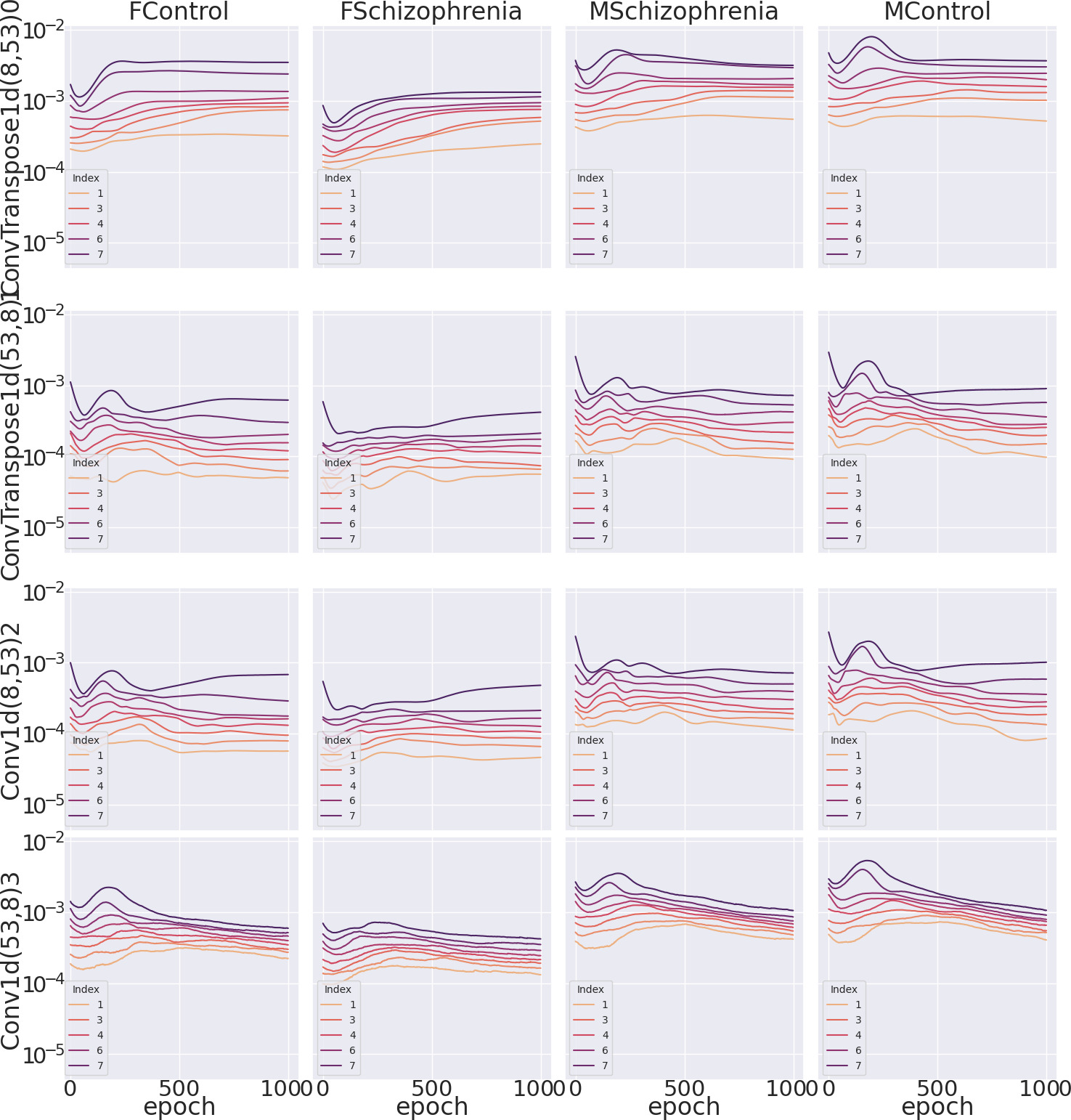}
         \caption{CNN1D Autoencoder with hidden dim 8 and ReLU activations}
         \label{ch6-fig:cnn1d_cobre_e}
     \end{subfigure}
\end{figure*}
\begin{figure*}\ContinuedFloat
          \begin{subfigure}[b]{\linewidth}
         \centering
         \includegraphics[width=\linewidth] {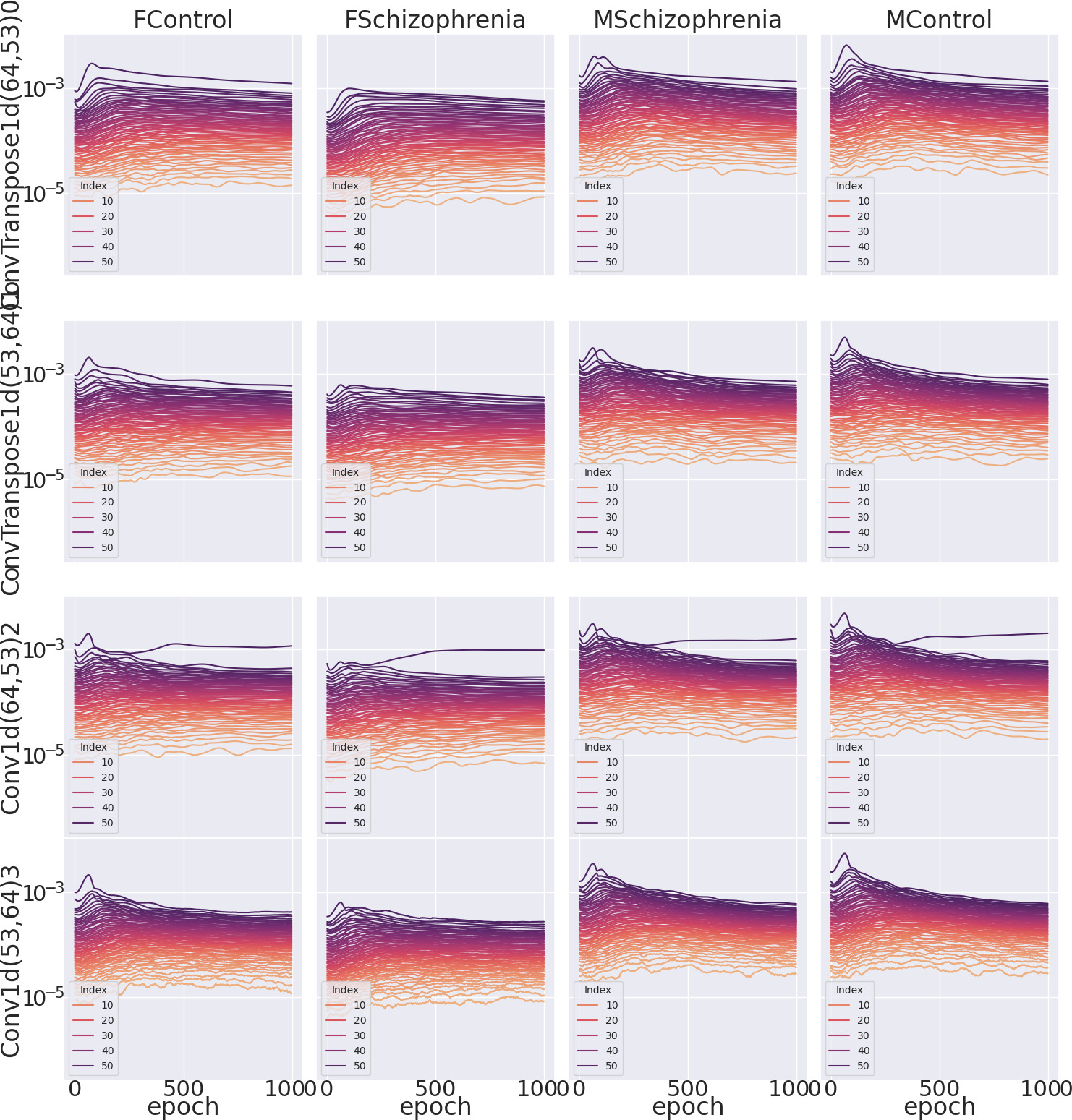}
         \caption{CNN1D Autoencoder with hidden dim 64 and ReLU activations}
         \label{ch6-fig:cnn1d_cobre_f}
     \end{subfigure}
\end{figure*}
\begin{figure*}\ContinuedFloat
\centering
     \begin{subfigure}[b]{1\linewidth}
         \centering
         \includegraphics[width=\linewidth] {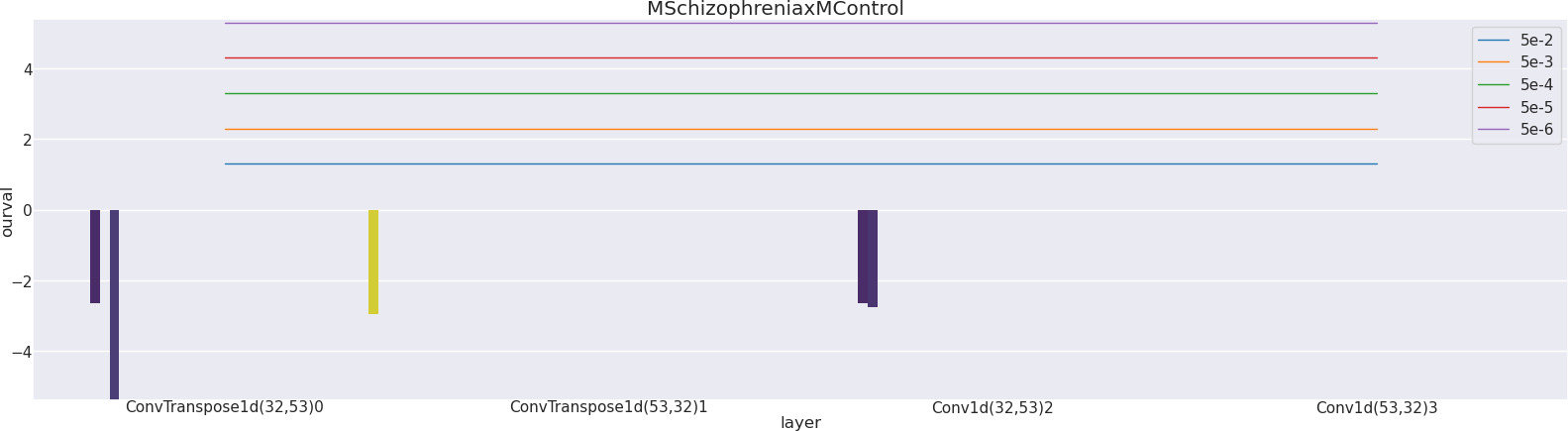}
         \caption{Significant spectral differences between HC, SZ and Sex: CNN1D Autoencoder with hidden dim 32 and ReLU activations}
         \label{ch6-fig:cnn1d_cobre_g}
     \end{subfigure}
     \begin{subfigure}[b]{1\linewidth}
         \centering
         \includegraphics[width=\linewidth] {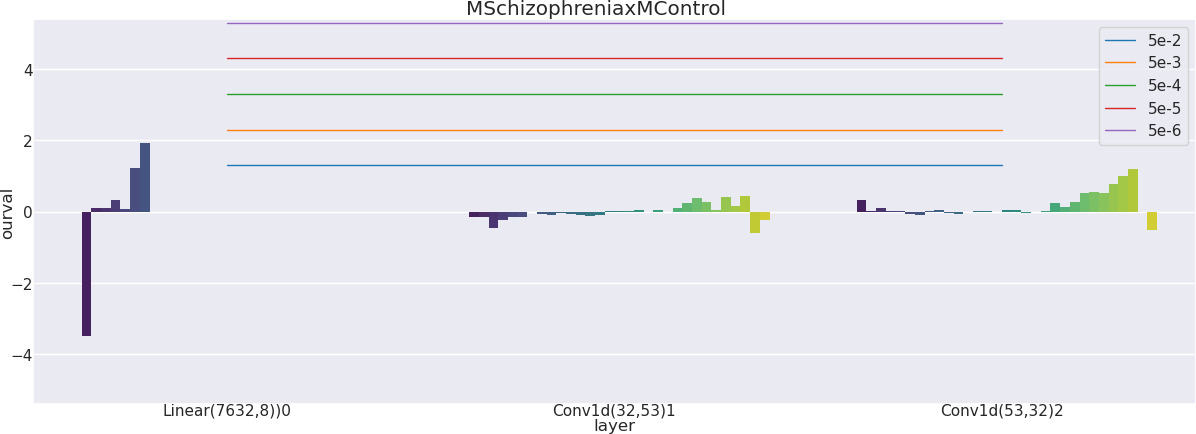}
         \caption{Significant spectral differences between HC, SZ and Sex: CNN1D Classifier with hidden dim 32 and ReLU activations}
         \label{ch6-fig:cnn1d_cobre_h}
     \end{subfigure}
     \caption{Differences in Auto-Differentiation Spectra Dynamics on the COBRE data set, trained with various architectures and tasks with a 1D CNN.}
     \label{ch6-fig:cnn1d_cobre}
\end{figure*}

\clearpage
\begin{figure*}
     \caption{Differences in Auto-Differentiation Spectra Dynamics on the COBRE data set, trained with various architectures and tasks with a 3D CNN.}
     \centering
         \begin{subfigure}[b]{\linewidth}
         \centering
         \includegraphics[width=\linewidth] {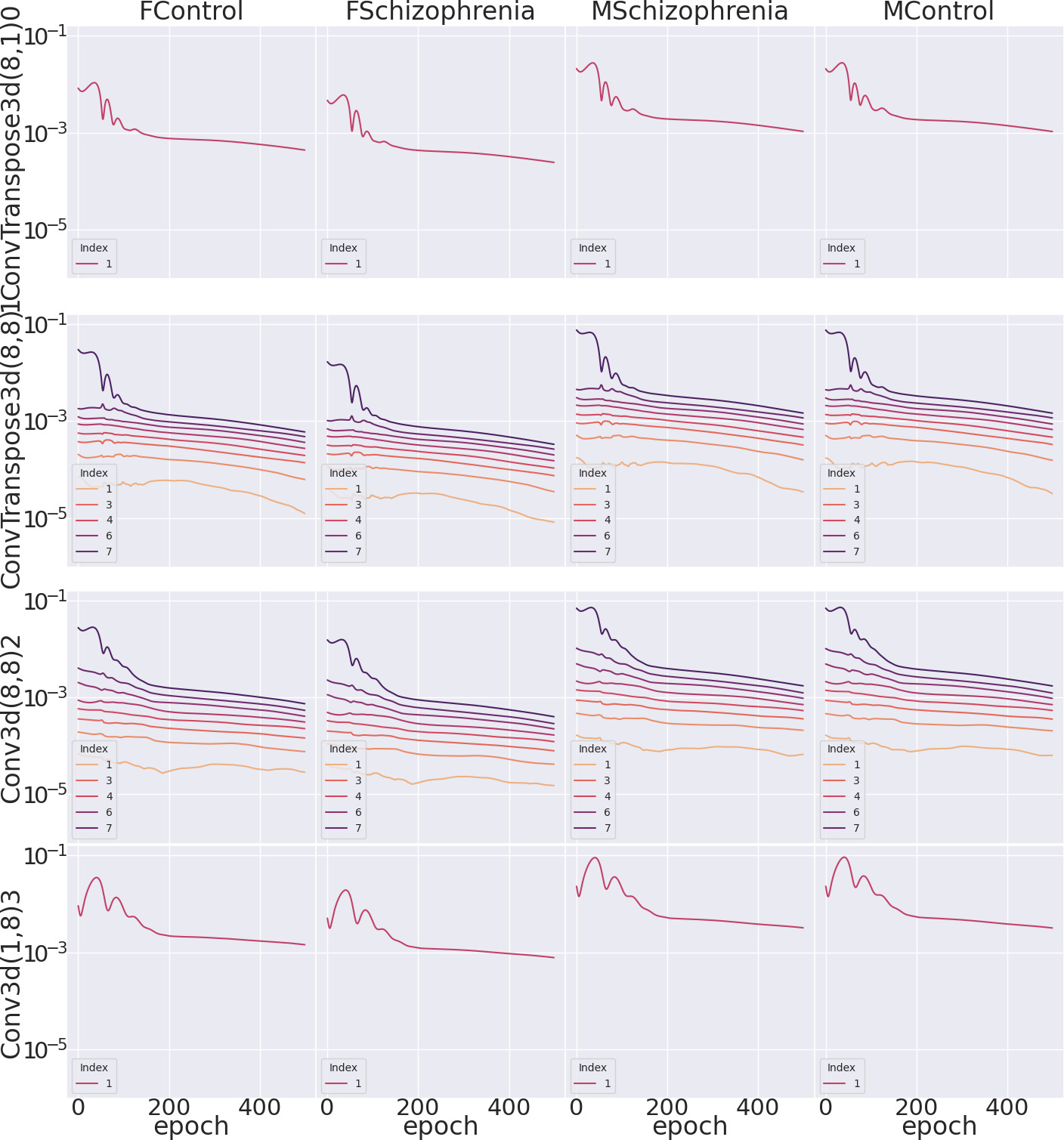}
         \caption{CNN3D Autoencoder with hidden dim 32 and ReLU activations}
         \label{ch6-fig:cnn3d_cobre_a}
     \end{subfigure}
\end{figure*}
\begin{figure*}\ContinuedFloat
              \begin{subfigure}[b]{\linewidth}
         \centering
         \includegraphics[width=\linewidth] {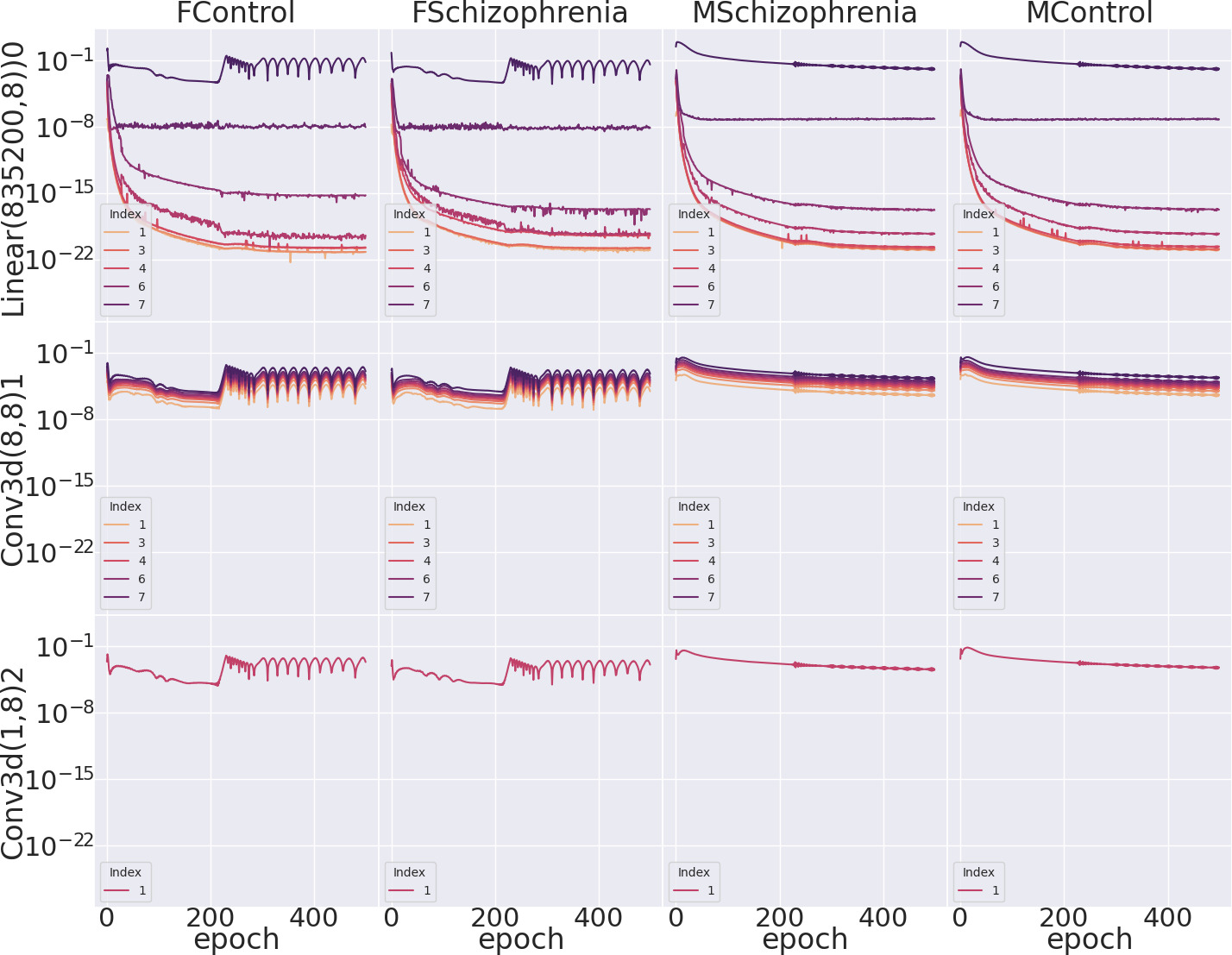}
         \caption{CNN3D Classifier with hidden dim 32 and ReLU activations}
         \label{ch6-fig:cnn3d_cobre_b}
     \end{subfigure}
\end{figure*}
\begin{figure*}\ContinuedFloat
\centering
     \begin{subfigure}[b]{\linewidth}
         \centering
         \includegraphics[width=\linewidth] {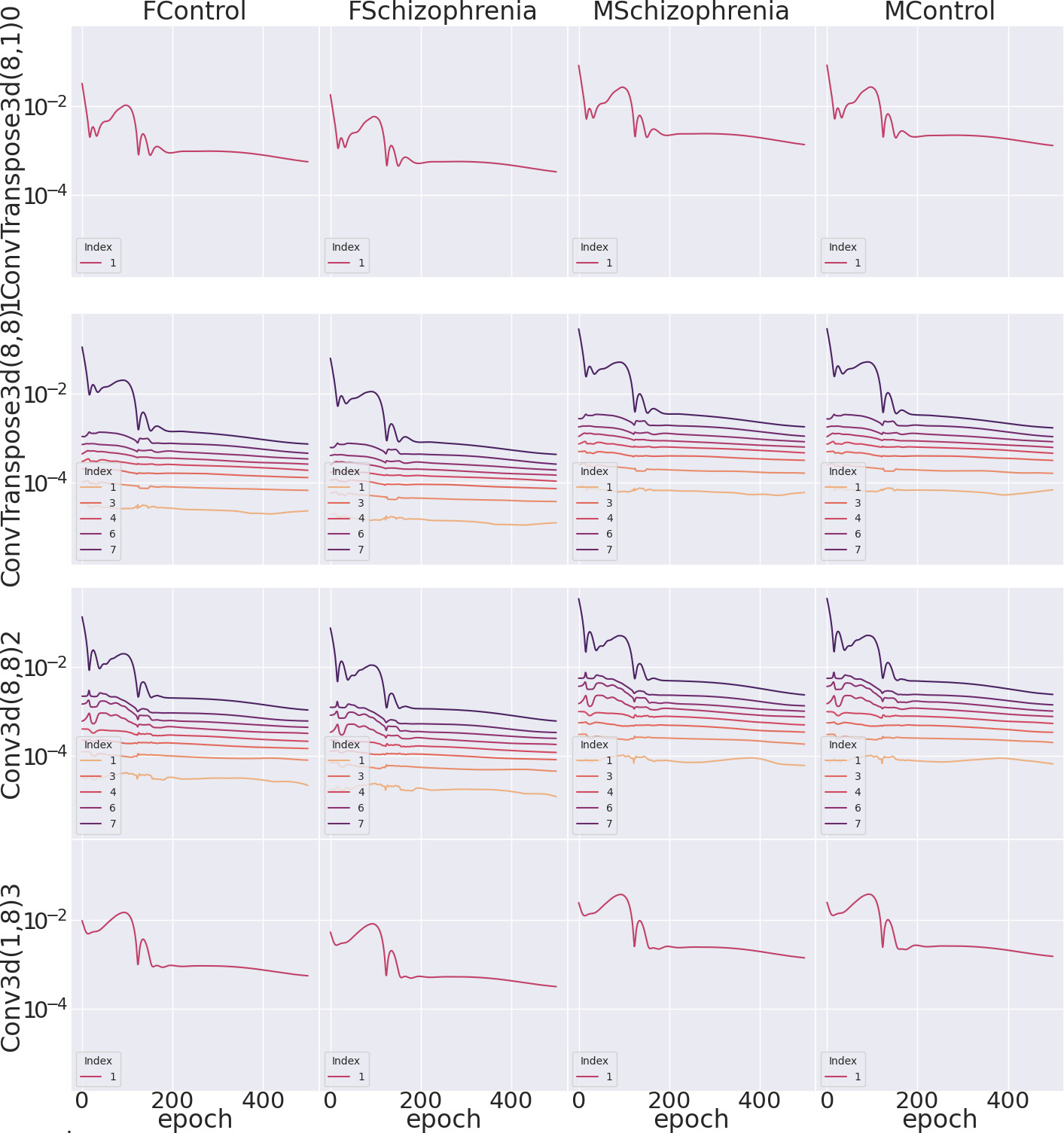}
         \caption{CNN3D Autoencoder with hidden dim 32 and Sigmoid activations}
         \label{ch6-fig:cnn3d_cobre_c}
     \end{subfigure}
\end{figure*}
\begin{figure*}\ContinuedFloat
     \begin{subfigure}[b]{\linewidth}
         \centering
         \includegraphics[width=\linewidth] {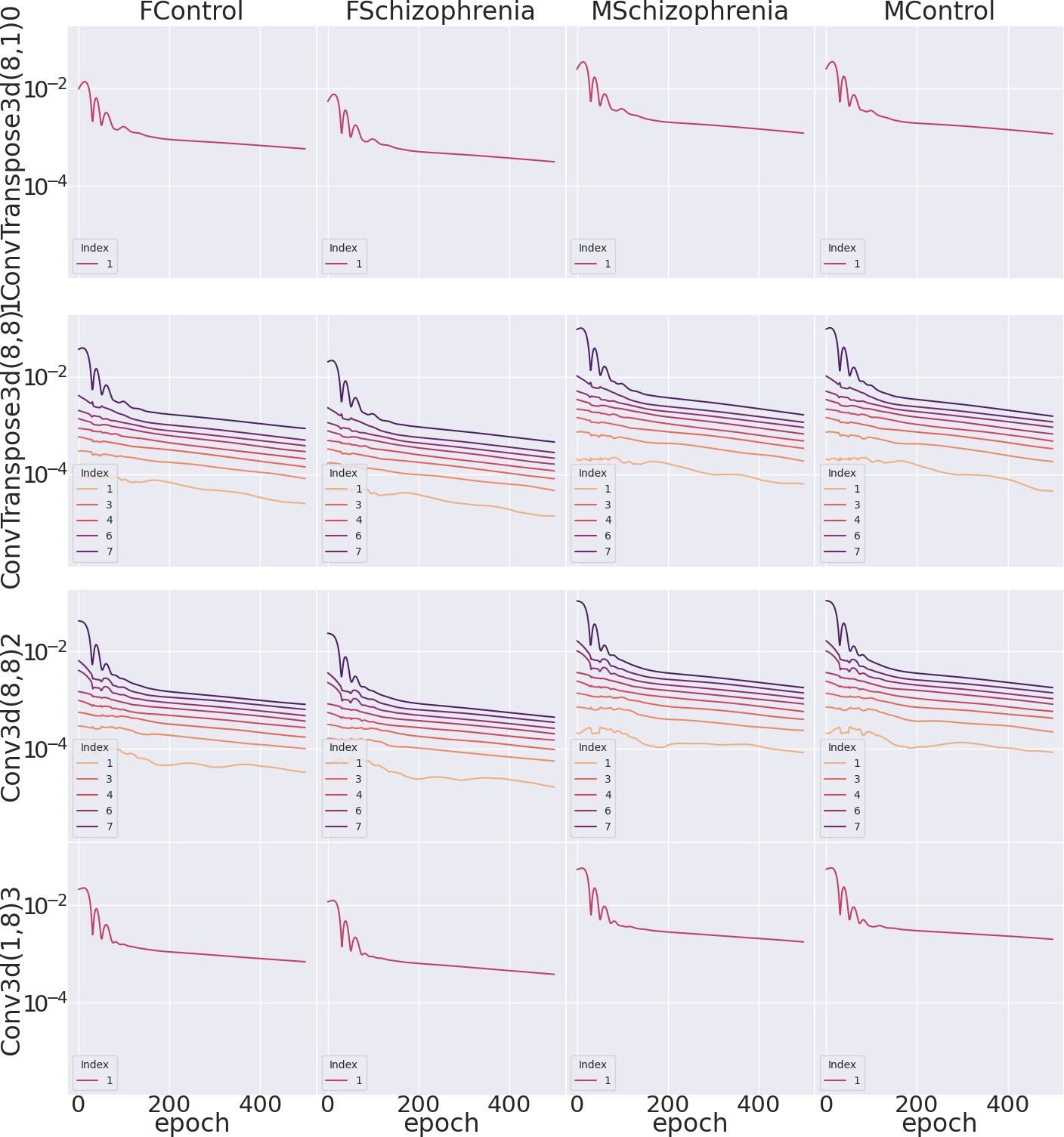}
         \caption{CNN3D Autoencoder with hidden dim 32 and Tanh activations}
         \label{ch6-fig:cnn3d_cobre_d}
     \end{subfigure}
\end{figure*}
\begin{figure*}\ContinuedFloat
\centering
     \begin{subfigure}[b]{\linewidth}
         \centering
         \includegraphics[width=\linewidth] {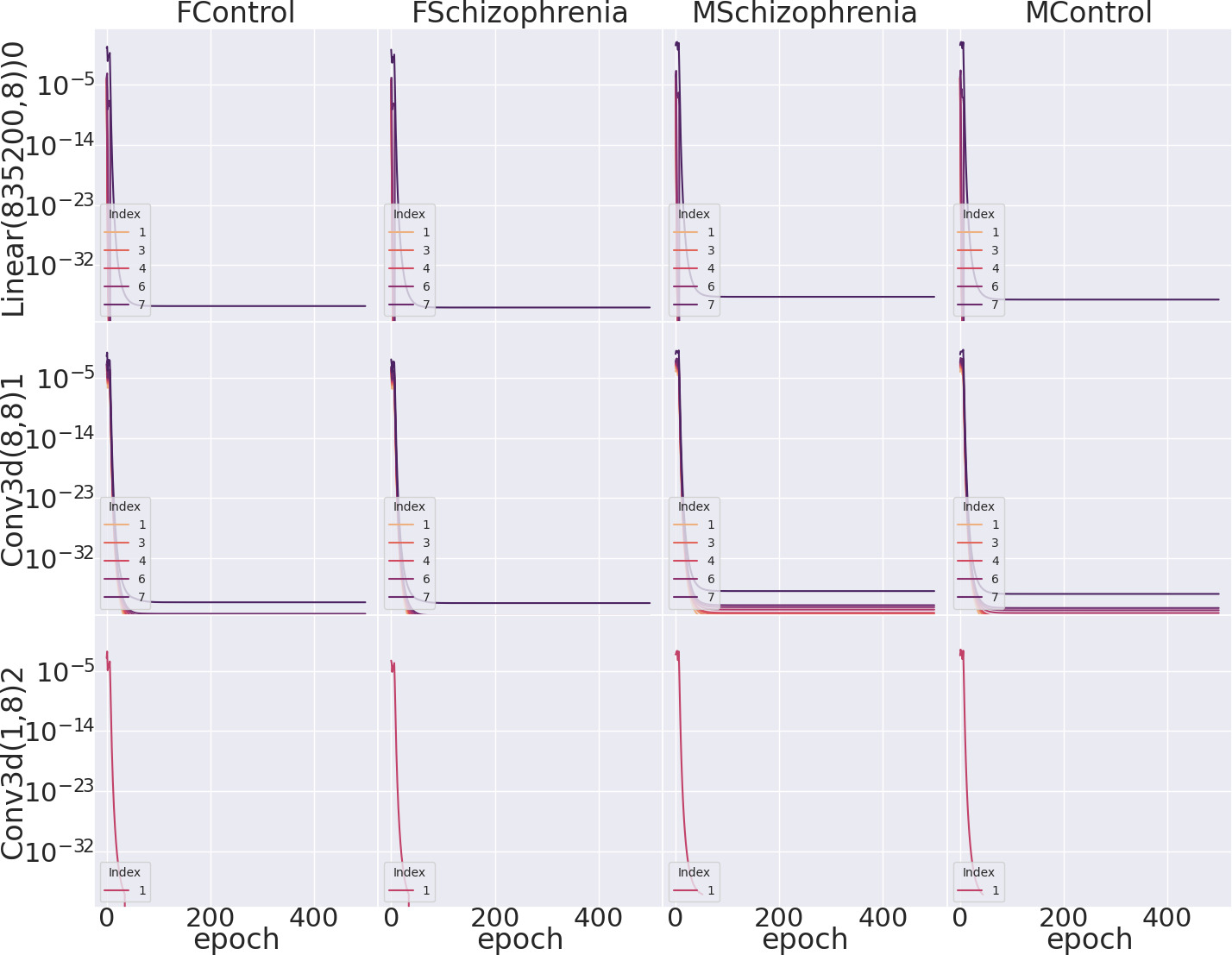}
         \caption{CNN3D Autoencoder with hidden dim 8 and ReLU activations}
         \label{ch6-fig:cnn3d_cobre_e}
     \end{subfigure}
\end{figure*}
\begin{figure*}\ContinuedFloat     
          \begin{subfigure}[b]{\linewidth}
         \centering
         \includegraphics[width=\linewidth] {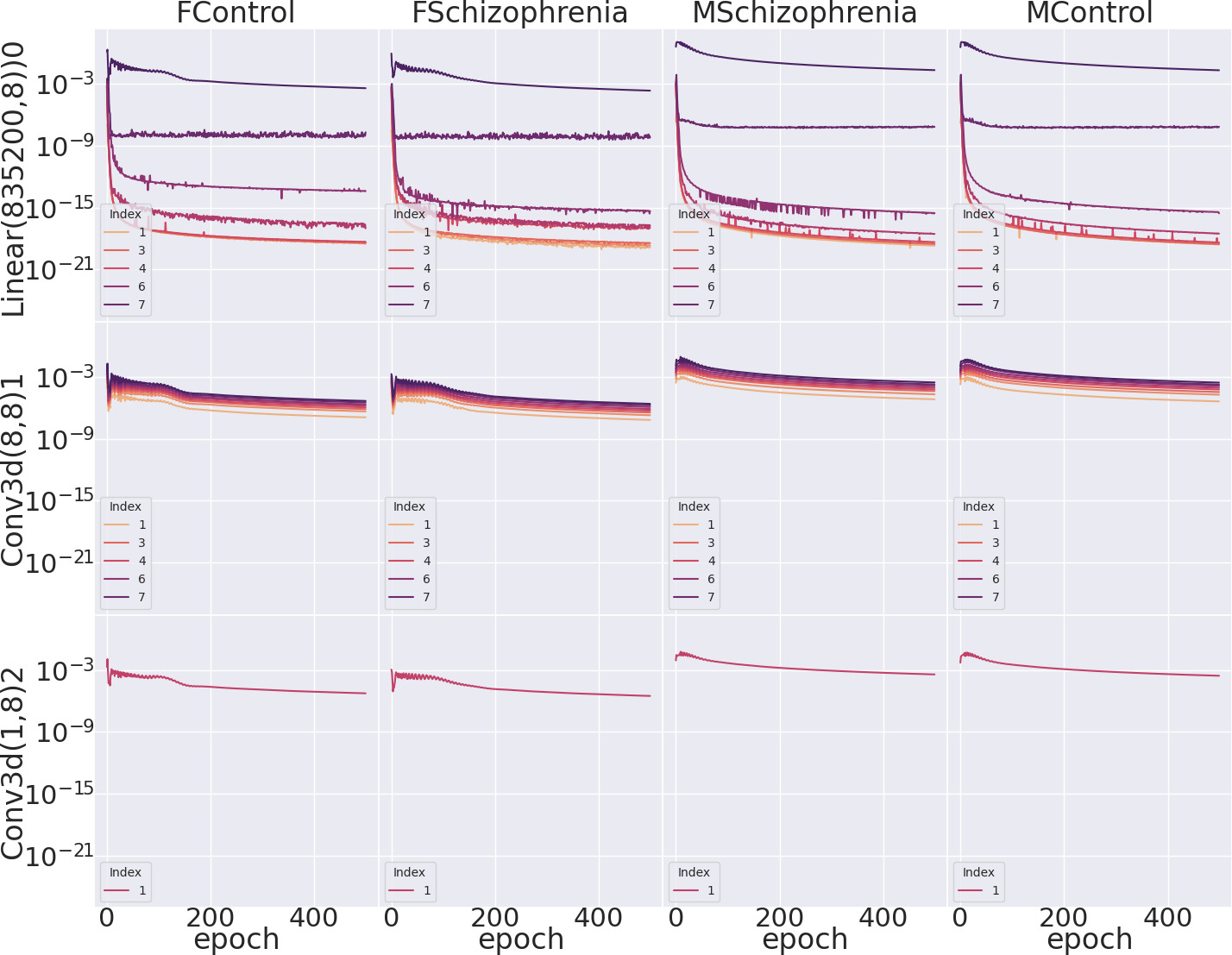}
         \caption{CNN3D Autoencoder with hidden dim 64 and ReLU activations}
         \label{ch6-fig:cnn3d_cobre_f}
     \end{subfigure}
\end{figure*}
\begin{figure*}\ContinuedFloat
\centering
     \begin{subfigure}[b]{1\linewidth}
         \centering
         \includegraphics[width=\linewidth] {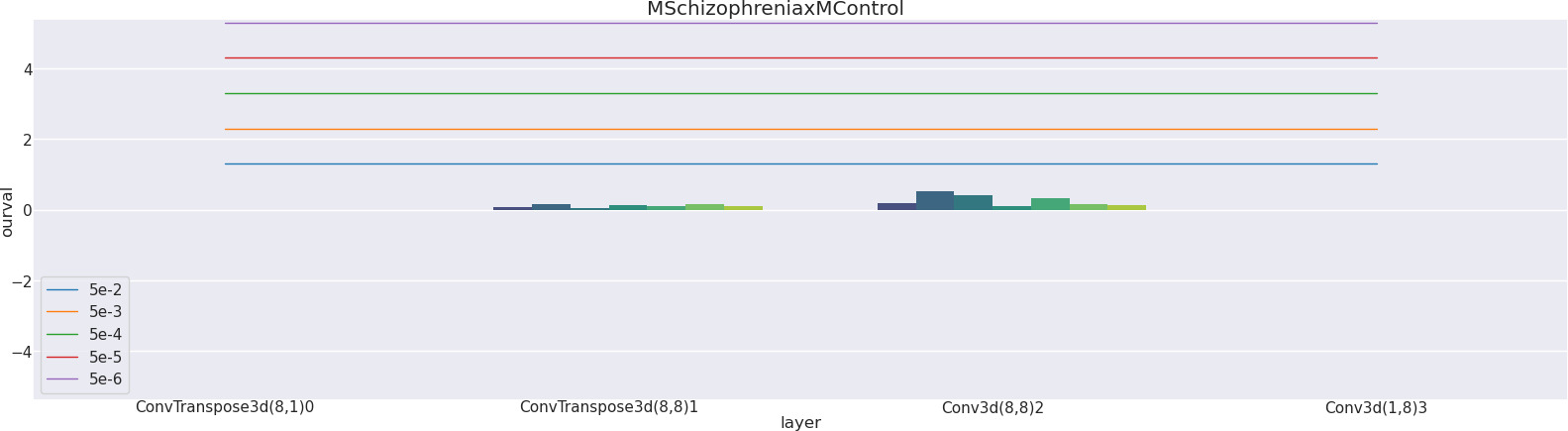}
         \caption{Significant spectral differences between HC, SZ and Sex: CNN3D Autoencoder with hidden dim 32 and ReLU activations}
         \label{ch6-fig:cnn3d_cobre_g}
     \end{subfigure}
     \begin{subfigure}[b]{1\linewidth}
         \centering
         \includegraphics[width=\linewidth] {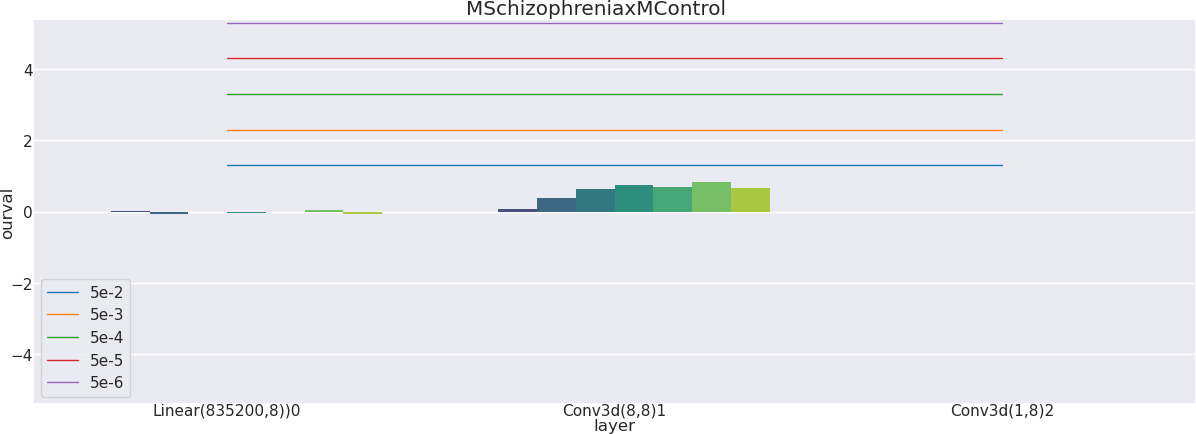}
         \caption{Significant spectral differences between HC, SZ and Sex: CNN3D Classifier with hidden dim 32 and ReLU activations}
         \label{ch6-fig:cnn3d_cobre_h}
     \end{subfigure}
     \caption{Differences in Auto-Differentiation Spectra Dynamics on the COBRE data set, trained with various architectures and tasks with a 3D CNN.}
     \label{ch6-fig:cnn3d_cobre}
\end{figure*}

\bibliographystyle{ieeetr}
\bibliography{main}

\end{document}